\documentclass[10pt,twocolumn,letterpaper]{article}

\usepackage{cvpr}              %

\usepackage[pagebackref,breaklinks,colorlinks,allcolors=cvprblue]{hyperref}
\usepackage[capitalise, capitalize]{cleveref}
\crefname{figure}{Figure}{Figures}
\crefname{table}{Table}{Tables}
\usepackage{amsmath}

\usepackage{array}
\usepackage{multirow}
\usepackage{tabularx}
\usepackage{makecell} %
\usepackage{graphicx}
\usepackage{enumitem}
\usepackage{float}
\usepackage{amssymb}
\usepackage{tikz}
\usetikzlibrary{quotes,arrows.meta,angles,calc,3d,shapes,intersections,plotmarks,positioning,backgrounds,external,shapes.geometric}
\usetikzlibrary{shadows}
\usetikzlibrary{shadows.blur}

\usepackage{epigraph}

\creflabelformat{equation}{#2#1#3}

\newcolumntype{C}{>{\centering\arraybackslash}X}

\graphicspath{{figures}}

\usepackage{amsmath,amsfonts,bm}

\def\1{\bm{1}}

\def\rvc{{\mathbf{c}}}
\def\rvd{{\mathbf{d}}}

\def\rvn{{\mathbf{n}}}

\def\rvr{{\mathbf{r}}}

\def\rvx{{\mathbf{x}}}

\DeclareMathAlphabet{\mathsfit}{\encodingdefault}{\sfdefault}{m}{sl}
\SetMathAlphabet{\mathsfit}{bold}{\encodingdefault}{\sfdefault}{bx}{n}

\newcommand{\transpose}{^\mathsf{T}}

\usepackage{xspace}
\makeatletter
\DeclareRobustCommand\onedot{\futurelet\@let@token\@onedot}
\def\@onedot{\ifx\@let@token.\else.\null\fi\xspace}

\def\eg{e.g\onedot}

\usepackage{amssymb}
\usepackage{pifont}

\newcommand{\method}{Snellcaster\xspace}

\newcommand{\refl}{\text{\reflectbox{R}}}
\newcommand{\refr}{\text{\rotatebox[origin=c]{20}{R}}}

\makeatletter
\renewcommand\paragraph{%
  \@startsection{paragraph}%
                {4}%
                {\z@}%
                {-1.25ex \@plus -1ex \@minus -0.2ex}%
                {-1.5ex \@plus 0.2ex}%
                {\normalfont\normalsize\bfseries}}%
\makeatother

\definecolor{cvprblue}{rgb}{0.21,0.49,0.74}

\def\confName{CVPR}
\def\confYear{2026}

\newcommand{\splashimage}[9]{%
    \def\mainimage{#1}%
    \def\greenimage{#2}%
    \def\trimA{#3}%
    \def\trimB{#4}%
    \def\trimC{#5}%
    \def\trimD{#6}%
    \def\trimE{#7}%
    \def\trimF{#8}%
    \def\trimG{#9}%
    \expandafter\splashimagecontinued
}

\newcommand{\splashimagecontinued}[1]{%
    \def\trimH{#1}%
    \begin{tikzpicture}[baseline=(mainimage.base)]

        \node[inner sep=0pt] (mainimage)
            {\includegraphics[width=\linewidth]{\mainimage}};

        \node[
            anchor=north east,
            xshift=-0.15pt,
            yshift=-0.15pt,
            inner sep=0pt,
            draw=green,
            line width=1pt
        ] at (mainimage.north east)
        {
            \includegraphics[
                width=0.2\linewidth,
                trim={{\trimA} {\trimB} {\trimC} {\trimD}},
                clip
            ]{\greenimage}
        };

        \node[
            anchor=north east,
            xshift=-0.2\linewidth-1pt,
            yshift=-0.19pt,
            inner sep=0pt,
            draw=red,
            line width=1pt
        ] at (mainimage.north east)
        {
            \includegraphics[
                width=0.2\linewidth,
                trim={{\trimE} {\trimF} {\trimG} {\trimH}},
                clip
            ]{\mainimage}
        };

    \end{tikzpicture}%
}

\newcommand\splashfigure{%
    \centering
    \captionsetup{type=figure}
    \setlength{\tabcolsep}{1pt}
    \begin{tabularx}{\linewidth}{@{}l@{\hskip 3pt}CCCC@{}}

        \rotatebox{90}{\parbox{2.5cm}{\centering Inpaint}} &
        \splashimage{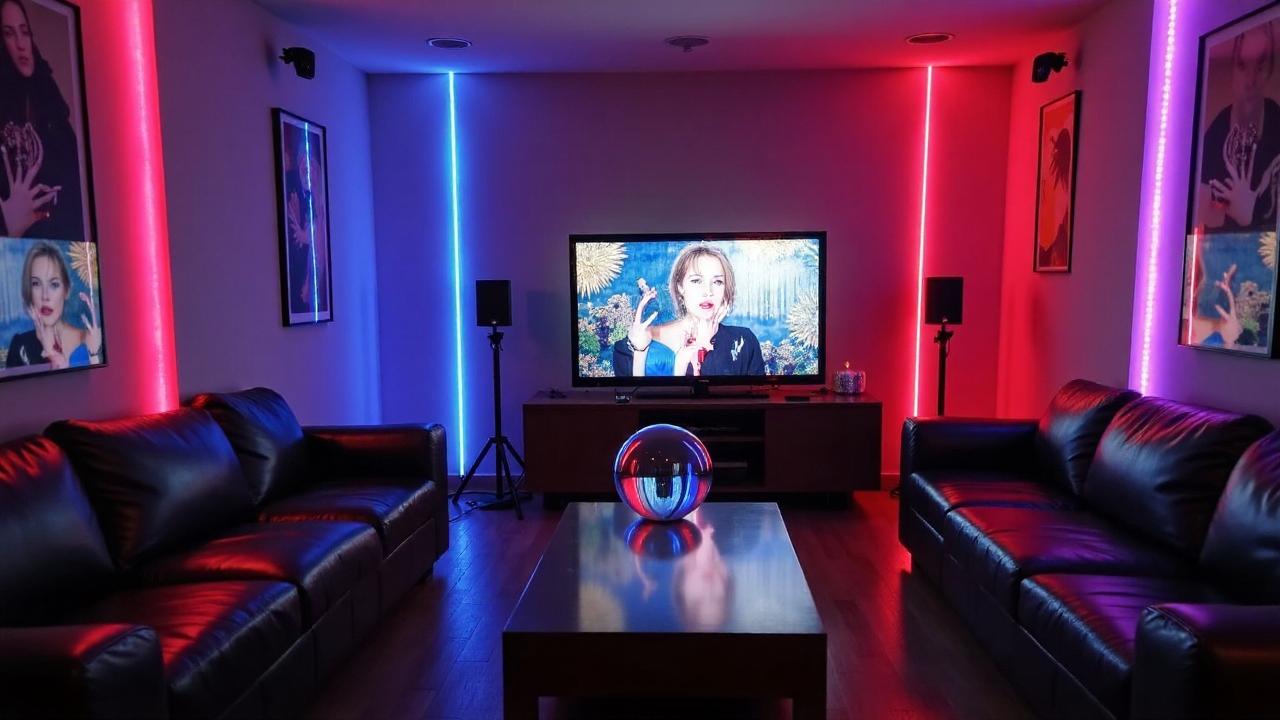}{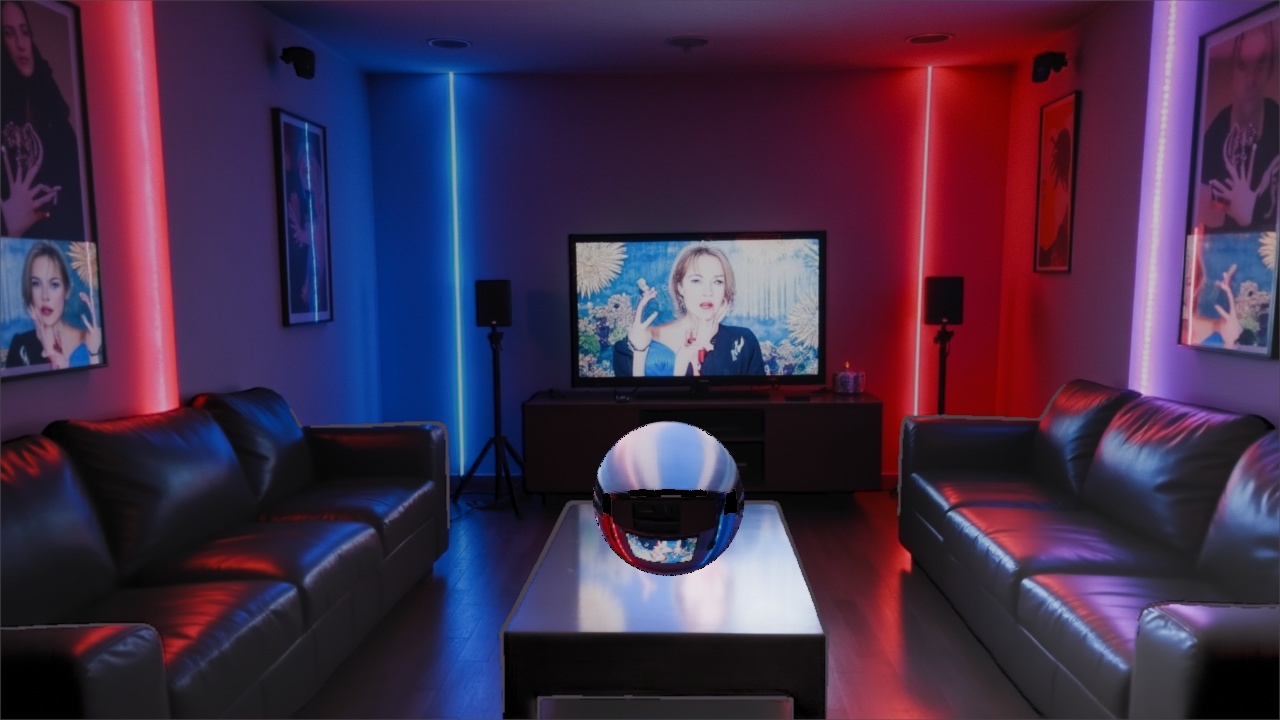}{585}{140}{525}{410}{612.5}{195}{562.5}{420} &
        \splashimage{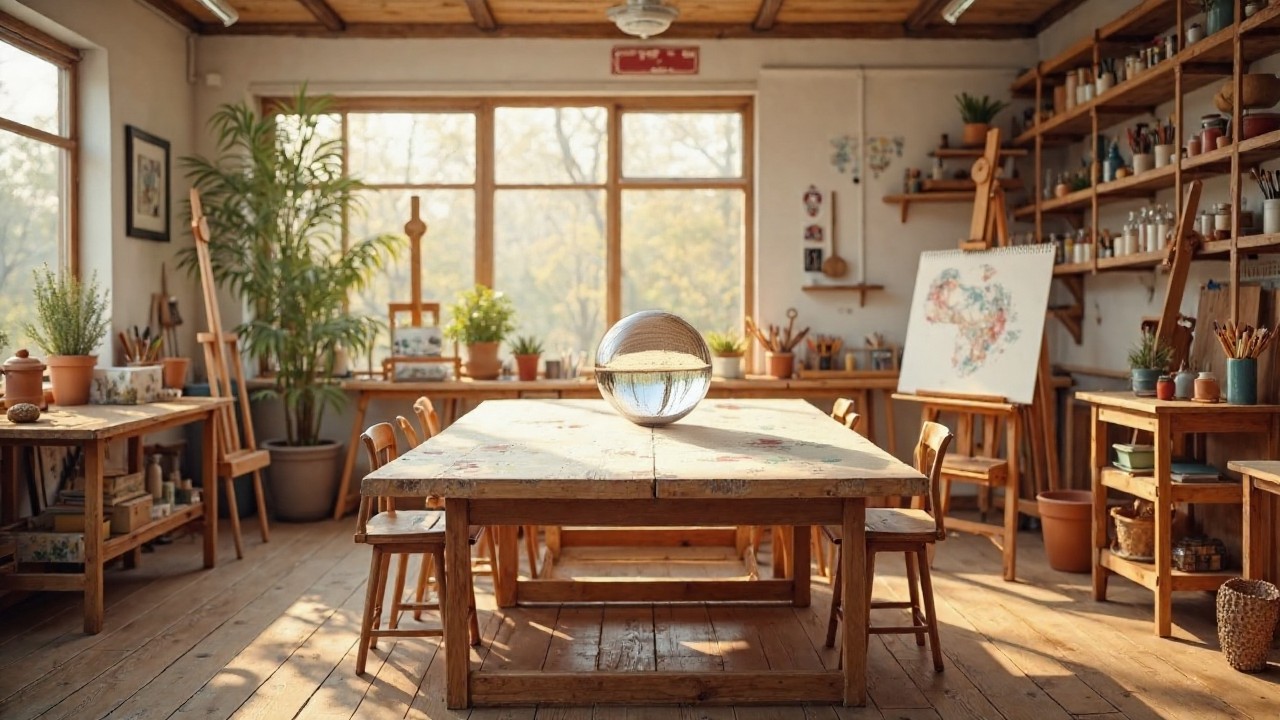}{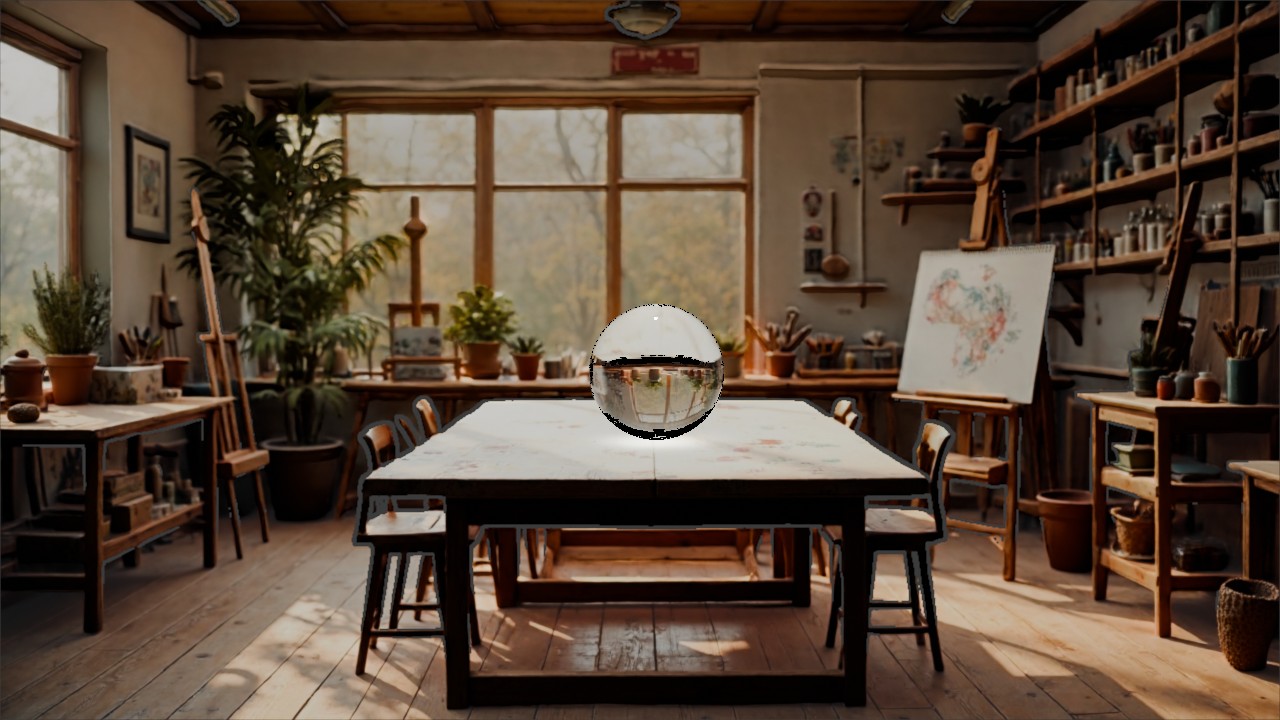}{582.5}{274}{552.5}{301}{588}{285}{562}{305} &
        \splashimage{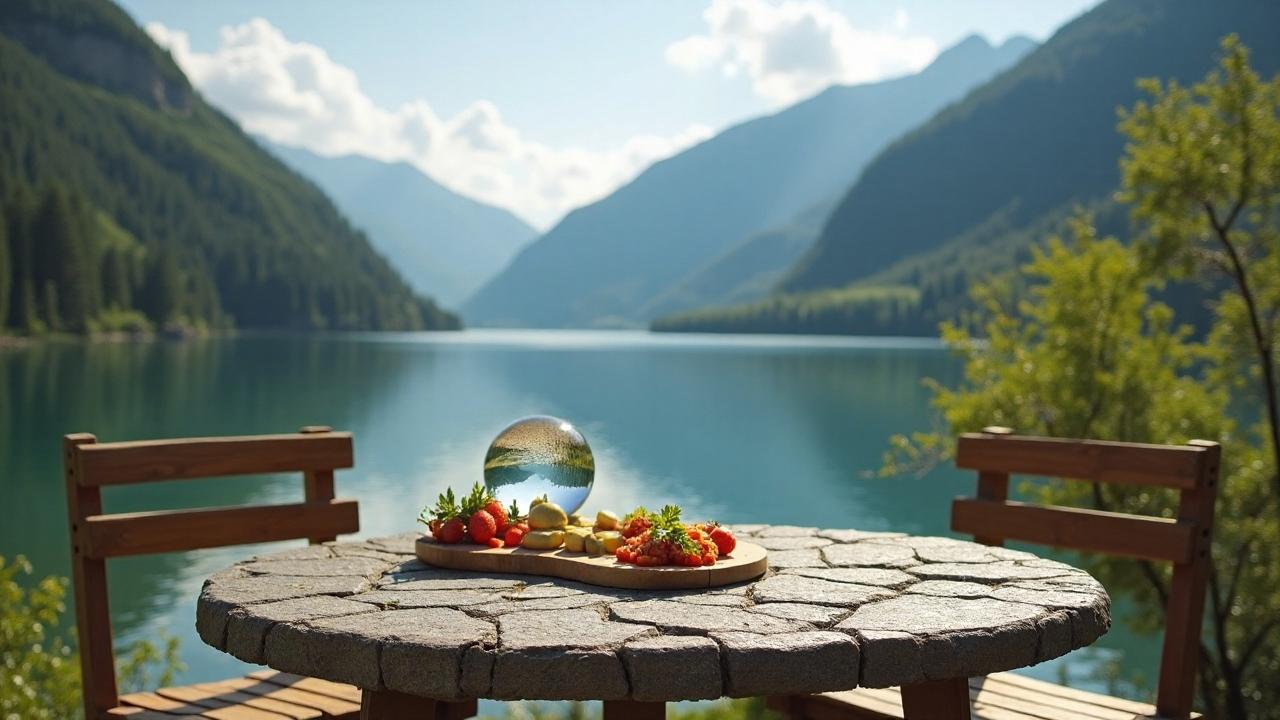}{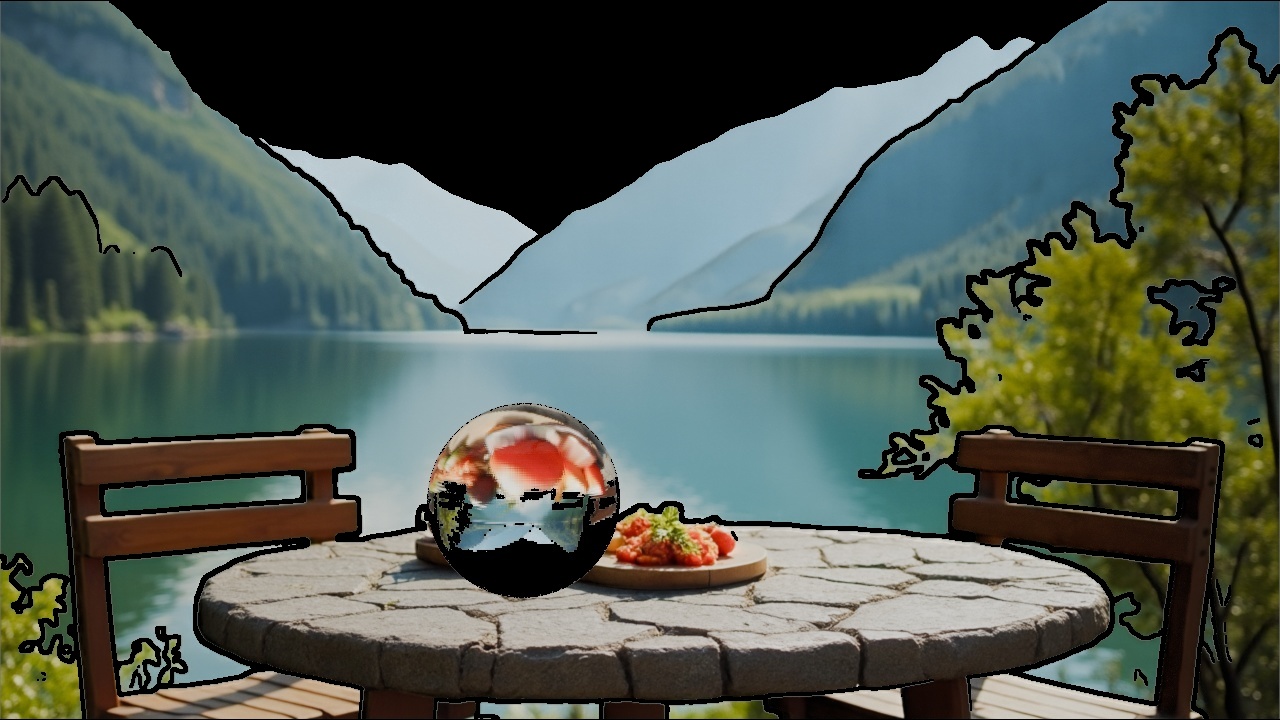}{428}{116}{660}{396}{483}{179}{685}{413} &
        \splashimage{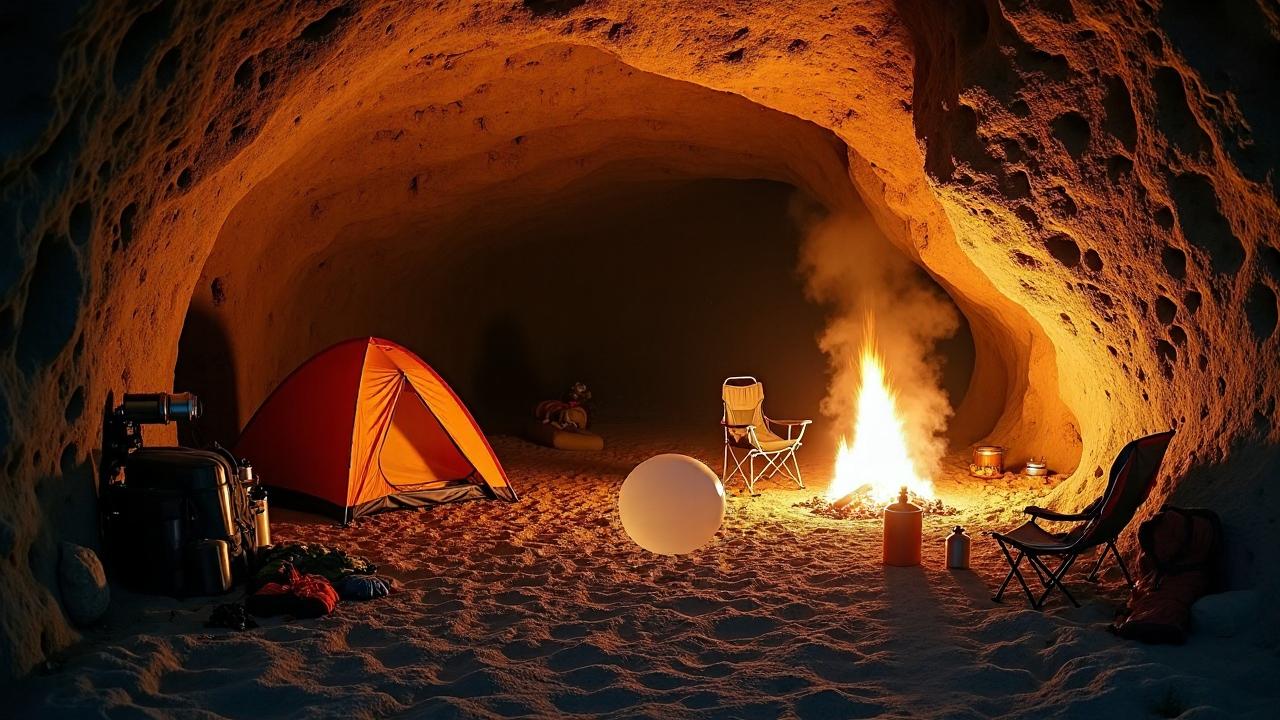}{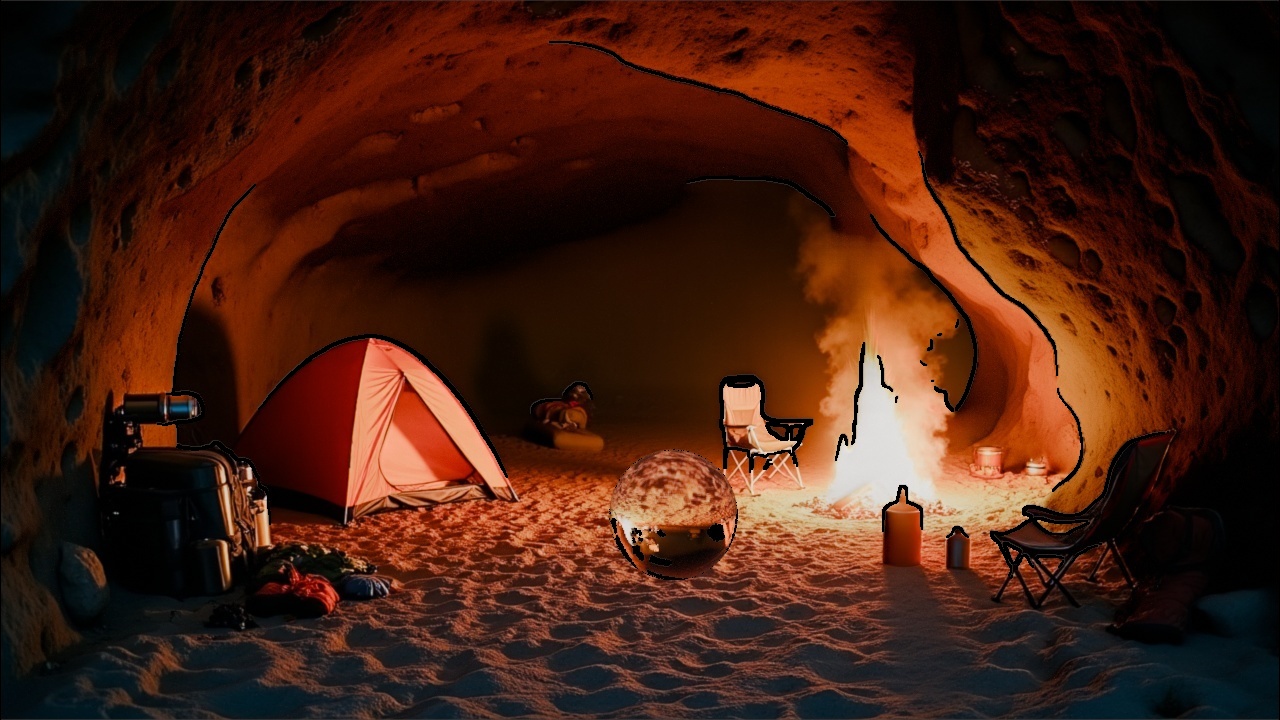}{605}{136}{538}{447}{615}{160}{552}{447}
        \\[-5pt]
        
        \rotatebox{90}{\parbox{2.5cm}{\centering Ours}} &
        \splashimage{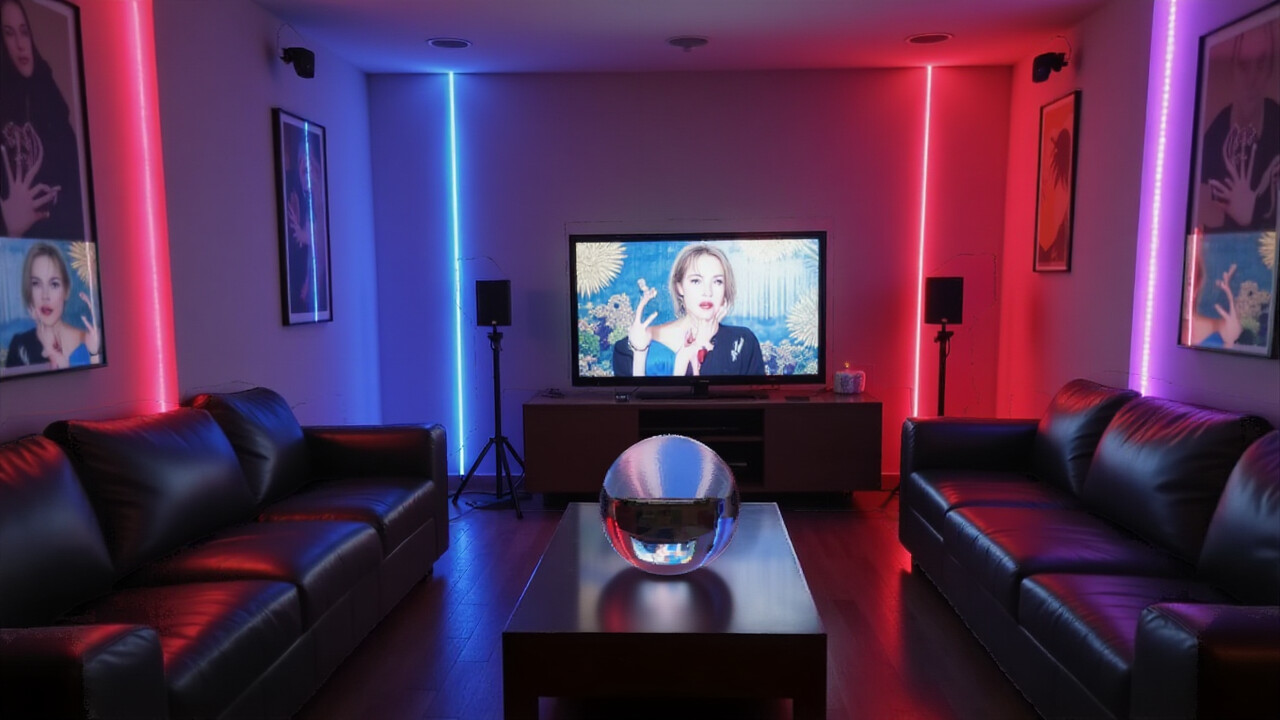}{figures/splash/blender_karaoke.jpg}{585}{140}{525}{410}{592.5}{140}{532.5}{425} &
        \splashimage{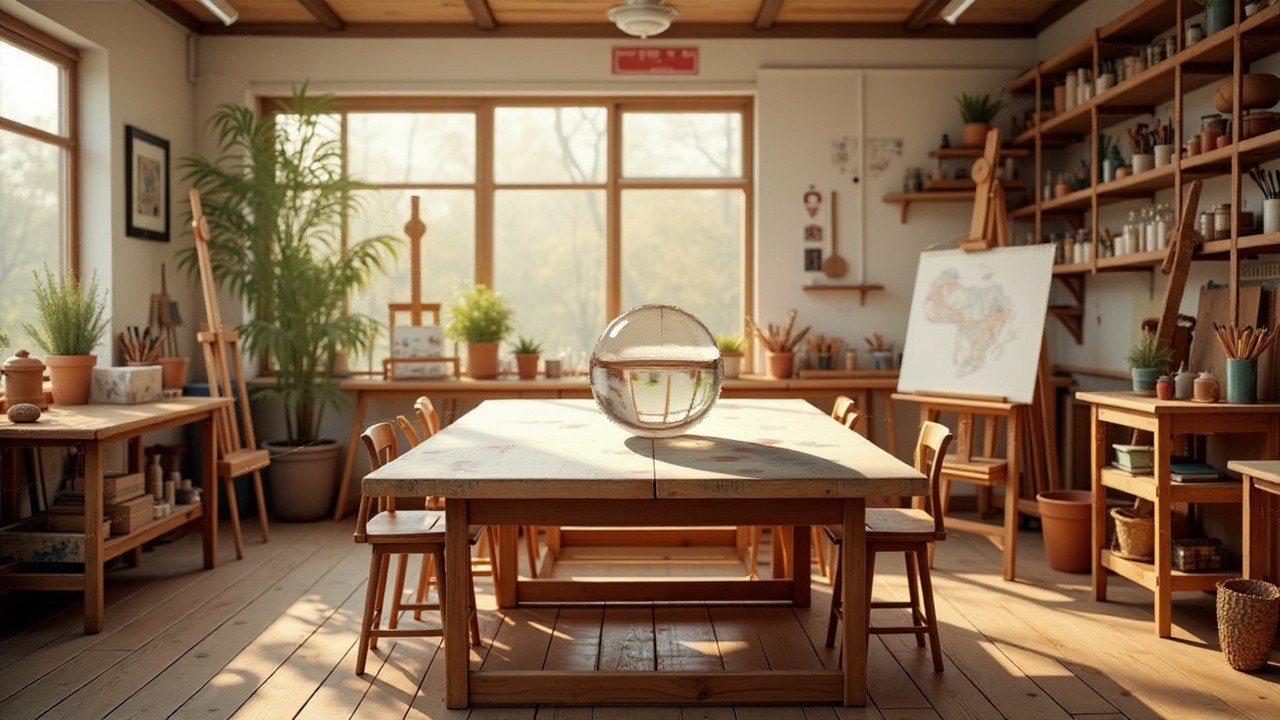}{figures/splash/artroom_9577_maskedgt.jpg}{582.5}{274}{552.5}{301}{582.5}{274}{552.5}{301} &
        \splashimage{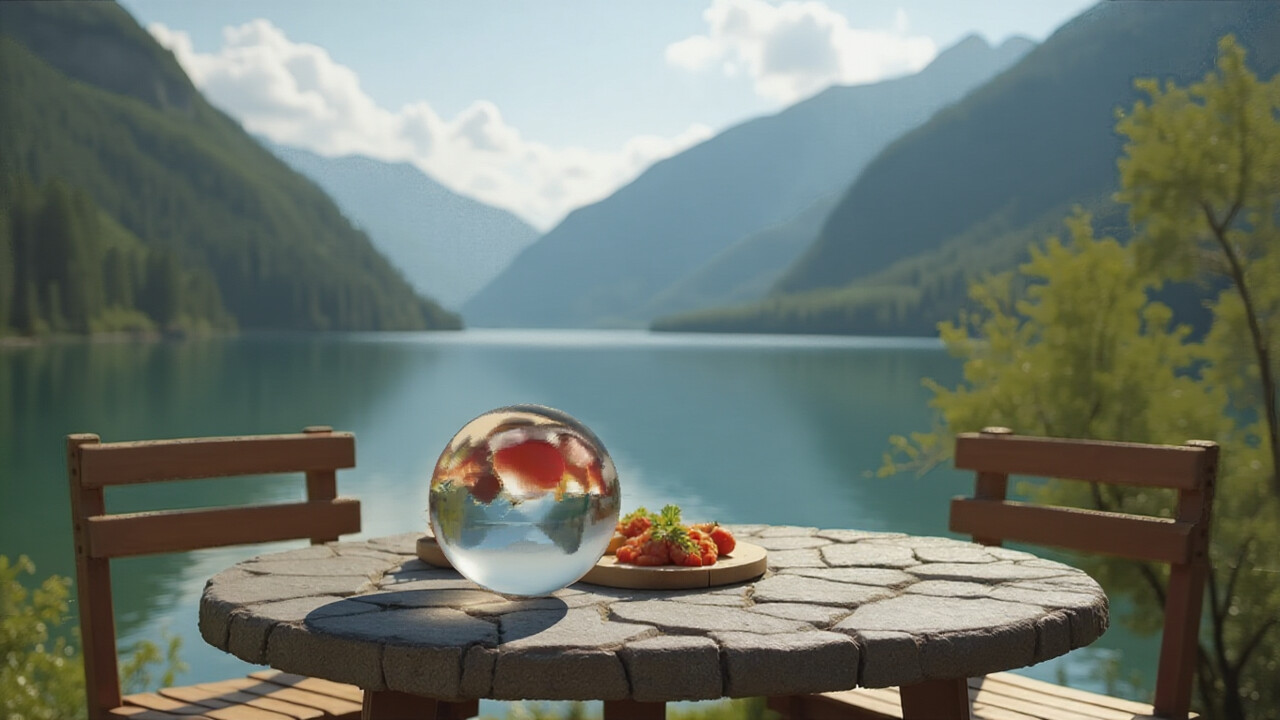}{figures/splash/blender_land.jpg}{428}{116}{660}{396}{428}{116}{660}{396} &
        
        \splashimage{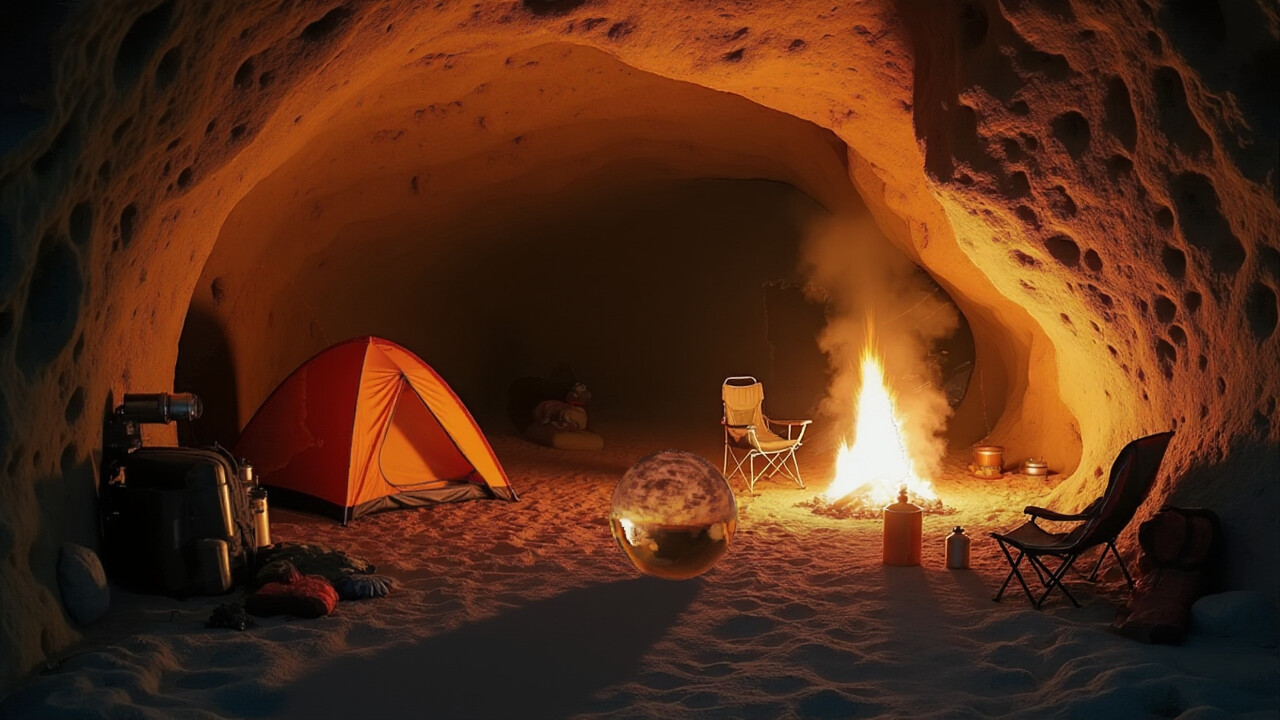}{figures/splash/blender_cave.jpg}{605}{136}{538}{447}{605}{136}{538}{447}
        
    \end{tabularx}
    \vspace{-8pt}
    \setcounter{figure}{0}
    \captionof{figure}{
    Image generation with optically-accurate refractions and consistent reflections.
    (Top)~Images generated by FLUX~\cite{labs2025flux1kontextflowmatching} with masked transparent sphere inpainting, exhibiting physically impossible refractions.
    (Bottom)~Images generated by our method, \method, with refractions and reflections that follow physical laws.
    (Inset)~Close-up of the refractive object as synthesized (left) and rendered (right) with an occlusion mask by Blender from a lifted 3D model. %
    The latter provides approximate ground truth, albeit with incorrect lighting. %
    }
    \label{fig:splash}
}

\makeatletter
\apptocmd{\@maketitle}{\par\vspace{-12pt}\splashfigure\par\vspace{16pt}}{}{}
\makeatother

\title{\raisebox{-0.85em}[0pt][-0.85em]{Refracting \rotatebox[origin=l]{15}{Reality:}} Generating Images with Realistic Transparent Objects} %

\author{Yue Yin
\quad \quad 
Enze Tao
\quad \quad 
Dylan Campbell
\\
The Australian National University\\
{\tt\small \{yue.yin1, enze.tao, dylan.campbell\}@anu.edu.au}
}

\begin{document}
\maketitle

\begin{abstract}
Generative image models can produce convincingly real images, with plausible shapes, textures, layouts and lighting.
However, one domain in which they perform notably poorly is in the synthesis of transparent objects, which exhibit refraction, reflection, absorption and scattering.
Refraction is a particular challenge, because refracted pixel rays often intersect with surfaces observed in other parts of the image, providing a constraint on the color.
It is clear from inspection that generative models have not distilled the laws of optics sufficiently well to accurately render refractive objects.
In this work, we consider the problem of generating images with accurate refraction, given a text prompt.
We synchronize the pixels within the object's boundary with those outside by warping and merging the pixels using Snell's Law of Refraction, at each step of the generation trajectory.
For those surfaces that are not directly observed in the image, but are visible via refraction or reflection, we recover their appearance by synchronizing the image with a second generated image---a panorama centered at the object---using the same warping and merging procedure.
We demonstrate that our approach generates much more optically-plausible images that respect the physical constraints.
\end{abstract}
\section{Introduction}
\label{sec:intro}

\epigraph{
First, there's the room you can see through the glass ---that's just the same as our drawing room, only the things go the other way.
}{\textit{Lewis Carroll}}

Text-to-image generation has become increasingly proficient at synthesizing images that faithfully model many aspects of the real world, while conforming to the text prompt.
In many cases, the illusion that the image was taken by a real camera under a physical imaging process is very convincing, with apparently correct shading, shadows, specularities, and perspective distortion, as well as detailed textures, shapes and plausible configurations.
However, transparent objects, and refraction in particular, are not well modeled, as shown in \cref{fig:splash}.
Unlike reflection, where the indirectly observed surfaces are usually out-of-frame and so can be hallucinated, refraction allows surfaces to be viewed both directly \textit{and} indirectly.
For example, in \cref{fig:splash}, the cushions on the sofa are imaged directly, but they are also visible (with some warping) on the sphere.
This provides a constraint on the color of the sphere pixels, which cannot just be hallucinated.
Nonetheless, generative models typically do just hallucinate this, or merely apply an arbitrary small-scale warping to (inaccurately) mimic refraction;
the underlying optical principles have not been learned.

We remedy this by directly enforcing the optical principles of refraction at every step of the image generation process.
We do so by synchronizing \cite{kim2024synctweedies} corresponding pixels on an imaged refractive object with those outside the object boundary, where correspondences are found using Snell's Law \citep{born2013principles}, the object's geometry and material properties, and the scene's geometry.
In our setup, we assume that we have a text prompt describing a scene containing a refractive object, access to a 3D model of the object (such as from a text-to-3D generator), its material properties (including the refractive indices and absorption properties), and its pose (we use a heuristic to place the object on a horizontal surface near the optical axis).
We further assume that there is only a single refractive object that does not scatter light and has a uniform index of refraction.

However, this synchronization process is not able to handle refracted or reflected pixel rays that intersect with surfaces that are occluded or outside the camera's field-of-view.
These unconstrained generated pixels may then not conform to a realistic completion of the scene observed in the image.
To ensure plausibility, we propose synchronizing the image with another generated image: a panorama centered at the refractive object.
The panorama is warped and merged with the perspective view, including the transparent region, so that the refractions and reflections capture a consistent scene.

Concretely, our proposed method generates an image from a text prompt, estimates its depth map, and then generates a new image, synchronized at each timestep to the refraction-warped original and a concurrently generated panorama centered at the transparent object.
Our contributions are:
\begin{enumerate}
    \item a method for generating images involving a single transparent, refractive object;
    \item a synchronization approach that facilitates constrained generation of corresponding pixels within an image, here used for refracted pixel rays; and
    \item a synchronization approach that facilitates consistent generation of unconstrained, unseen pixels, here used to infer plausible occluded or out-of-frame refracted and reflected pixel rays.
\end{enumerate}

\section{Related Work}
\label{sec:relwork}

While the generation of images with correct refraction has not been widely considered in the literature, other aspects of light transport have been addressed.

\paragraph{Image Generation with Light Transport.}
Image relighting aims to change the illumination conditions of a scene or object while preserving material appearance and geometry. 
Learning-based methods \cite{sunrelighting, zhou2023relightable, nestmeyer2020learning, kim2024switchlight, sengupta2021light, wang2023sunstage} are widely used in image relighting, where a network is trained to predict a harmonized image from the input composite.
Recently, some works \cite{zeng2024dilightnet, kocsis2024lightit, deng2024flashtex, jin2024neural, ren2024relightful, choi2025scribblelight, zhang2025scaling} exploit generative models \cite{ldms, labs2025flux1kontextflowmatching} to improve relighting by manipulating the illumination of the image foreground using information from the background.
For shadow generation, existing methods can be divided into rendering based and non-rendering based approaches.
Rendering-based methods \cite{sheng2021ssn, sheng2023pixht, kee2014exposing, karsch2014automatic, sheng2022controllable} require explicit knowledge of lighting, reflectance, and scene geometry to generate shadows for inserted virtual objects using rendering techniques.
Non-rendering methods add realistic shadows to composite images without inferring 3D information, and are mostly GAN-based \cite{liu2020arshadowgan, zhang2019shadowgan, hu2019mask, hong2022shadow} and diffusion-based \cite{zhao2025shadow, liu2024shadow, tao2024shadow}.
For our task, we do not employ the aforementioned relighting or shadow generation models; instead, we use a finetuned FLUX Kontext image editing model to jointly perform relighting and shadow synthesis as a post-processing step, enhancing the realism of the generated transparent object within our scene.

\paragraph{Image Generation with Reflections.}
Looking more specifically at the case of generative modeling with reflections,
\citet{dhiman2025reflecting} introduce SynMirror, a large-scale dataset of mirror scenes, together with MirrorFusion, a depth-conditioned diffusion inpainting method that generates geometrically consistent and photorealistic mirror reflections.
Their results show that diffusion models can be adapted to faithfully render reflections when augmented with geometry cues.
Similarly, \citet{phongthawee2024diffusionlight} propose DiffusionLight, which leverages diffusion models to insert chrome balls into images for single-image lighting estimation.
By fine-tuning Stable Diffusion XL \citep{sdxl} with LoRA \citep{hu2022lora}, they enable HDR light probe estimation from standard LDR images, achieving convincing illumination across diverse in-the-wild scenes.
Both approaches illustrate how targeted conditioning can significantly improve realism for reflective effects.

\paragraph{Synchronized Generation.}
Numerous studies have explored synchronized generation by sampling synchronously from multiple diffusion paths while maintaining consistency across them.
SyncDiffusion \cite{lee2023syncdiffusion} introduces a joint diffusion synchronization module using a perceptual similarity loss to generate coherent montages.
Visual Anagrams \cite{geng2024visual} synchronises diffusion paths from different viewpoints by averaging the predicted noise at each step.
SyncTweedies \cite{kim2024synctweedies} averages the predicted clean images using Tweedie estimates \cite{robbins1992empirical}.
SyncSDE \cite{lee2025syncsde} proposes a probabilistic framework that explicitly models correlations between diffusion trajectories.
SaFa \cite{dai2025latent} applies latent swapping as a high-performance alternative to averaging, which supports the generation of seamless audio and panoramas.
Closer to the problem of refractive and reflective image generation, \citet{chang2025lookingglass} explore generative frameworks for creating anamorphic images with latent rectified flow models.
Their goal is to produce an image that reveals a second image when a specific mirrored object is placed on a physical copy of the image and is viewed from the correct viewpoint.
Our work builds on this underlying idea of synchronizing warped views of the same image to, in our case, model accurate refraction and reflection.

\section{Preliminaries}
\label{sec:prelim}

In this section, we outline the technical background used in our approach, relating to flow matching and light transport.

\subsection{Flow Matching}
In flow matching \cite{lipman2023flow}, a clean latent $z_0 \sim p_{\mathrm{data}}$ can be transformed into pure noise $z_1 \sim \mathcal{N}(0, I)$ by evolving the latent under an ordinary differential equation (ODE):
\begin{equation}
dz_t = v_\theta(z_t, t, p)\,dt, \quad t \in [0,1],
\end{equation}
where $z_t$ denotes the latent at time $t$, and the velocity $v_\theta$ is typically parameterized by a neural network that can be optionally conditioned on a text prompt $p$.
At inference, samples can be generated by reversing this process, integrating the ODE from $z_1$ back to $z_0$. The ODE can be discretized and solved using classical numerical integration schemes such as forward Euler:
\begin{equation}
z_{t-\Delta t} = z_t + \Delta \sigma \, v_\theta(z_t, t, p),
\end{equation}
where $\Delta\sigma = \sigma_{t-\Delta t} - \sigma_t$ and $\sigma_t$ denotes the noise schedule at time $t$, with $\sigma_0 = 0$ (clean) and $\sigma_1 = 1$ (pure noise).
Optionally, stochasticity can be added by turning the backward process into a stochastic differential equation (SDE):
\begin{equation}
z_{t-\Delta t} = z_t + \Delta \sigma \, v_\theta(z_t, t, p) + |\sigma_t - \sigma_{t-\Delta t}| \, \epsilon,
\end{equation}
where $\epsilon \sim \mathcal{N}(0, I)$ is Gaussian noise.

\paragraph{Classifier-free guidance (CFG).}
To enhance the quality of generated samples, classifier-free guidance is usually used.
The guided velocity is obtained by linearly combining the conditional and unconditional predictions:
\begin{equation}
\hat{v}_t = (1 + \omega) v_\theta(z_t, t, p) - \omega v_\theta(z_t, t, \emptyset),
\end{equation}
where $\omega$ denotes the guidance scale.
Increasing $\omega$ amplifies the influence of the conditioning prompt, typically improving visual consistency while reducing diversity and occasionally causing over-saturated outputs.

\paragraph{Euler estimates.}
At any intermediate timestep $t$ with noise level $\sigma_t$, an estimate of the clean latent $z_0$ can be obtained by taking a single Euler step using the current velocity prediction $v_\theta(z_t, t, p)$:
\begin{equation}
\label{eq:euler}
z_{0|t} = z_t - \sigma_t v_\theta(z_t, t, p).
\end{equation}
This equation can be seen as the flow matching equivalent of Tweedie's formula \cite{robbins1992empirical} in diffusion models.

\subsection{Light Transport: Refraction and Reflection}%
\label{sec:prelim_rays}

In this section, we outline the equations governing refraction and reflection, as well as our ray casting procedure used to compute the pixel--pixel mappings.
Since we are considering materials with piecewise constant refractive indices, we represent light paths as piecewise linear functions.
They are parameterized by $N$ points $\{\rvx_i\}_{i=0}^{N-1}$ and unit direction vectors $\{\rvd_i\}_{i=0}^{N-1}$,
\begin{equation}
\label{eq:piecewise_linear}
\rvr(t) = \sum_{i = 0}^{N-1} \mathbf{1}_{[\tau_i, \tau_{i+1})}(t) \left( \rvx_i + (t - \tau_i) \rvd_i \right),
\end{equation}
where $\mathbf{1}_{A}(t)$ is an indicator function testing $t\in A$, 
the initial pair $(\rvx_0, \rvd_0)$ is given by the camera origin and pixel direction vector, and the cumulative distance is given by
$\tau_i = \sum_{j=1}^{i} \|\rvx_j - \rvx_{j-1}\|$, excepting for $\tau_0=0$ and $\tau_N=\infty$.

We compute a refraction path $\rvr^\refr$, which may include multiple refractions and total internal reflections, and a reflection path $\rvr^\refl$, which only considers the dominant first reflection, since secondary reflections usually have negligible impact.
For the refraction path $\rvr^\refr$, we cast a ray from position $\rvx_i$ in direction $\rvd_i$ until it hits a surface of the refractive object at point $\rvx_{i+1}$.
Then the next direction $\rvd_{i+1}$ is computed using Snell's Law \citep{born2013principles},
\begin{equation}
\label{eq:snell}
    \rvd_{i+1} = \alpha_i\rvd_{i} + \left(\alpha_i \beta_i - \sqrt{\gamma_i} \right)\rvn(\rvx_{i+1}),
\end{equation}
where 
$\alpha_i = \nu_i / \nu_{i+1}$,
$\beta_i = -\rvd_{i}\transpose\rvn(\rvx_{i+1})$,
$\gamma_i = 1 - \alpha_i^2 (1 - \beta_i^2 )$,
$\nu_i$ is the refractive index of the $i$\textsuperscript{th} material, and
$\rvn(\rvx)$ is the unit surface normal at point $\rvx$.
If $\gamma_i < 0$, total internal reflection occurs instead and the direction is given by the Law of Reflection,
\begin{equation}
    \label{eq:reflection}
    \rvd_{i+1} = \rvd_{i} - 2(\rvd_{i}\transpose \rvn(\rvx_{i+1})) \rvn(\rvx_{i+1}). 
\end{equation}
For the reflection path $\rvr^\refl$, the first reflected direction $\rvd_1^\refl$ is given by \cref{eq:reflection}.

\begin{figure*}[!t]
    \centering
    \hspace*{-1.25em}
    \resizebox{\linewidth}{!}{\input{figures/pipeline.tikz}\unskip}%
    \vspace{-6pt}
    \caption{
        \method flowchart.
        (Top)~An initial image $I_0^-$ is generated using Flux~\cite{labs2025flux1kontextflowmatching}.
        Its depth map $D^-$ is estimated (not shown) and is used to place the transparent object on a horizontal surface near the optical axis.
        Rays cast through this object are refracted according to Snell's Law \citep{born2013principles} and intersect with the estimated geometry, defining the warping functions.
        These are used to synthesize an image with correct refraction for the surfaces visible in the original image.
        (Middle)~In the main branch, we generate the perspective image from prompt $p$, including the transparent object.
        (Bottom)~In the auxiliary branch, we concurrently generate a panoramic image from augmented prompt $p^{360}$, centered at the transparent object's location, to consistently fill in occluded or out-of-frame surfaces.
        (All)~At each denoising step $t$, we compute Euler estimates of the clean images for both branches $I_{0|t}$ and warp them using the precomputed geometric correspondences.
        These are blended with the warped original image to obtain a complete perspective and panoramic image.
        Finally, we combine refractive and reflective contributions using Fresnel's equations, before encoding back into latent space for the next denoising step.
    }
    \label{fig:pipe}
\end{figure*}

\section{Transparent Object Image Generation}
\label{sec:method}

Our training-free image generation method, which we name \method since it casts rays that are curved according to Snell's Law \citep{born2013principles}, is visualized in \cref{fig:pipe}.
First, an image $I_0^-$ is generated given a text prompt $p^-$, which has any reference to the transparent object removed, and its depth map $D^-$ is estimated using MoGe-2 \cite{wang2025moge}.
We then generate two images concurrently: a perspective image guided by prompt $p$, which includes the transparent object, and a panoramic image guided by augmented prompt $p^{360}$, centered at the transparent object's location.
At each denoising timestep $t$, we compute Euler estimates $I_{0|t}$ and $I_{0|t}^{360}$ using \cref{eq:euler} for both views, warp them using precomputed geometric correspondences from ray tracing, and blend them together to enforce physical consistency.
Finally, we combine refracted and reflected contributions using the Fresnel equations to produce the final image, which is encoded back into latent space for the next denoising step.

\subsection{Computing the Warping Functions}
Before generation begins, we pre-compute pixel-to-pixel warping functions by ray tracing through the 3D scene geometry, as outlined in \cref{sec:prelim_rays}.
We first convert the depth map $D^-$ into a 3D mesh and place the transparent object mesh onto a horizontal surface near the optical axis.

\paragraph{Self-warping.}
To model refraction through the transparent object, we cast rays from the perspective camera through each pixel, tracing their paths through the foreground mesh according to Snell's Law~\cite{born2013principles} and intersecting with the background mesh.
Projecting these intersections back to the image plane yields $\pi^\refr$, which maps pixels inside the transparent object to their corresponding background locations in $I_0^-$, as caused by refraction.

\paragraph{Cross-view Warping.}
The perspective camera does not observe occluded or out-of-view surfaces.
We therefore generate a panoramic view centered at the transparent object, which approximates the surrounding scene as viewed from each point on its surface.
To get the panorama-to-perspective warping, rays are cast from the perspective camera to refract through the foreground mesh and intersect with the background mesh, including its bounding box.
The intersection directions are then projected onto the panorama to yield $\pi^{-1}$, the refraction component for panorama-to-perspective warping, with the refracted color $\rvc^\refr$ at each pixel sampled from the corresponding intersection direction.
Similarly, tracing reflected ray paths from the perspective camera and projecting them onto the panorama yields $\pi^{-\refl}$, the reflection component for panorama-to-perspective warping, with the reflected color $\rvc^\refl$ sampled along the reflected direction $\rvd_1^\refl$ on the equirectangular panorama.
Conversely, casting straight rays from the panoramic camera and projecting their directions onto the perspective image plane yields $\pi$, which warps the perspective view to the panoramic view.
These warping functions are computed once and reused throughout the generation process.

\subsection{Cross-view Synchronization}
\label{sec:sync}
At each denoising timestep $t$, we synchronize the perspective and panoramic views to ensure appearance and geometry consistency.
We obtain Euler estimates for both views using \cref{eq:euler}, and decode the latent states to $I_{0|t}$ and $I_{0|t}^{360}$ in pixel space for the warping and blending, due to the fact that warping in latent space does not correspond to the same geometric transformation in pixel space \cite{chang2025lookingglass}.

\paragraph{Synthesizing Physics-Based Refractions.}
\label{para:refraction}
To incorporate the known background geometry, we apply the self-warping function $\pi^\refr$ to the clean object-free image $I_0^-$ to obtain $\pi^{\refr}(I_0^-)$, which shows the background as it would appear through the transparent object.
Similarly, we warp the panoramic estimate to the perspective view and get $\pi^{-1}(I_{0|t}^{360})$, incorporating scene content invisible from the perspective camera.
We then blend the three components in the perspective view together,
\begin{align}
\hat{I}_{0|t}^{\refr} &= \phi ((I_{0|t}, I_{0|t}^{360}, I_0^-), (\pi^\mathbb{I}, \pi^{-1}, \pi^{\refr}), \lambda ),
\end{align}
where $\pi^\mathbb{I}$ is the identity warp and $\phi$ denotes an occlusion-masked blending operation with value-weighted averaging that preserves fine details of the images \cite{chang2025lookingglass}, given by
\begin{align}
\phi(\mathcal{X}, \mathcal{Y}, \lambda) =& (1-\lambda) \frac{\sum_i M(\mathcal{Y}_i) \odot \mathcal{Y}_i(\mathcal{X}_i)}{\sum_i M(\mathcal{Y}_i)} \notag\\
&+ \lambda \frac{\sum_i M(\mathcal{Y}_i) \odot |\mathcal{Y}_i(\mathcal{X}_i)| \odot \mathcal{Y}_i(\mathcal{X}_i)}{\sum_i M(\mathcal{Y}_i) \odot |\mathcal{Y}_i(\mathcal{X}_i)|},
\end{align}
where $M(\mathcal{Y}_i)$ is the occlusion mask associated with the $i$th element of the sequence of warping functions $\mathcal{Y}$, and the divisions are element-wise.
For the panoramic view, we perform an analogous synchronization procedure:
\begin{align}
\hat{I}_{0|t}^{360} &= \phi \big( (I_{0|t}^{360}, I_{0|t}, I_0^-), (\pi^\mathbb{I}, \pi, \pi), \lambda \big).
\end{align}

\paragraph{Synthesizing Plausible Reflections.}
\label{para:reflection}
Transparent objects exhibit both refraction and reflection, with their relative contributions determined by the viewing angle.
Using the reflection warping $\pi^{-\refl}$ we warp the panoramic estimate to obtain the reflected appearance $I_{0|t}^\refl = \pi^{-\refl}(I_{0|t}^{360})$.
The refractive and reflective color contributions are then combined using the Fresnel equations~\cite{hecht2012optics},
\begin{align}
    \rvc' &= \frac{1}{2}(R_p + R_s) (\rvc^\refl - \rvc^\refr) + \rvc^\refr, \label{eq:fresnel}\\
    R_p &=\! \left( \frac{\nu_1 \beta_0 - \nu_0 \sqrt{\gamma_0}}{\nu_1 \beta_0 + \nu_0 \sqrt{\gamma_0}} \right)^{\!\!2} \!\!, 
    R_s \!=\! \left( \frac{\nu_0 \beta_0 - \nu_1 \sqrt{\gamma_0}}{\nu_0 \beta_0 + \nu_1 \sqrt{\gamma_0}} \right)^{\!\!2} \!\!,
\end{align}
where $R_p$ and $R_s$ are the reflection coefficients for parallel and perpendicular polarized light.
The blending is performed in linear color space, with the result $\rvc'$ converted to sRGB to obtain the final pixel color $\rvc$~\cite{verbin2022ref}.

\paragraph{Updated Denoising Step.}
These final synchronized Euler estimates $\hat{I}_{0|t}$ and $\hat{I}_{0|t}^{360}$ are then encoded back to latent space. 
The resulting latents $\hat{z}_{0|t}$ and $\hat{z}_{0|t}^{360}$ are used to guide the subsequent denoising step:
\begin{equation}
z_{t-1} = z_t + \frac{\sigma_{t-1} - \sigma_t}{\sigma_t} (z_t - \hat{z}_{0|t}),
\label{eq:step}
\end{equation}
ensuring that both views remain consistent.

\paragraph{Further Details and Post-processing.}
All warping operations are performed using Laplacian pyramid warping, as in \cite{chang2025lookingglass}.
This reduces boundary artifacts and aliasing, since pixel data is sourced from the image with the appropriate resolution for the level of expansion or contraction caused by the warp.
Like FreeDoM \cite{yu2023freedom} and LookingGlass \cite{chang2025lookingglass}, we use time travel \cite{wangzero} to improve multi-view consistency.
All details are in the supplement.
After the entire generation process is complete, we apply a foreground object relighting (harmonization) model \cite{labs2025flux1kontextflowmatching} as a post-processing step to add shadows and specularities conforming to the light sources in the rest of the image.

\begin{figure*}[!t]
    \centering
    \makebox[0.025\linewidth]{} %
    \begin{minipage}[]{0.237\linewidth}\centering
        Blender Reference
    \end{minipage}\hspace{2pt}%
    \begin{minipage}[]{0.237\linewidth}\centering
        Snellcaster (Ours)
    \end{minipage}\hspace{2pt}%
    \begin{minipage}[]{0.237\linewidth}\centering
        FLUX Inpainting
    \end{minipage}\hspace{2pt}%
    \begin{minipage}[]{0.237\linewidth}\centering
        FLUX-dev
    \end{minipage}\\[1.0pt]

    \begin{minipage}[c]{0.025\linewidth}\centering
        \rotatebox{90}{Artroom (Inset)}
    \end{minipage}%
    \begin{subfigure}[]{0.237\linewidth}\centering
        \includegraphics[width=\linewidth,
            trim={500pt 255pt 475pt 280pt}, clip]{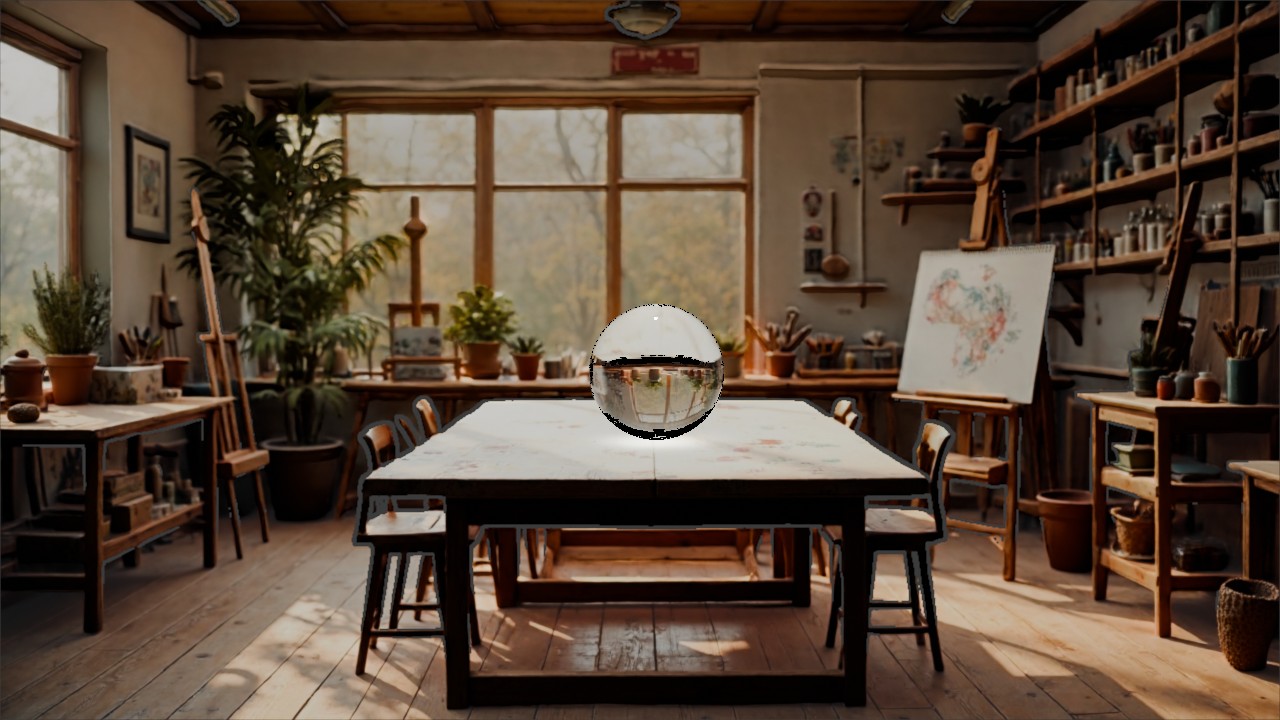}
    \end{subfigure}\hspace{2pt}%
    \begin{subfigure}[]{0.237\linewidth}\centering
        \includegraphics[width=\linewidth,
            trim={500pt 255pt 475pt 280pt}, clip]{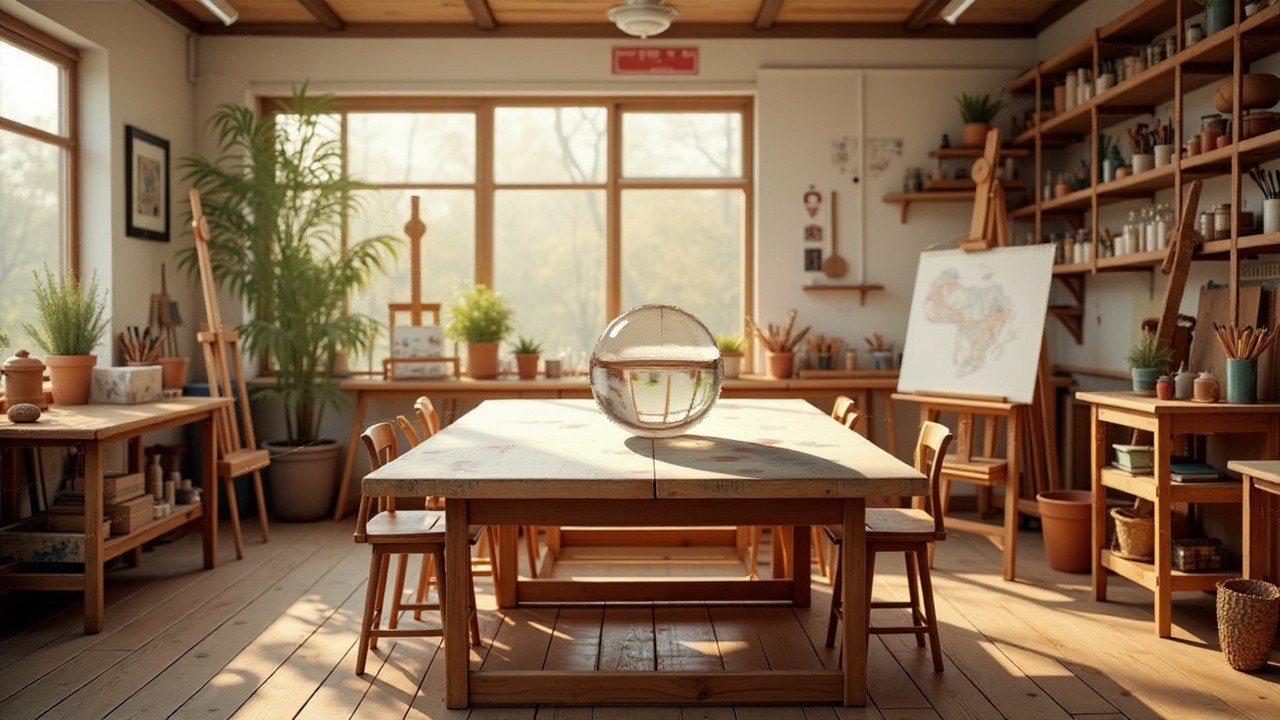}
    \end{subfigure}\hspace{2pt}%
    \begin{subfigure}[]{0.237\linewidth}\centering
        \includegraphics[width=\linewidth,
            trim={500pt 255pt 475pt 280pt}, clip]{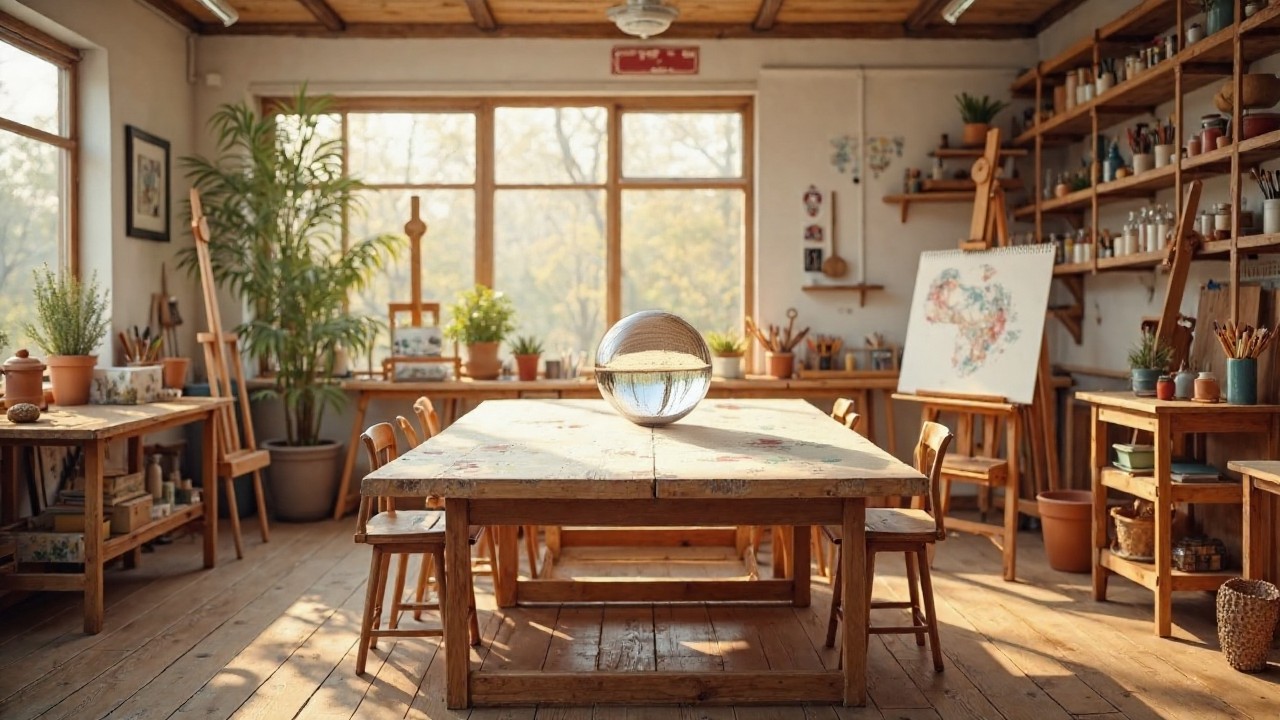}
    \end{subfigure}\hspace{2pt}%
    \begin{subfigure}[]{0.237\linewidth}\centering
        \includegraphics[width=\linewidth,
            trim={500pt 255pt 475pt 280pt}, clip]{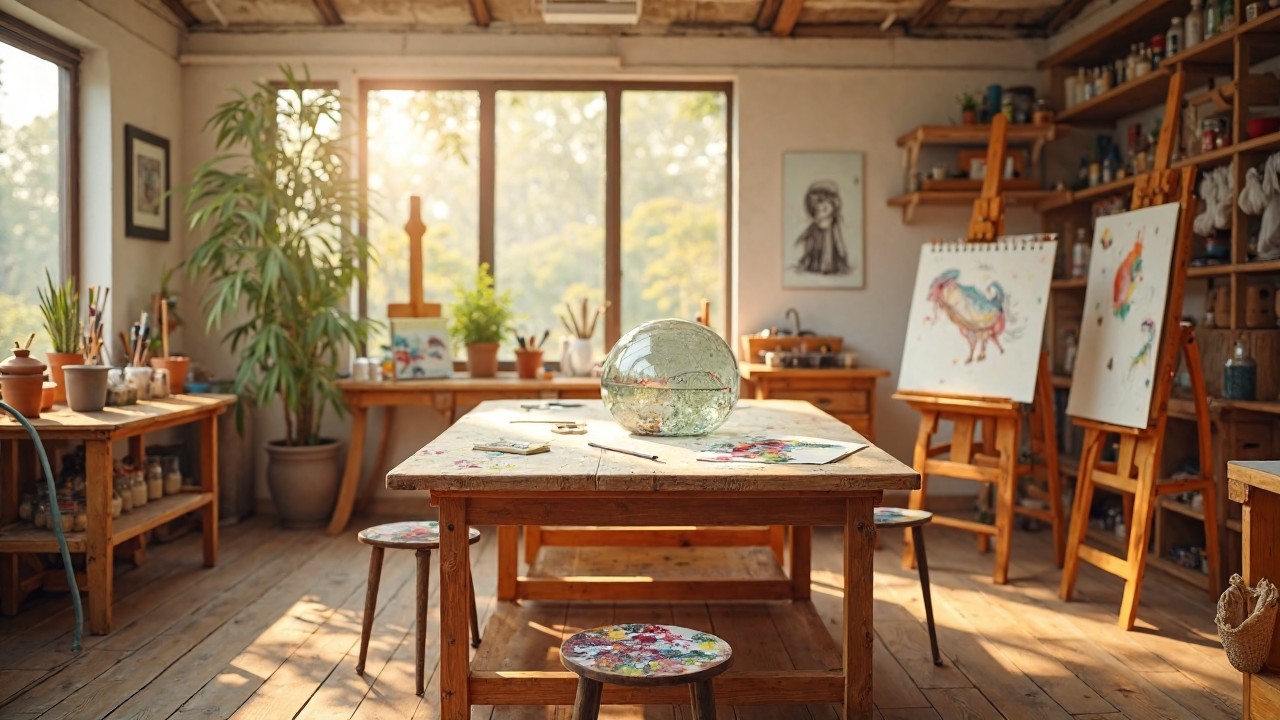}
    \end{subfigure}\\[1.0pt]

    \begin{minipage}[c]{0.025\linewidth}\centering
        \rotatebox{90}{Artroom}
    \end{minipage}%
    \begin{subfigure}[]{0.237\linewidth}\centering
        \includegraphics[width=\linewidth]{figures/results/artroom_maskedgt.jpg}
    \end{subfigure}\hspace{2pt}%
    \begin{subfigure}[]{0.237\linewidth}\centering
        \includegraphics[width=\linewidth]{figures/results/artroom.jpg}
    \end{subfigure}\hspace{2pt}%
    \begin{subfigure}[]{0.237\linewidth}\centering
        \includegraphics[width=\linewidth]{figures/results/artroom_flux_fill.jpg}
    \end{subfigure}\hspace{2pt}%
    \begin{subfigure}[]{0.237\linewidth}\centering
        \includegraphics[width=\linewidth]{figures/results/artroom_flux.jpg}
    \end{subfigure}\\[1.0pt]

    \begin{minipage}[c]{0.025\linewidth}\centering
        \rotatebox{90}{Karaoke}
    \end{minipage}%
    \begin{subfigure}[]{0.237\linewidth}\centering
        \includegraphics[width=\linewidth,
            trim={128pt 77pt 128pt 77pt}, clip]{figures/splash/blender_karaoke.jpg}
    \end{subfigure}\hspace{2pt}%
    \begin{subfigure}[]{0.237\linewidth}\centering
        \includegraphics[width=\linewidth,
            trim={128pt 77pt 128pt 77pt}, clip]{figures/splash/karaoke_main.jpg}
    \end{subfigure}\hspace{2pt}%
    \begin{subfigure}[]{0.237\linewidth}\centering
        \includegraphics[width=\linewidth,
            trim={128pt 77pt 128pt 77pt}, clip]{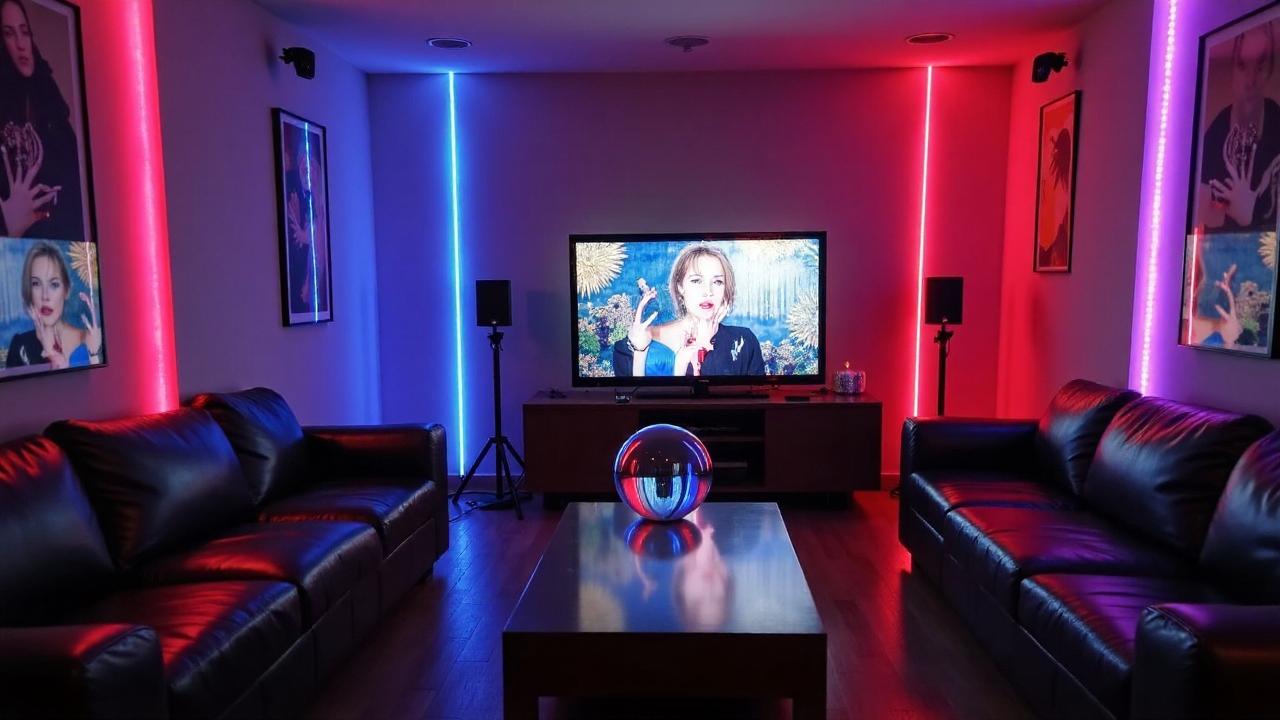}
    \end{subfigure}\hspace{2pt}%
    \begin{subfigure}[]{0.237\linewidth}\centering
        \includegraphics[width=\linewidth]{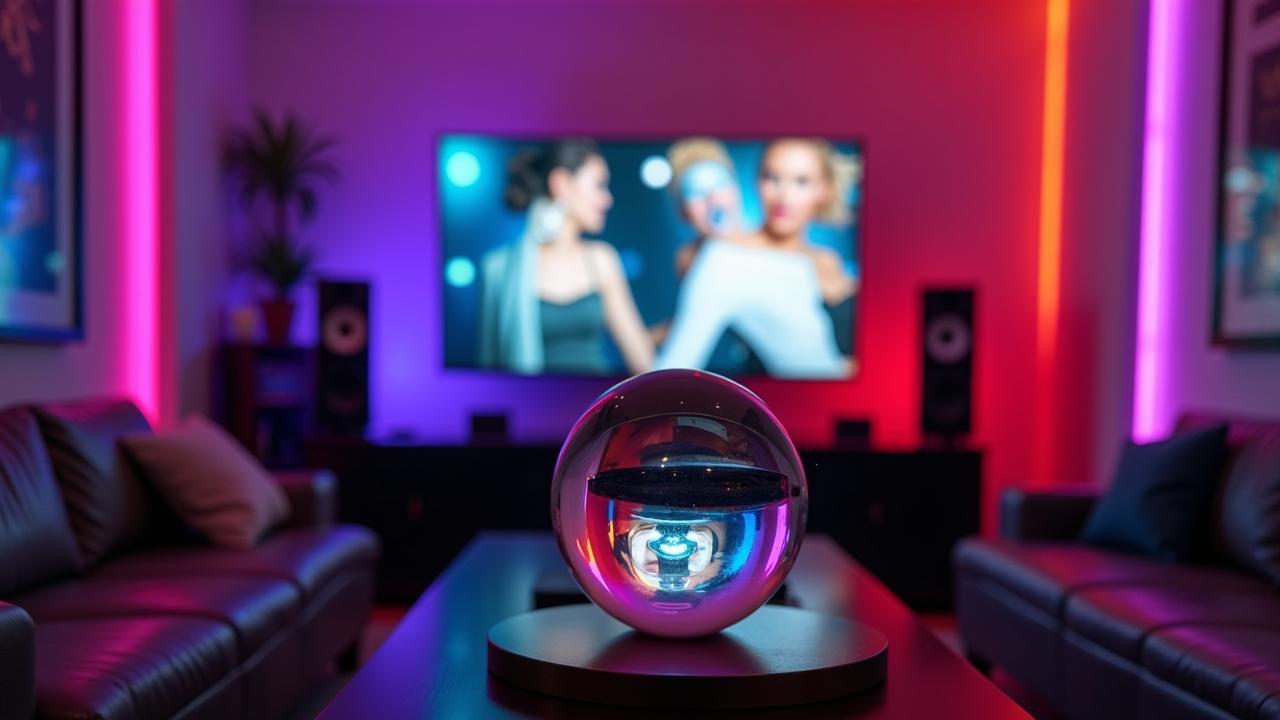}
    \end{subfigure}\\[1.0pt]

    \begin{minipage}[c]{0.025\linewidth}\centering
        \rotatebox{90}{Landscape}
    \end{minipage}%
    \begin{subfigure}[]{0.237\linewidth}\centering
        \includegraphics[width=\linewidth]{figures/splash/blender_land.jpg}
    \end{subfigure}\hspace{2pt}%
    \begin{subfigure}[]{0.237\linewidth}\centering
        \includegraphics[width=\linewidth]{figures/splash/landscape_main.jpg}
    \end{subfigure}\hspace{2pt}%
    \begin{subfigure}[]{0.237\linewidth}\centering
        \includegraphics[width=\linewidth]{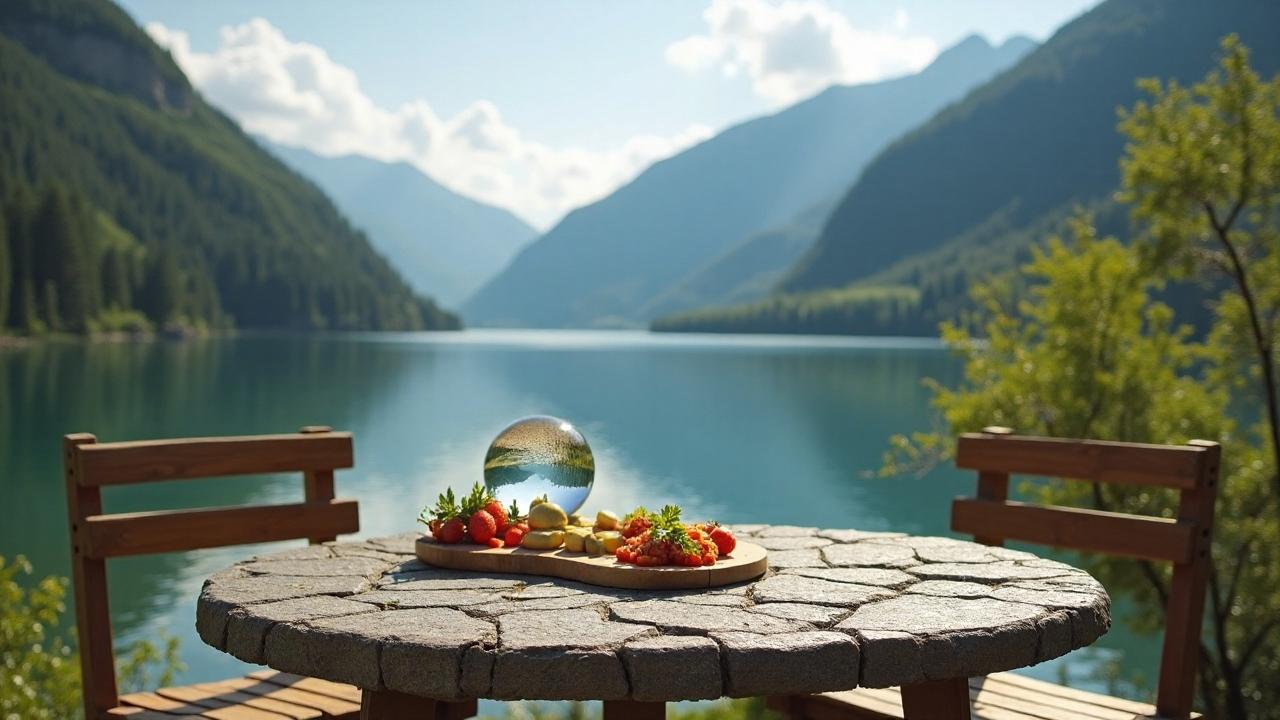}
    \end{subfigure}\hspace{2pt}%
    \begin{subfigure}[]{0.237\linewidth}\centering
        \includegraphics[width=\linewidth]{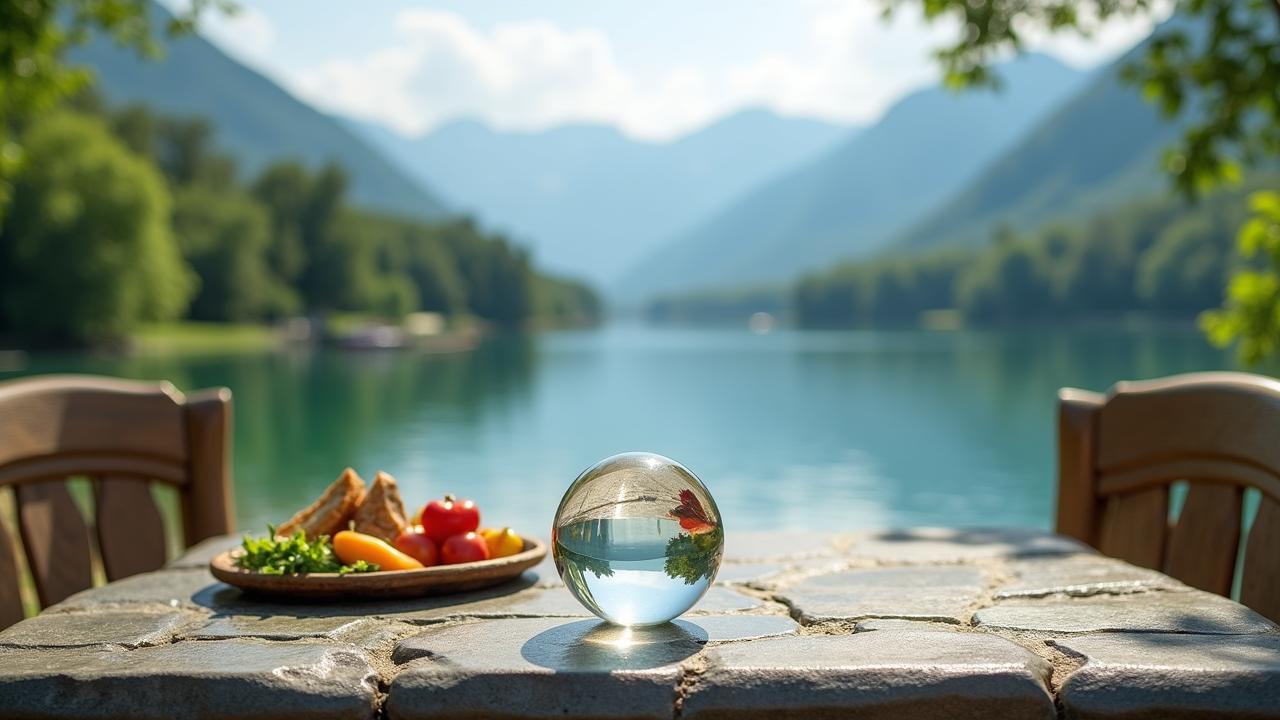}
    \end{subfigure}\\[1.0pt]
    
    \begin{minipage}[c]{0.025\linewidth}\centering
        \rotatebox{90}{Living Room}
    \end{minipage}%
    \begin{subfigure}[]{0.237\linewidth}\centering
        \includegraphics[width=\linewidth]{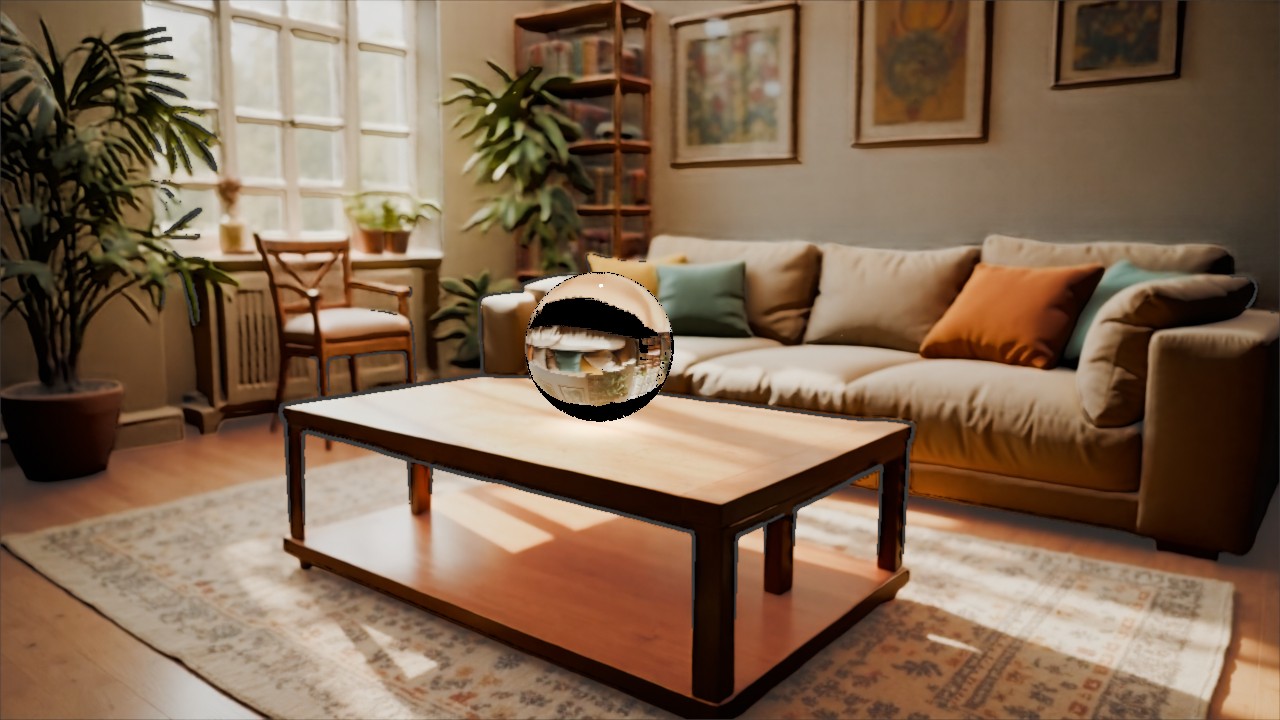}
    \end{subfigure}\hspace{2pt}%
    \begin{subfigure}[]{0.237\linewidth}\centering
        \includegraphics[width=\linewidth]{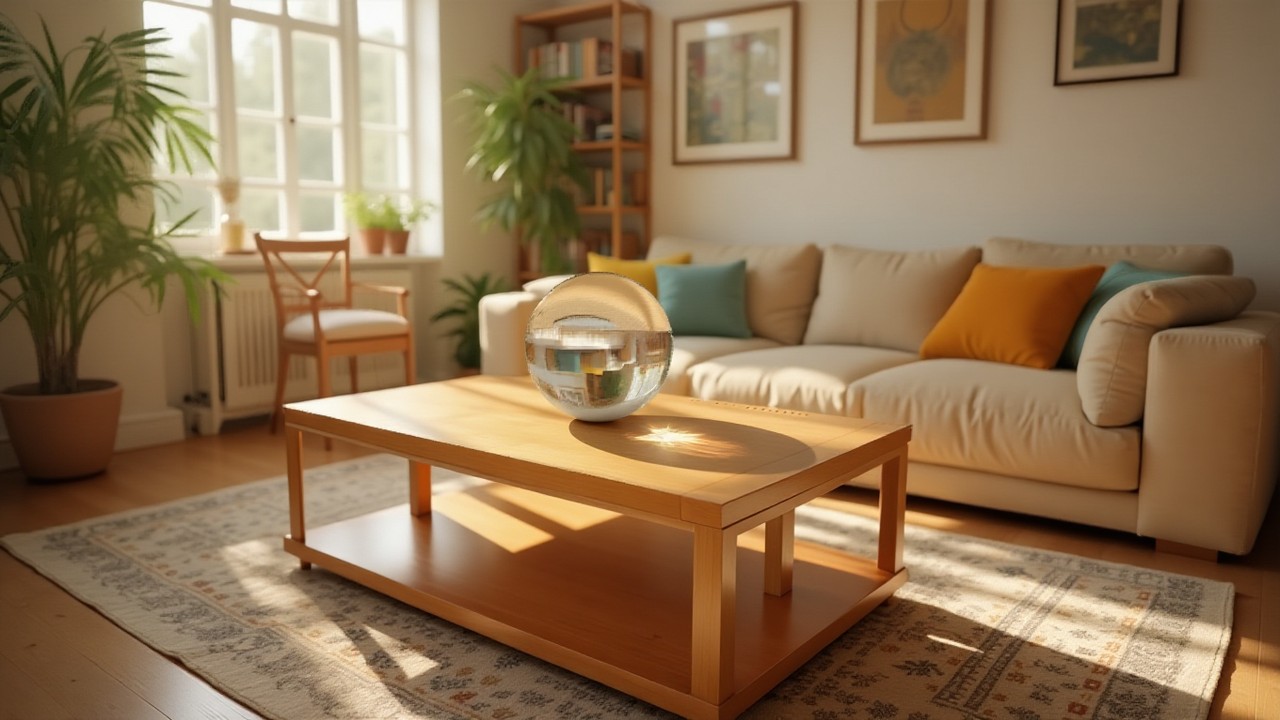}
    \end{subfigure}\hspace{2pt}%
    \begin{subfigure}[]{0.237\linewidth}\centering
        \includegraphics[width=\linewidth]{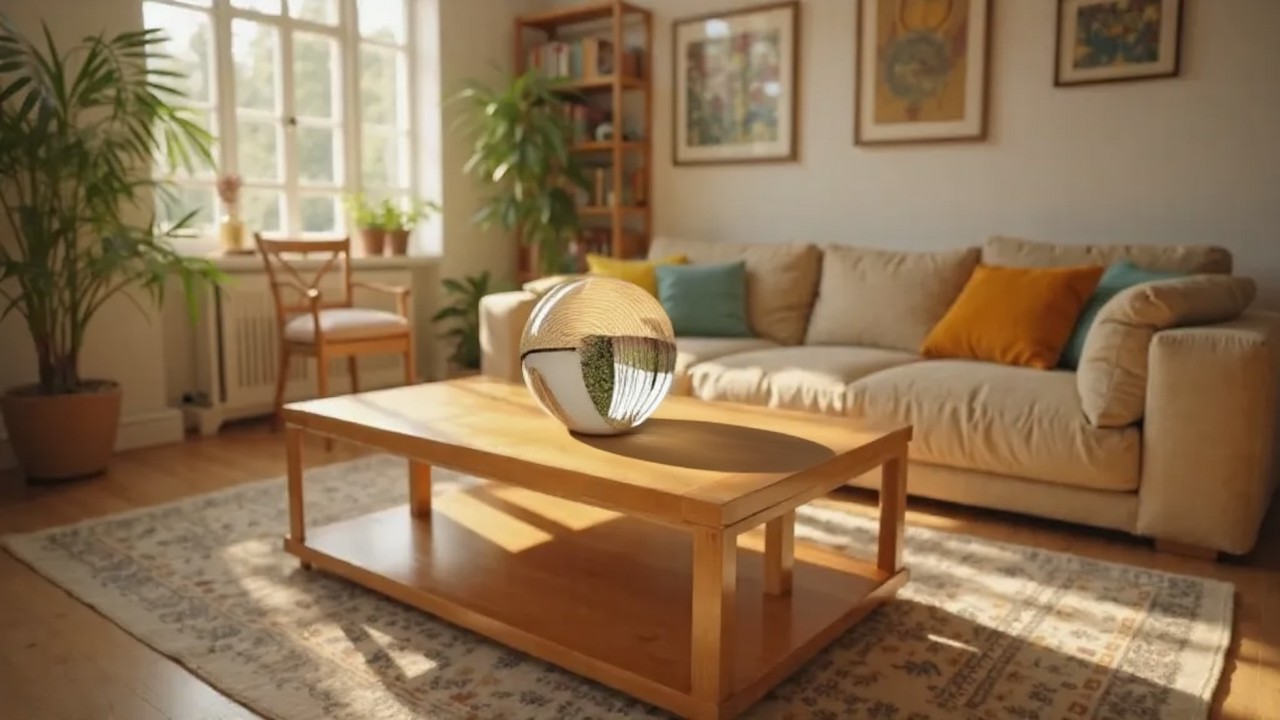}
    \end{subfigure}\hspace{2pt}%
    \begin{subfigure}[]{0.237\linewidth}\centering
        \includegraphics[width=\linewidth]{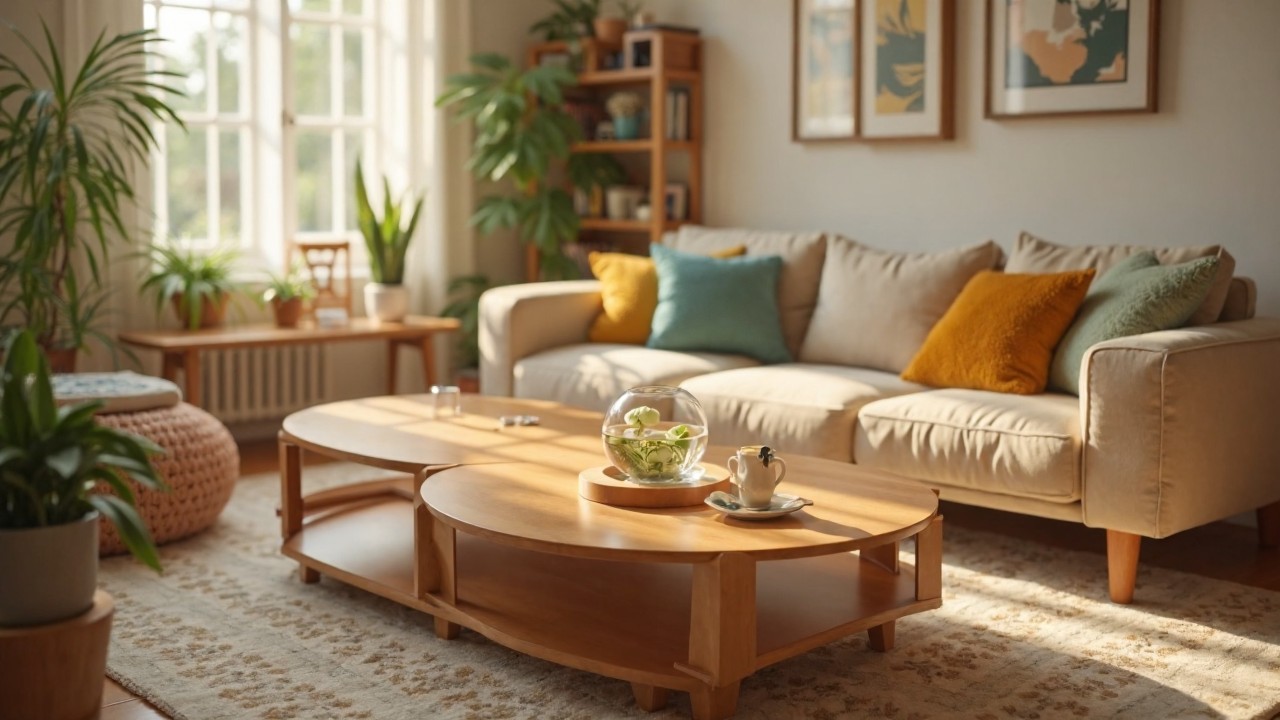}
    \end{subfigure}\\[1.0pt]

    \begin{minipage}[c]{0.025\linewidth}\centering
        \rotatebox{90}{Cave (cylinder)}
    \end{minipage}%
    \begin{subfigure}[]{0.237\linewidth}\centering
        \includegraphics[width=\linewidth,
            trim={325pt 72pt 315pt 288pt}, clip]{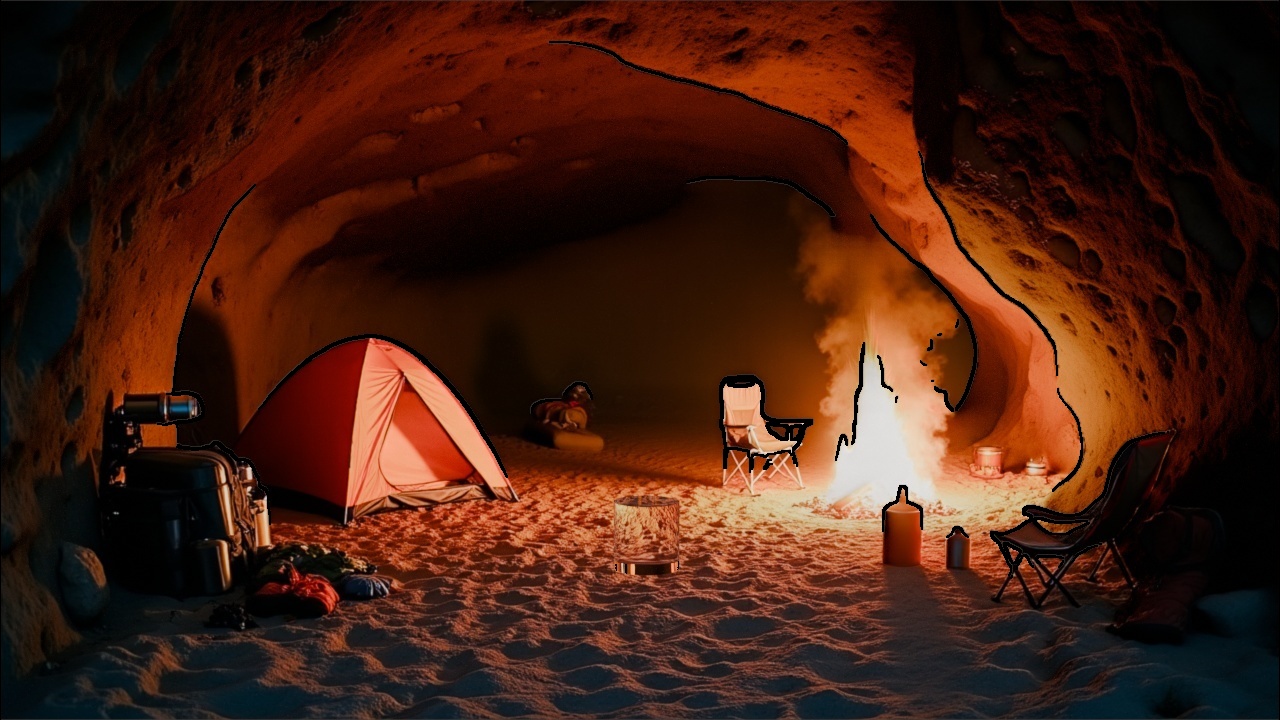}
    \end{subfigure}\hspace{2pt}%
    \begin{subfigure}[]{0.237\linewidth}\centering
        \includegraphics[width=\linewidth,
            trim={325pt 72pt 315pt 288pt}, clip]{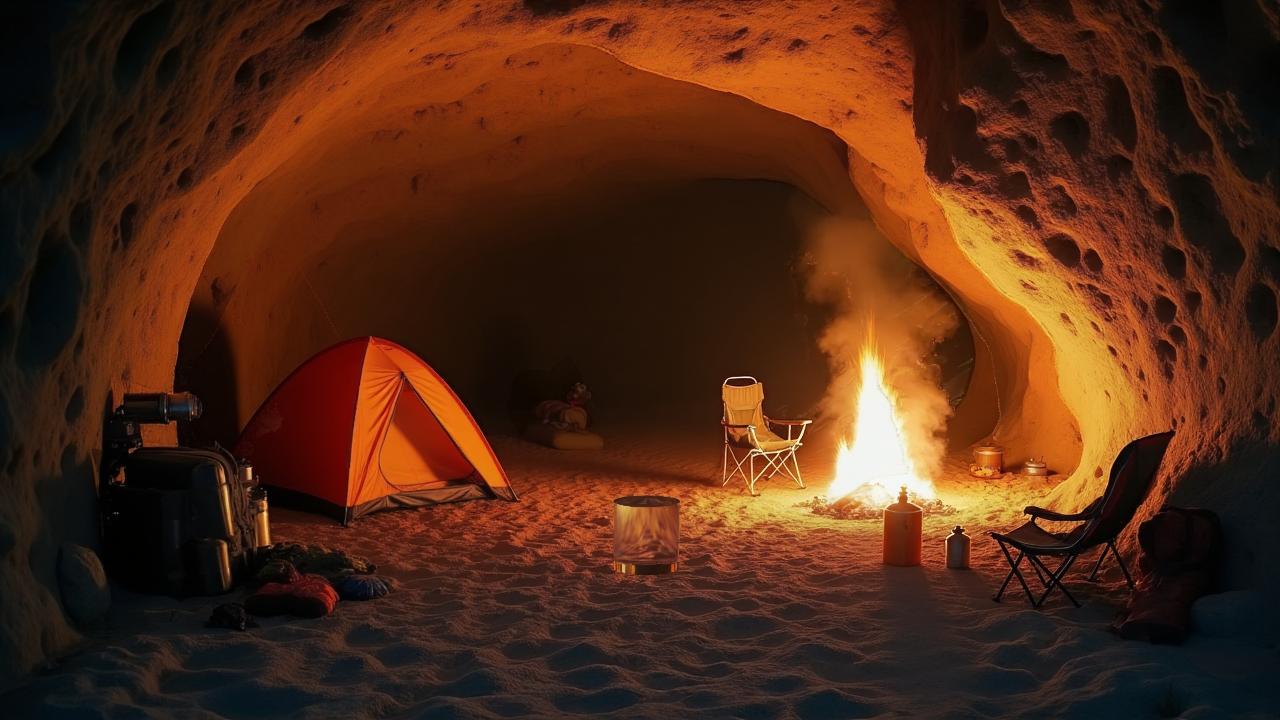}
    \end{subfigure}\hspace{2pt}%
    \begin{subfigure}[]{0.237\linewidth}\centering
        \includegraphics[width=\linewidth,
            trim={325pt 72pt 315pt 288pt}, clip]{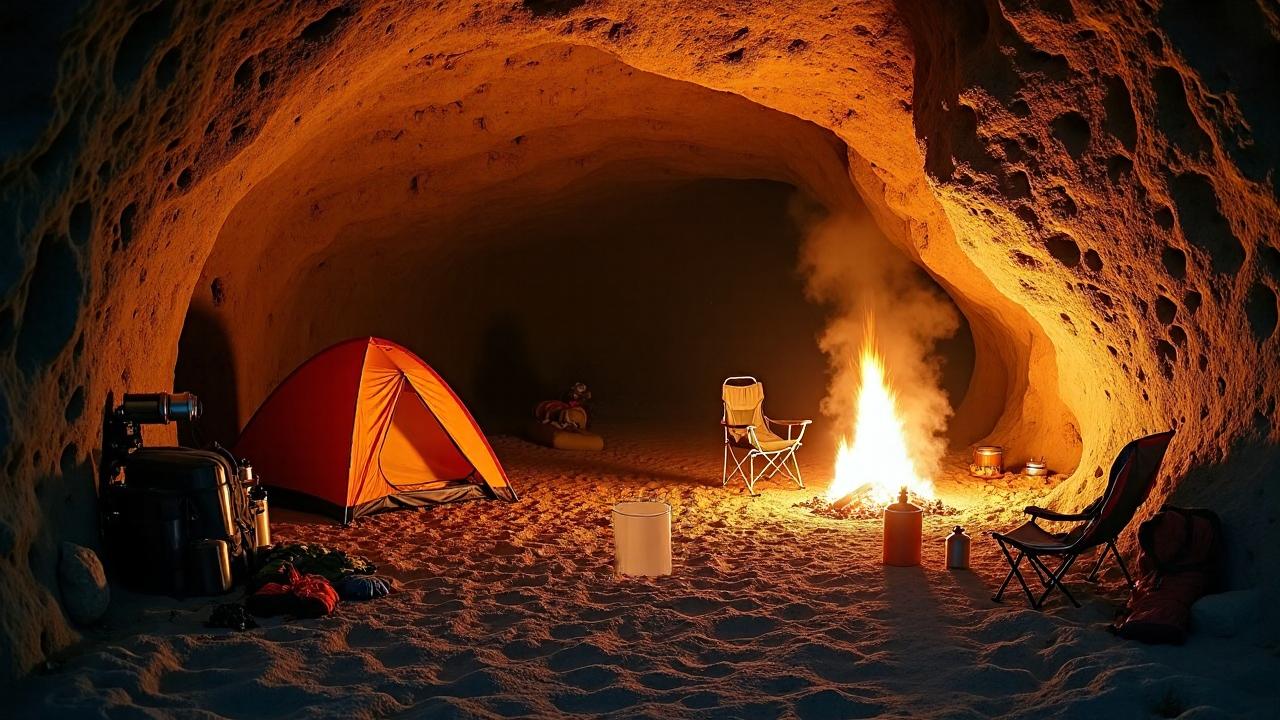}
    \end{subfigure}\hspace{2pt}%
    \begin{subfigure}[]{0.237\linewidth}\centering
        \includegraphics[width=\linewidth,
            trim={325pt 72pt 315pt 288pt}, clip]{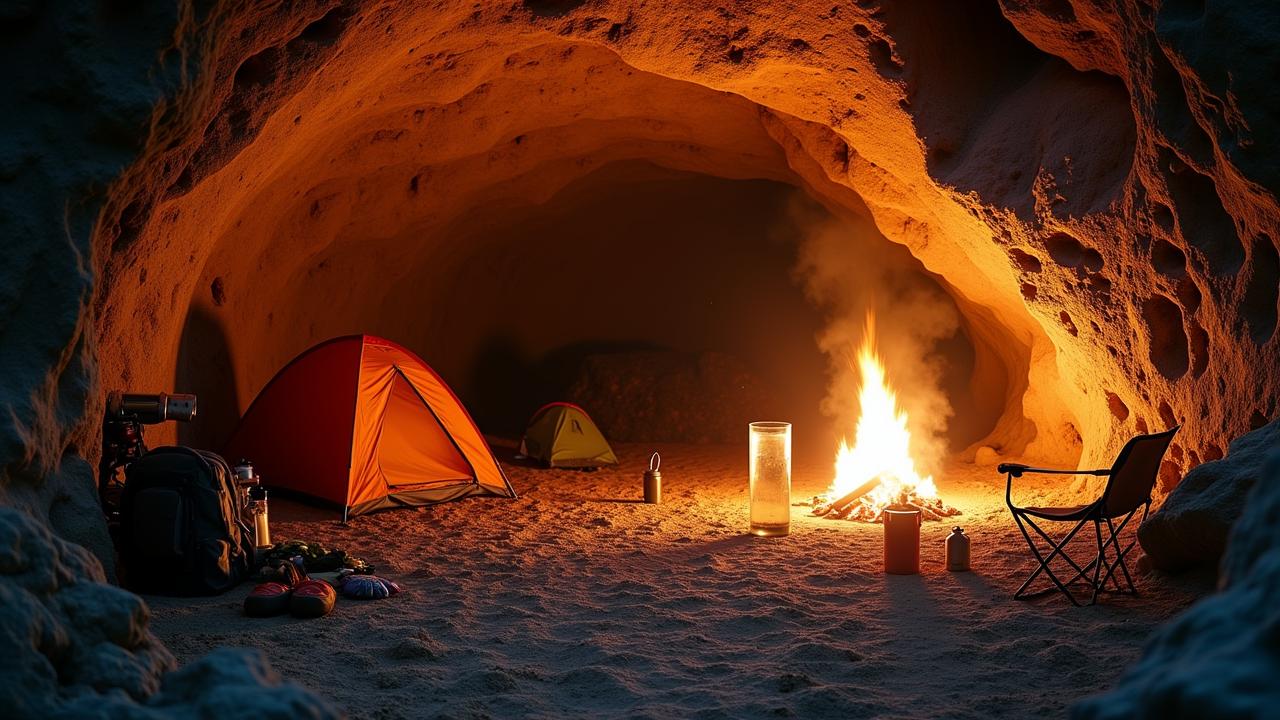}
    \end{subfigure}\\[1.0pt]

    \begin{minipage}[c]{0.025\linewidth}\centering
        \rotatebox{90}{Artroom (dog)}
    \end{minipage}%
    \begin{subfigure}[]{0.237\linewidth}\centering
        \includegraphics[width=\linewidth,
            trim={256pt 144pt 256pt 144pt}, clip]{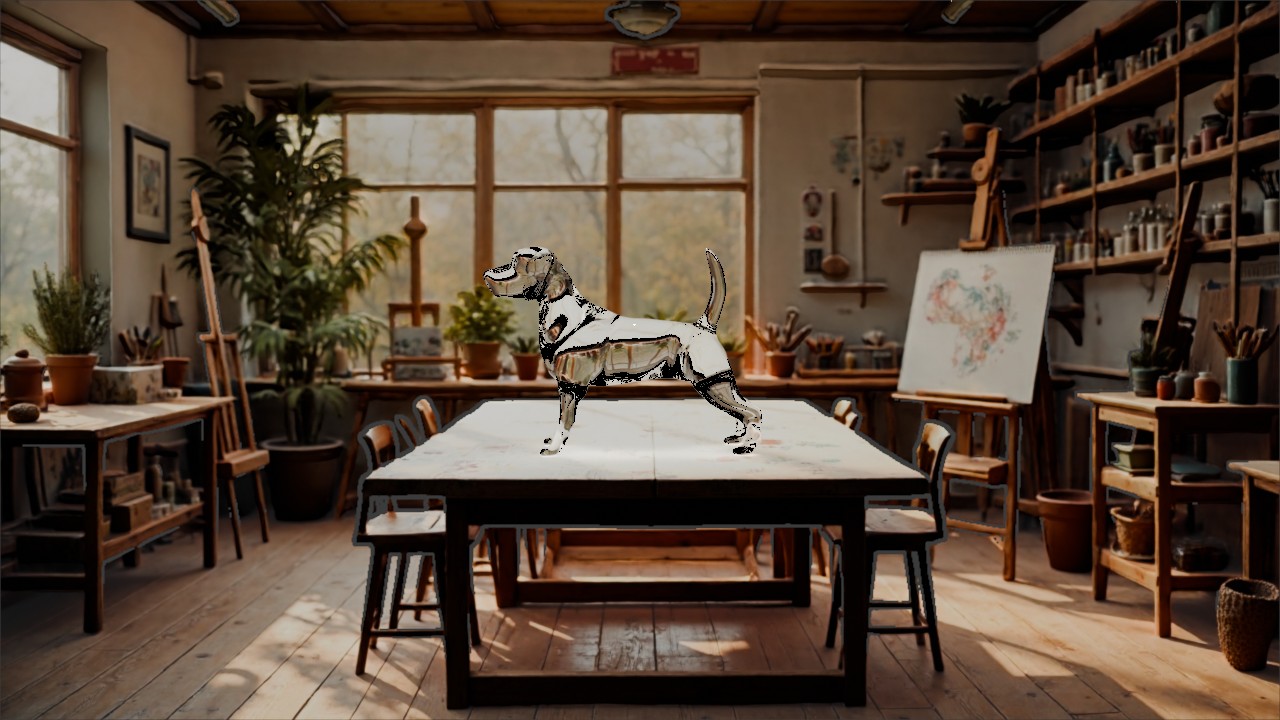}
    \end{subfigure}\hspace{2pt}%
    \begin{subfigure}[]{0.237\linewidth}\centering
        \includegraphics[width=\linewidth,
            trim={256pt 144pt 256pt 144pt}, clip]{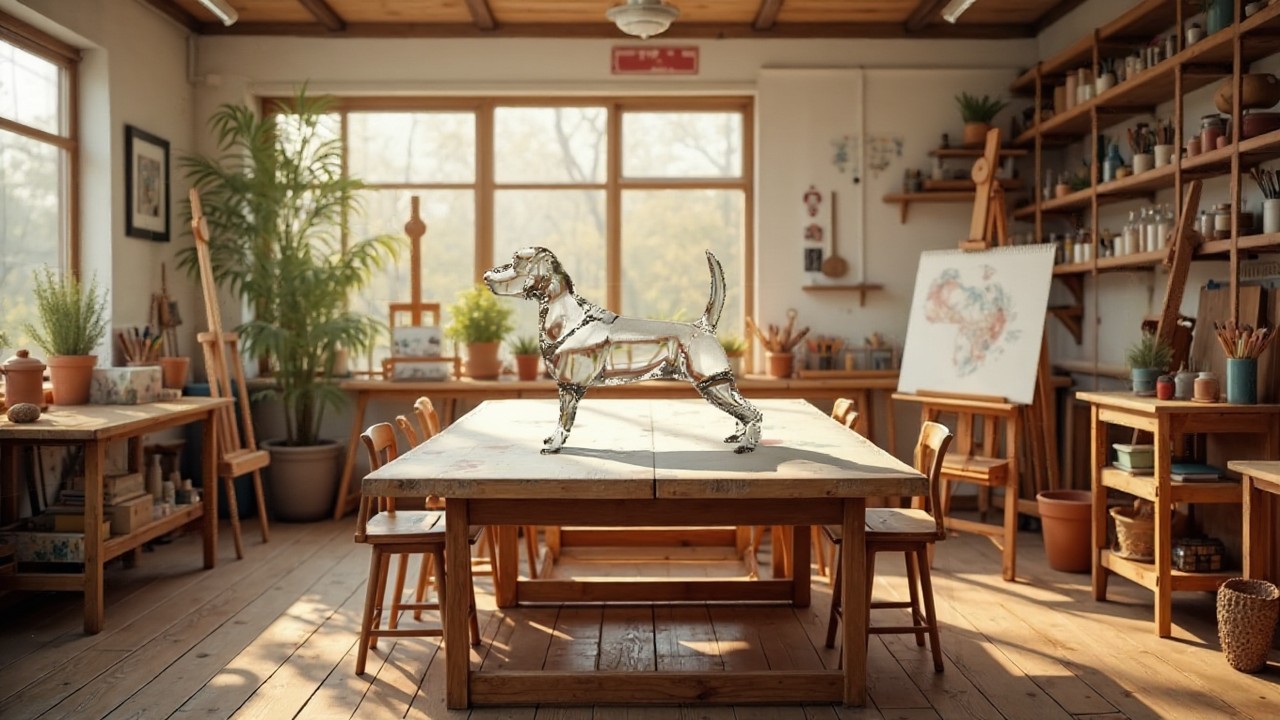}
    \end{subfigure}\hspace{2pt}%
    \begin{subfigure}[]{0.237\linewidth}\centering
        \includegraphics[width=\linewidth,
            trim={256pt 144pt 256pt 144pt}, clip]{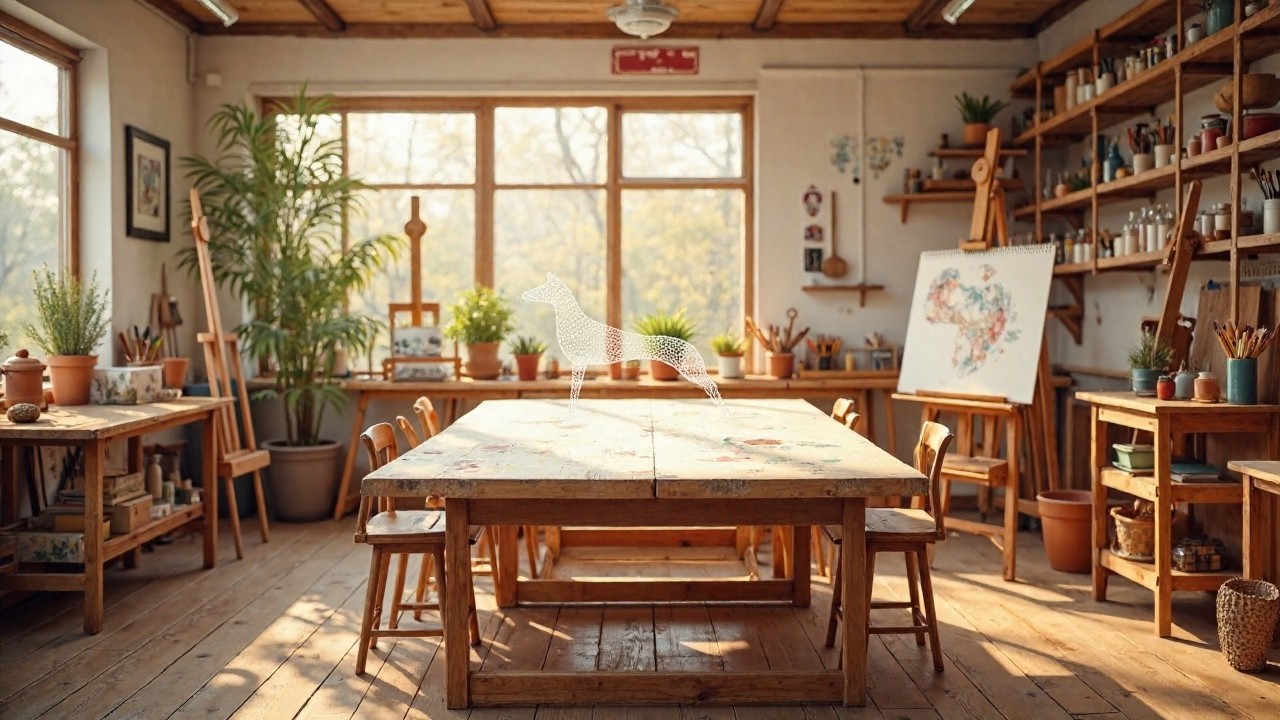}
    \end{subfigure}\hspace{2pt}%
    \begin{subfigure}[]{0.237\linewidth}\centering
        \includegraphics[width=\linewidth,
            trim={256pt 144pt 256pt 144pt}, clip]{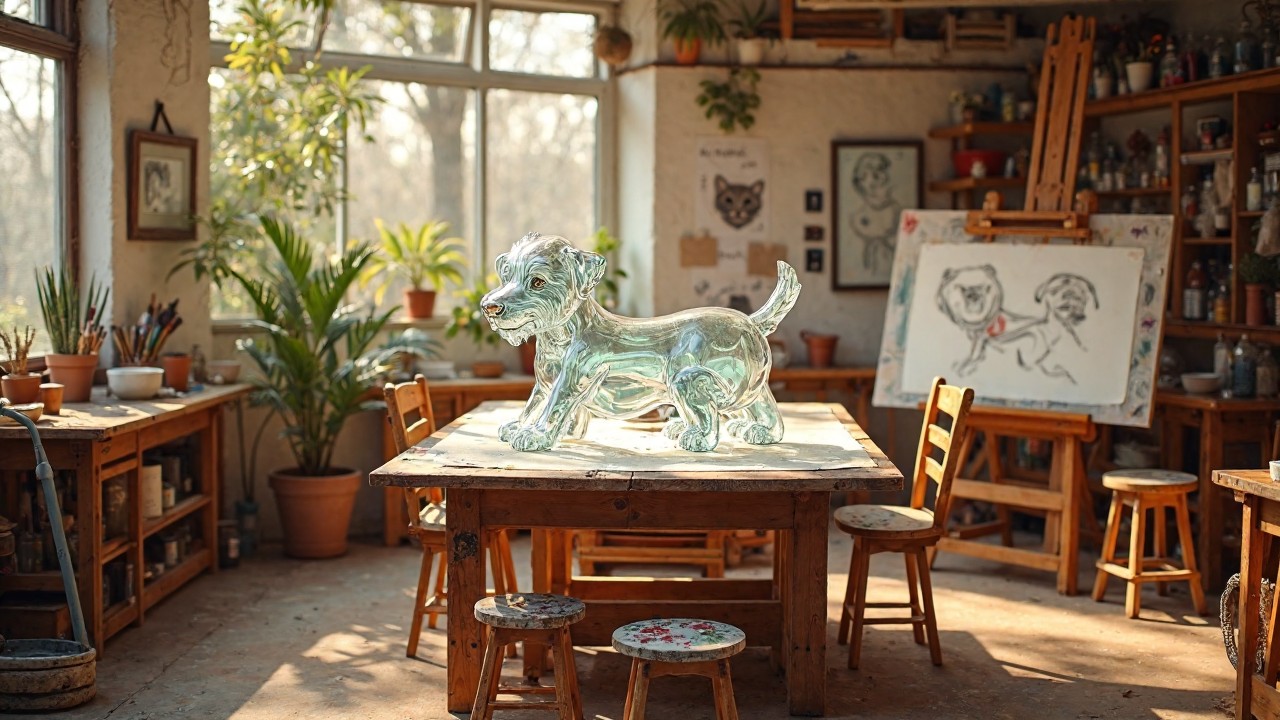}
    \end{subfigure}
    \vspace{-6pt}
    \caption{Qualitative comparison across six scenes.
    The first column is rendered in Blender with the estimated geometry and appearance from $I_0^-$, which provides a reference for the true refractions and reflections, under the caveats that the light sources and colors are incorrect and that there is missing data where refracted surfaces are not directly observed in the image $I_0^-$.
    The last three columns are generated using Snellcaster (ours), a FLUX inpainting model, and the standard FLUX-dev generative model using the same random seeds.
    Our approach conforms significantly better to the true refractions, with the expected left--right and up--down flips and radial warping in sphere scenes.
    }
    \label{fig:results}
\end{figure*}

\section{Experiments}
\label{sec:exp}

In this section, we outline the experimental setup, especially how we measure success in this physics-based generation task, present the qualitative and quantitative results of the method, and report some key analyses.

\subsection{Experimental Setup}

\paragraph{Dataset.}
Our dataset consists of 10 text prompts describing indoor and outdoor scenes that contain a transparent object.
Each prompt has 5 variations, resulting in 50 distinct scenes.
For each scene, we consider 6 different transparent objects (sphere, cylinder, pyramid, fox, dog, and sculpture), yielding a total of 300 unique scene–object combinations.
Indoor scenes, where other objects are closer to the transparent object, are significantly more challenging for modeling refractions, compared to outdoor scenes where the simplifying assumption that surfaces are infinite-distant is more reasonable.
In the main paper, we primarily consider the illustrative case of a glass sphere, which humans are relatively good at grokking: the background should appear flipped about the horizontal and vertical axes of the sphere, with distortion increasing towards the boundary.
Despite the simplicity of the geometry,
image generators are unable to plausibly synthesize images containing this object.
In the supplement, we show more examples of other objects, where the complexity of the refraction paths is challenging even for humans to intuit.

\paragraph{Compared methods.}
We compare our approach against one inpainting model and four state-of-the-art text-to-image models.
The inpainting baseline is a FLUX-based inpainting model~\cite{labs2025flux1kontextflowmatching}, which takes as input the clean reference image $I_0^-$ (corresponding to prompt $p^-$) together with a foreground mask indicating the location of the glass object, and generates the object within the specified region.
The text-to-image models include FLUX-dev~\cite{labs2025flux1kontextflowmatching}, FLUX.2-dev~\cite{flux-2-2025}, Qwen-Image~\cite{wu2025qwen}, and Stable Diffusion 3.5 (large)~\cite{sd3}, conditioned on the full prompt $p$ that explicitly includes the glass object.
For fair comparison, we use the same random seed as that used to generate the clean reference image.

\paragraph{Metrics.}
Evaluation of image generation quality is challenging, especially when assessing the physical correctness of the result.
To assess image quality and how well it conforms to the input text, we report the CLIP score \cite{hessel2021clipscore} and ImageReward score \cite{xu2023imagereward}.
Note that these models are not trained to prefer realistic refractions, so can only be used to evaluate whether our refraction synchronization procedure caused the image quality to deteriorate.
To assess refraction fidelity, we compare the refracted pixel rays to those rendered by computer graphics software (Blender \cite{blender}) from the same RGBD image, masking out those pixels that intersect with unobserved surfaces.
We report the masked peak signal-to-noise ratio (PSNR) on grayscale images with histogram matching, and the masked LPIPS distance.
The color adjustment is necessary, because the Blender image is not harmonized with respect to lighting, so the refracted region will have some expected differences in color and structure.
Nonetheless, our synthesized image should still conform well with the main structure in the ground-truth.

\paragraph{Implementation details.}

We use the FLUX-dev~\cite{labs2025flux1kontextflowmatching} flow-matching architecture, with $T=20$ denoising steps, the default guidance scale of 3.5, and generate images of resolution $720 \times 1280$ (perspective) and $1024 \times 2048$ (equirectangular panoramic).
The algorithm hyperparameters are set as follows:
detail-preserving averaging coefficient $\lambda$ is set to 0.5,
the number of Laplacian pyramid levels is set to 5,
and time travel is repeated 3 times for timesteps in $[0.2, 0.8]T$, following the settings from LookingGlass \cite{chang2025lookingglass}.
Full text prompts ($p, p^-, p^{360}$), along with the prompts for foreground object relighting are provided in the supplement.  
All experiments were run on a NVIDIA A100 80GB GPU.

\begin{table}[!t]
\centering
\small
\caption{
Quantitative comparison of our method against FLUX-dev~\cite{labs2025flux1kontextflowmatching}, FLUX.2-dev~\cite{flux-2-2025}, Qwen-Image~\cite{wu2025qwen}, Stable Diffusion 3.5 (Large)~\cite{sd3}, and FLUX-based inpainting model~\cite{labs2025flux1kontextflowmatching}, averaged across all 300 scene--object combinations (10 prompts $\times$ 5 variations $\times$ 6 objects).
Metrics include the CLIP score, ImageReward score, masked PSNR, and masked LPIPS.
Our method consistently outperforms the baselines in terms of refraction modeling (measured by PSNR and LPIPS), without sacrificing text alignment and image aesthetics (CLIP and ImageReward).
}
\label{tab:results}
\setlength{\tabcolsep}{4pt}
\begin{tabularx}{\linewidth}{@{}lcccc@{}}
    \toprule
    Method & CLIP$\uparrow$ & ImReward$\uparrow$ & PSNR$\uparrow$ & LPIPS$\downarrow$ \\
    [-0.2ex]
    \midrule
    FLUX (dev) & 32.36 & -0.20 & 12.68 & 0.48 \\
    FLUX.2 (dev) & 33.37 & \textbf{0.18} & 12.15 & 0.48 \\
    Qwen-Image & 32.67 & 0.01 & 12.55 & 0.48 \\
    Stable Diffusion 3.5 (L) & \textbf{34.56} & 0.08 & 12.25 & 0.53 \\
    FLUX Inpaint & 33.44 & -0.47 & 12.66 & 0.47 \\
    Snellcaster (ours) & 32.85 & -0.32 & \textbf{16.51} & \textbf{0.24} \\
    \bottomrule
\end{tabularx}
\end{table}

\subsection{Results}
Qualitative results, together with the generating image prompts, are shown in \cref{fig:results}, along with samples from the underlying text-to-image generator to illustrate the implausibility of the synthesized transparent object when generated without physical guidance.
Furthermore, in regions where geometry is missing (masked areas in the Blender reference images), our method produces plausible content by leveraging the synchronized panorama to fill in unseen regions.
By contrast, FLUX inpainting and the standard FLUX-dev model generate physically implausible results.
For instance, in the living room scene, FLUX inpainting fails to render the sofa behind the sphere, showing unrelated content instead.
In the karaoke scene, although both the FLUX inpainting model and FLUX-dev generate glass spheres with tones consistent with the room, background details such as the TV and speakers are completely absent within the sphere.
In the landscape scene, the standard FLUX-dev output appears reasonable at first glance, but closer inspection reveals that the mountains in the background are not refracted in the sphere at all.
Overall, standard FLUX-dev and FLUX inpainting both fail to synthesize correct physical refractions, either omitting the background objects or rendering something that does not conform to the rest of the image evidence.

Quantitative results are presented in \cref{tab:results}.
Across both per-pixel metrics, our method outperforms all other methods.
These results confirm that our approach not only produces visually more accurate images but also better aligns with the text prompts and scene geometry, demonstrating the effectiveness of cross-view synchronization and physics-based refraction modeling.

\begin{figure}[!t]
    \centering
    \begin{minipage}[t]{0.49\linewidth}
        \centering
        \includegraphics[width=\linewidth,
            trim={0pt 50pt 0pt 0pt}, clip]{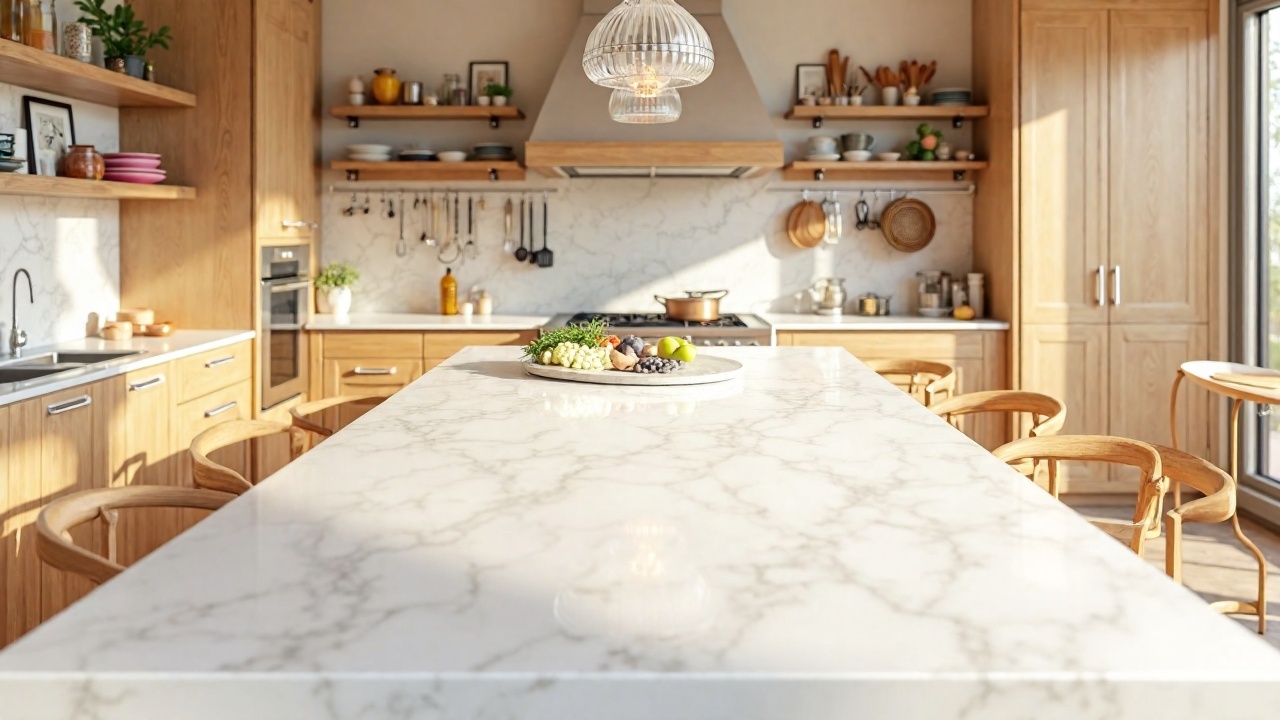}
    \end{minipage}
    \hfill
    \begin{minipage}[t]{0.49\linewidth}
        \centering
        \includegraphics[width=\linewidth,
            trim={0pt 50pt 0pt 0pt}, clip]{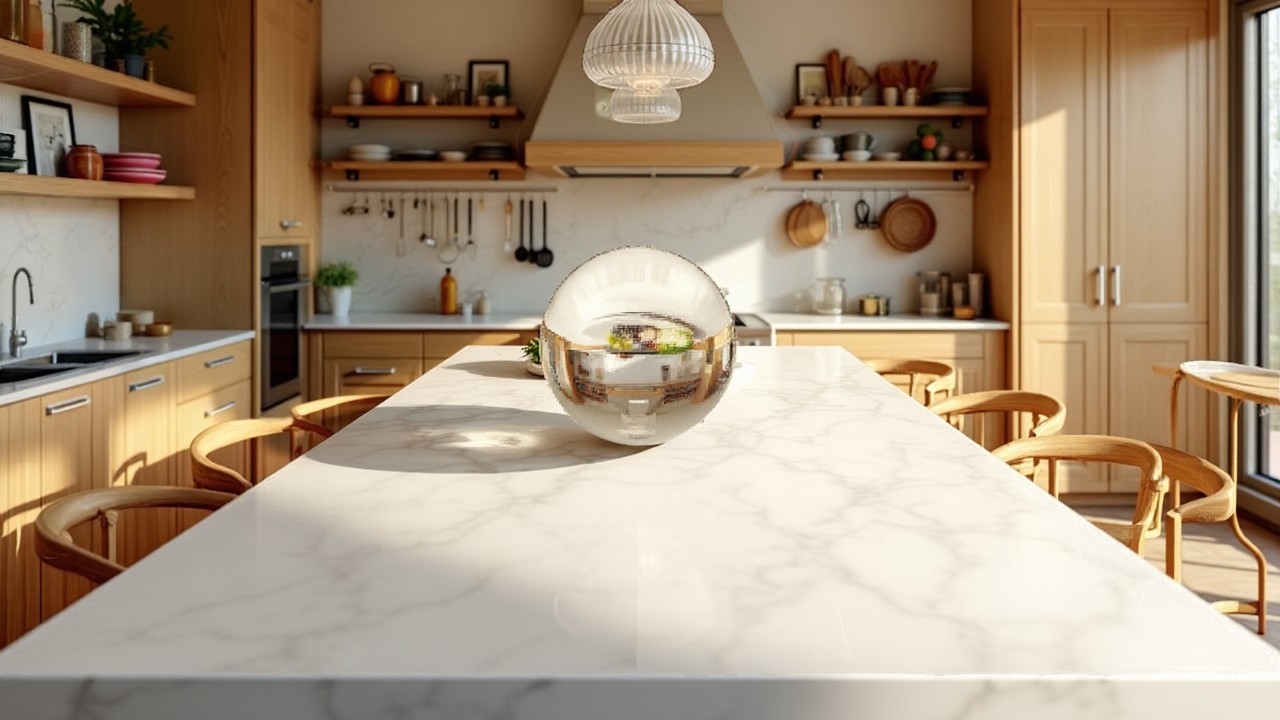}
    \end{minipage}
    \par\vspace{2pt}
    \begin{minipage}[t]{\linewidth}
        \centering
        \includegraphics[width=\linewidth,
            trim={0pt 100pt 0pt 0pt}, clip]{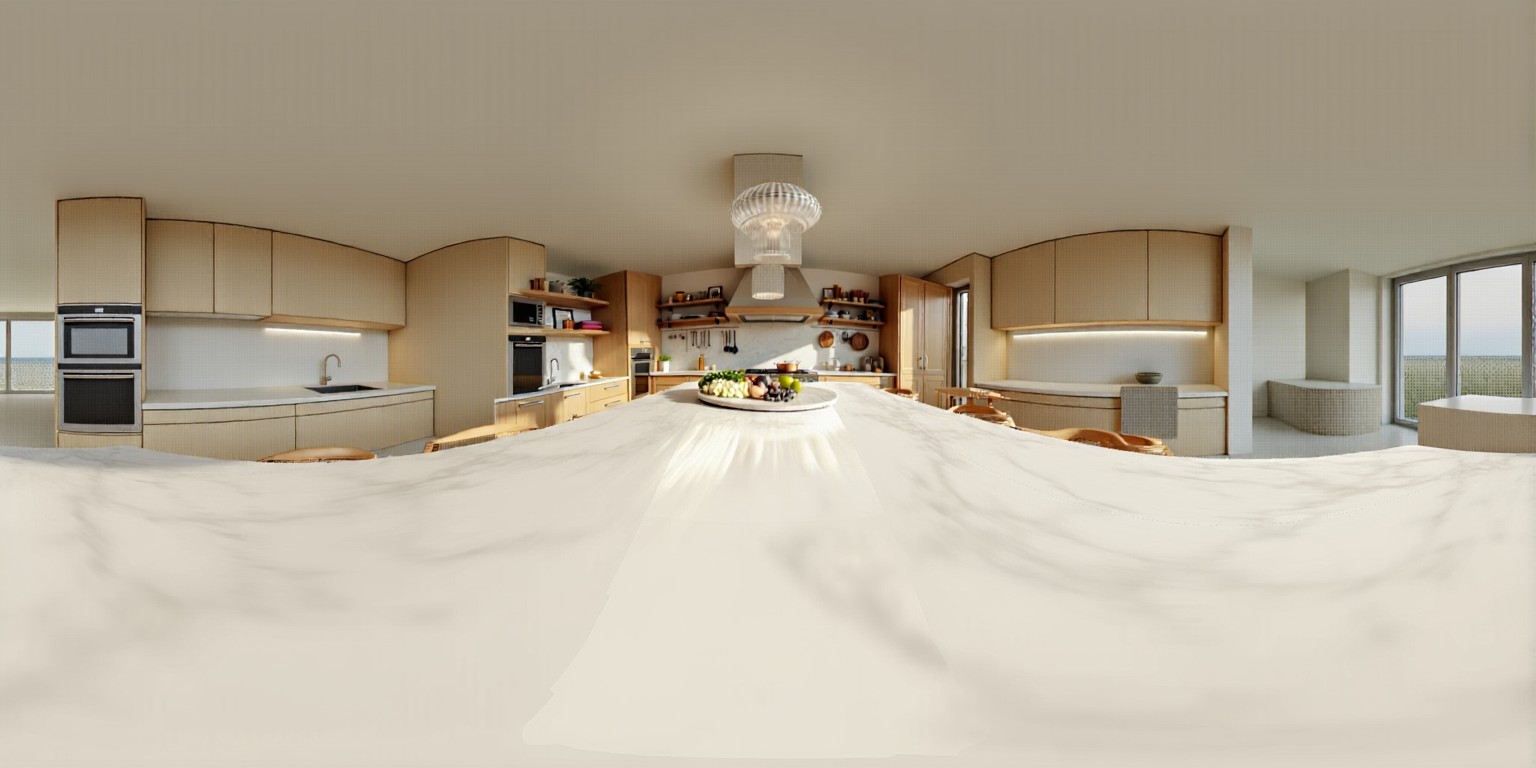}
    \end{minipage}
    \par\vspace{-8pt}
    \caption{
    Kitchen scene example of the synchronized object-free image $I_0^-$ (top left), the generated perspective image $I_0$ (top right), and the auxiliary generated panorama $I_0^{360}$ (bottom).
    The panorama extends the scene in a plausible way that is consistent with the perspective view.
    }
    \label{fig:pano}
\end{figure}

In \cref{fig:pano}, we give an example of the object-free image $I_0^-$ and the auxiliary generated panorama $I_0^{360}$, as well as the generated perspective image $I_0$, for one representative scene.
This shows that the panoramic completion of the scene is plausible and is consistent with the perspective view, with sufficient detail for the sphere reflections.

\begin{figure*}[!t]
    \centering
    \captionsetup[subfigure]{labelformat=empty,font=footnotesize}
    \begin{minipage}[t]{0.245\linewidth}
        \centering
        \includegraphics[width=\linewidth,
            trim={270pt 120pt 220pt 105.7pt}, clip]{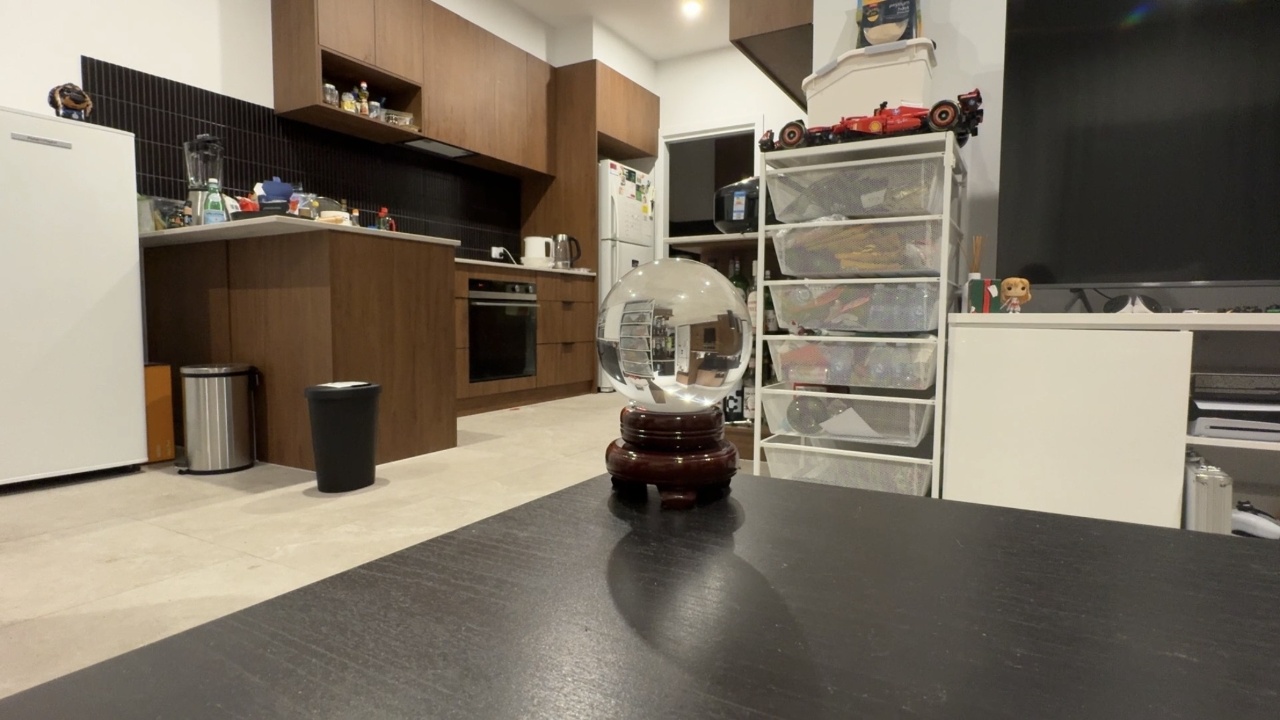}
            \vspace{-13pt}
            \subcaption{Indoor real scene}%
    \end{minipage}\hfill
    \begin{minipage}[t]{0.245\linewidth}
        \centering
        \includegraphics[width=\linewidth,
            trim={270pt 120pt 220pt 105.7pt}, clip]{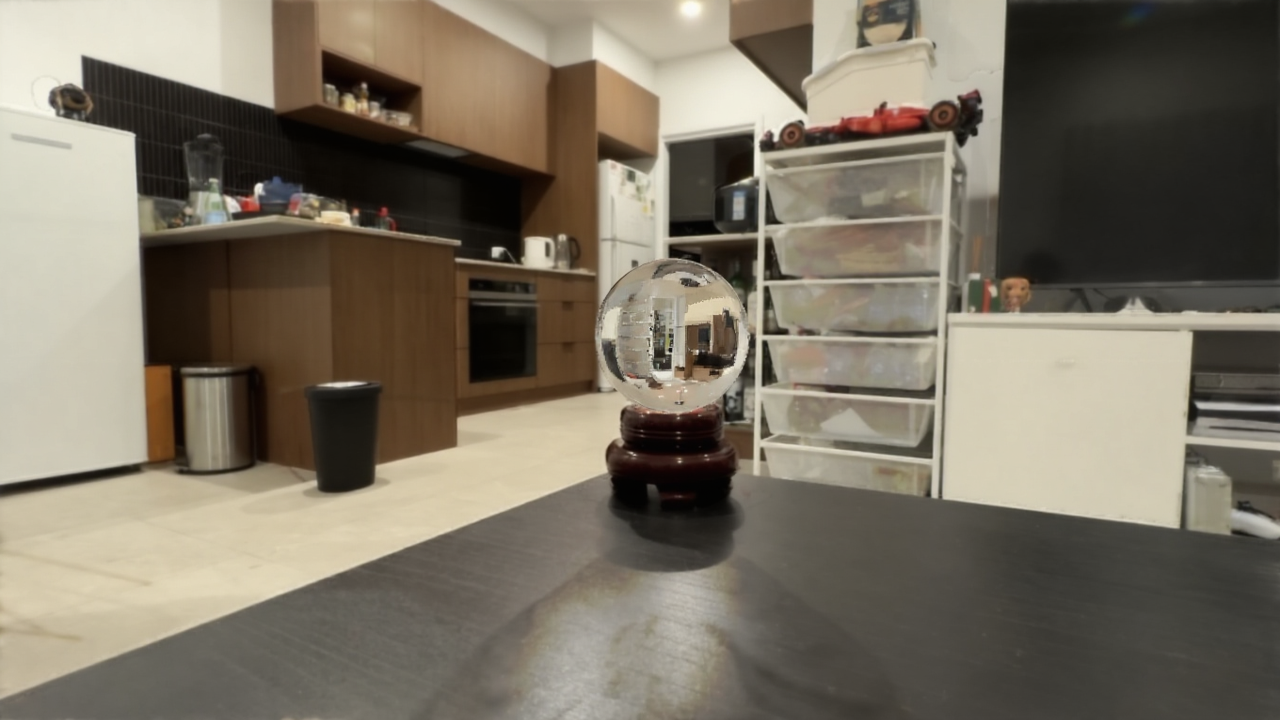}
            \vspace{-13pt}
            \subcaption{Indoor generated scene}%
    \end{minipage}
    \begin{minipage}[t]{0.245\linewidth}
        \centering
        \includegraphics[width=\linewidth,
            trim={0pt 210pt 0pt 90pt}, clip]{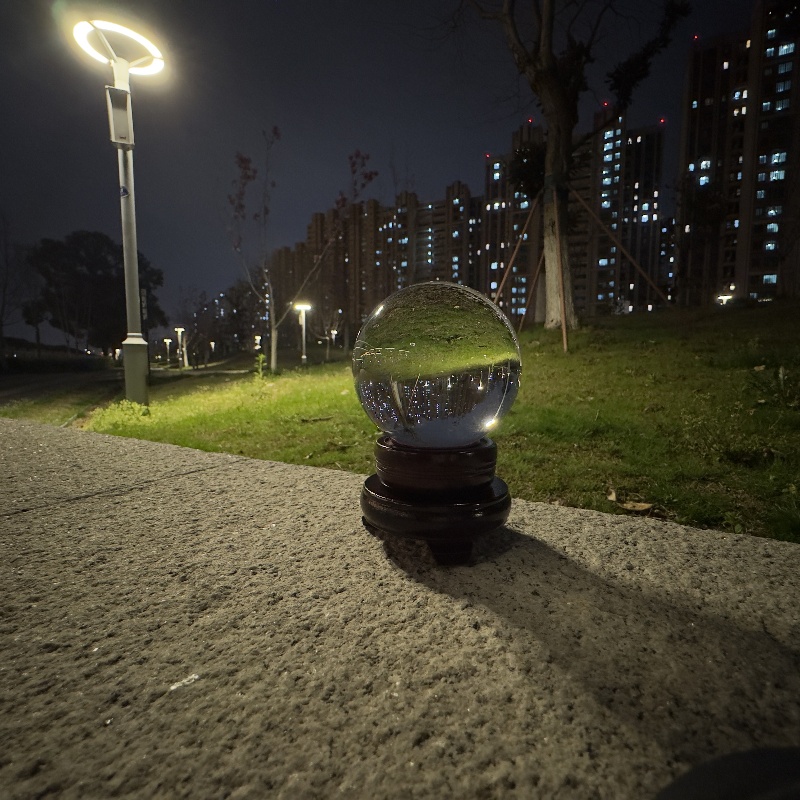}
            \vspace{-13pt}
            \subcaption{Outdoor real scene}%
    \end{minipage}\hfill
    \begin{minipage}[t]{0.245\linewidth}
        \centering
        \includegraphics[width=\linewidth,
            trim={0pt 210pt 0pt 90pt}, clip]{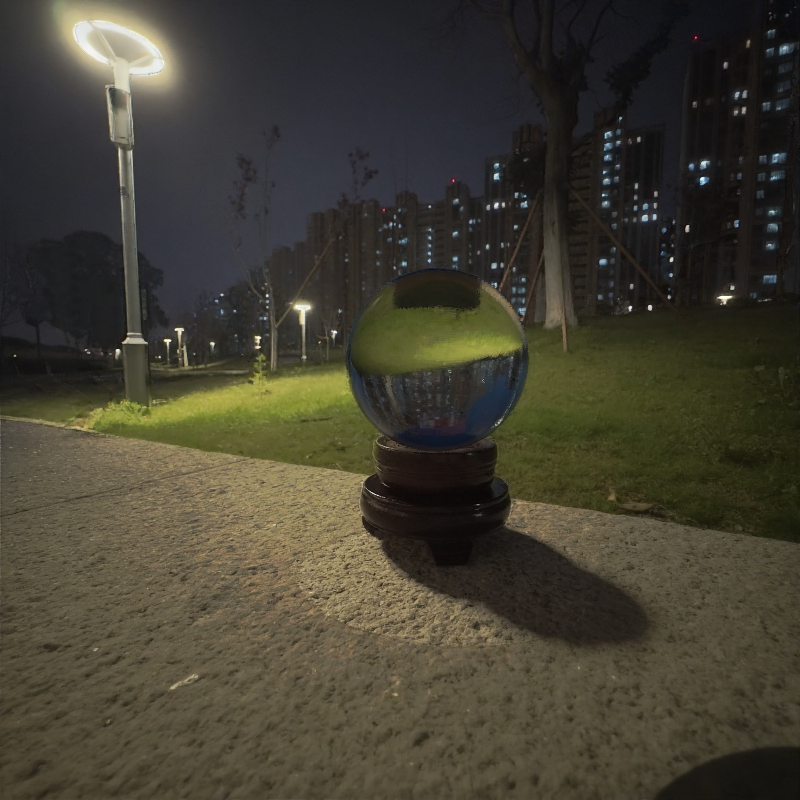}
            \vspace{-13pt}
            \subcaption{Outdoor generated scene}%
    \end{minipage}%
    \vspace{-5pt}
    \caption{Real-world evaluation on captured scenes.
    We show two pairs of results from real image pairs collected by a phone camera. 
    Each pair consists of a real photograph containing a transparent sphere (left) and a corresponding image synthesized by our method (right).
    The first pair shows an indoor kitchen scene, and the second pair shows an outdoor night park scene.
    Our method takes the image without the transparent sphere as $I_0^-$ and generates the final image with a physically plausible glass sphere inserted into the scene.}
    \label{fig:real}
\end{figure*}

Moreover, to evaluate our method on real-world data, we collect several image pairs, each consisting of one image without a transparent sphere and one with a transparent sphere, captured with a fixed camera.
The image without the transparent sphere is used as $I_0^-$.
Starting from random noise, we then apply our generation pipeline to synthesize the final image.
Two examples are shown in~\cref{fig:real}.

\subsection{Ablation Study}
Quantitative ablation results are shown in \cref{tab:ablation}, where we evaluate the effect of two key components: the reflection modeling and the panorama synchronization.
To ensure a fair comparison, all ablations are performed before relighting, since relighting introduces additional appearance changes that are not directly comparable across variants.
We observe that removing either component leads to worse performance, especially for the per-pixel metrics.
Among the two components, removing the panorama synchronization leads to a more significant performance drop than removing the reflection component.
This is because, while reflections are important, their contribution is relatively subtle compared to refraction, which dominates the appearance of the object.
In contrast, the panorama branch provides essential information for regions inside the object that are not directly observed, such as areas that are outside the camera’s field of view or occluded by objects.
Without this branch, the model must hallucinate these regions, often producing content that is inconsistent with the surrounding scene and does not align smoothly with the background.
We further provide a qualitative comparison in \cref{fig:ablation}, where we visualize this effect on a landscape scene.
We also show the quantitative effect of relighting in \cref{tab:ablation}.
Although relighting slightly degrades per-pixel metrics, this is expected because it introduces additional lighting and shadow effects that deviate from the ground truth at a pixel level.
However, these effects improve the overall visual plausibility of the results, especially the shadow cast by the object, highlighting a trade-off between numerical metrics and perceptual quality.
Overall, these results demonstrate that both reflection modeling and panorama synchronization contribute meaningfully to reconstruction accuracy, while relighting primarily improves realism rather than per-pixel metrics.

\section{Conclusion}
\label{sec:conc}

\begin{table}[!t] \centering 
\small
\caption{
Quantitative ablation study on a subset of six scenes (artroom, cafe, dining room, kitchen, living room, office) with the sphere geometry.
Ablating either the reflection and panorama components degrades the per-pixel performance (PSNR, LPIPS).
Adding in the relighting leads to worse per-pixel metrics, but incorporating shadows is important for visual plausibility.
}
\label{tab:ablation}
\setlength{\tabcolsep}{4pt}
\begin{tabular}{l c c c c c}
\toprule
Model Variant & MAE$\downarrow$ & PSNR$\uparrow$ & LPIPS$\downarrow$ & CLIP$\uparrow$ & ImgR$\uparrow$ \\
\midrule
Ours & \textbf{0.0953} & \textbf{18.21} & \textbf{0.24} & \underline{34.22} & 0.51 \\ %
w/o reflections  & \underline{0.0964} & \underline{18.11} & \underline{0.25} & 34.20 & \textbf{0.67} \\ %
w/o pano sync & 0.0983 & 17.98 & \underline{0.25} & 33.88 & \underline{0.64} \\ %
w/ relighting & 0.1137 & 17.10 & 0.28 & \textbf{34.61} & 0.59  \\
\bottomrule
\end{tabular}
\end{table}

\begin{figure}[!t]
    \centering
    \captionsetup[subfigure]{labelformat=empty,font=footnotesize}
    \begin{minipage}[t]{0.495\linewidth}
        \centering
        \includegraphics[width=\linewidth]{figures/splash/blender_land.jpg}
            \vspace{-13pt}
            \subcaption{Blender Reference}%
    \end{minipage}\hfill
    \begin{minipage}[t]{0.495\linewidth}
        \centering
        \includegraphics[width=\linewidth]{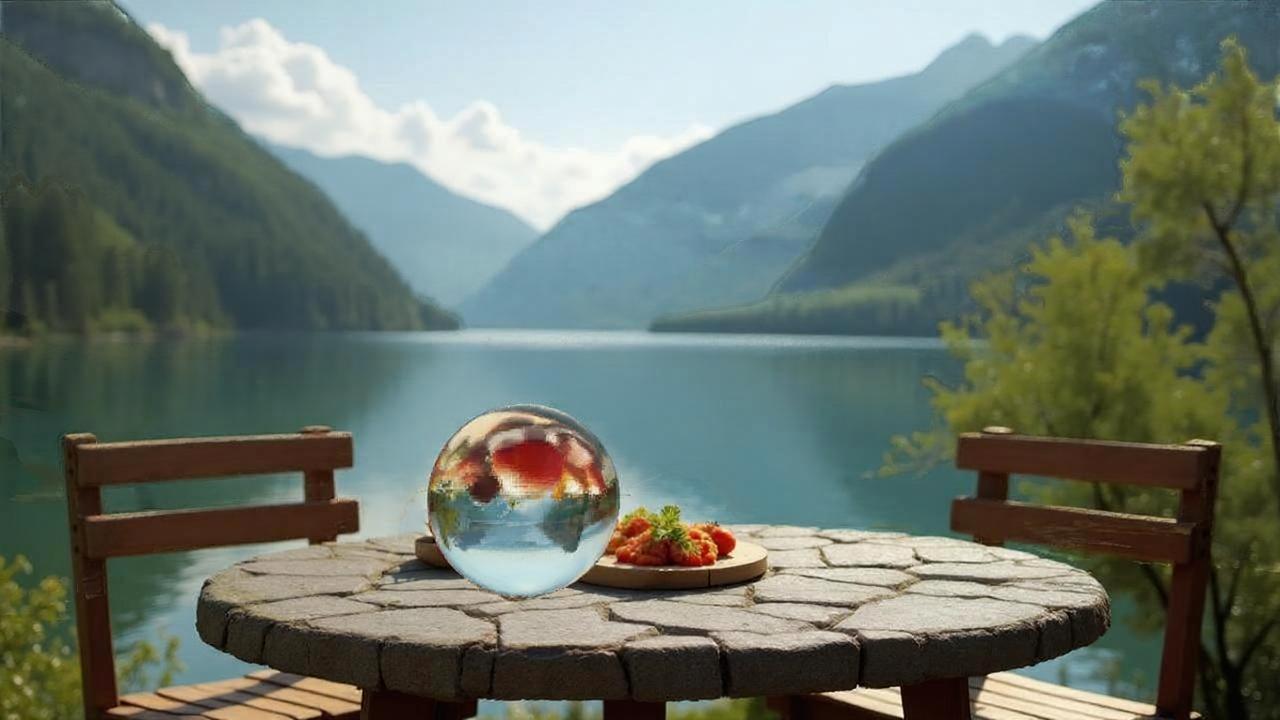}
            \vspace{-13pt}
            \subcaption{Ours}%
    \end{minipage}\\
    \begin{minipage}[t]{0.495\linewidth}
        \centering
        \includegraphics[width=\linewidth]{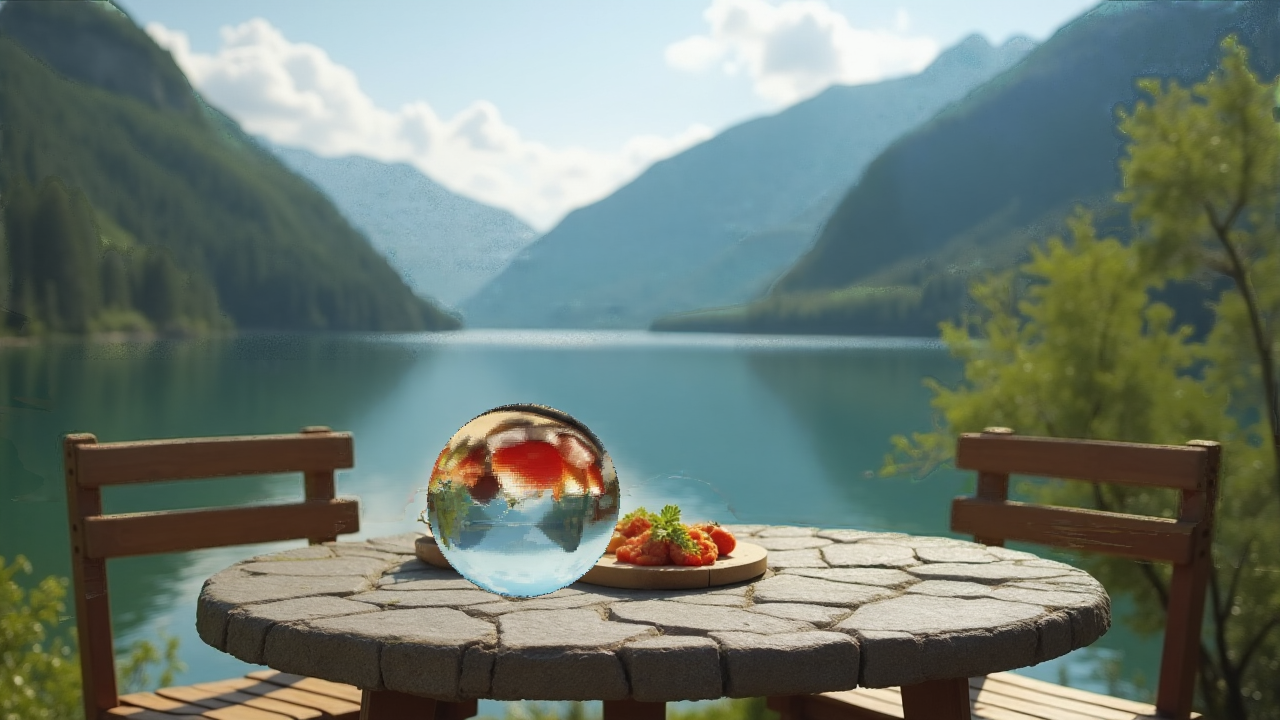}
            \vspace{-13pt}
            \subcaption{Ours w/o reflection}%
    \end{minipage}\hfill
    \begin{minipage}[t]{0.495\linewidth}
        \centering
        \includegraphics[width=\linewidth]{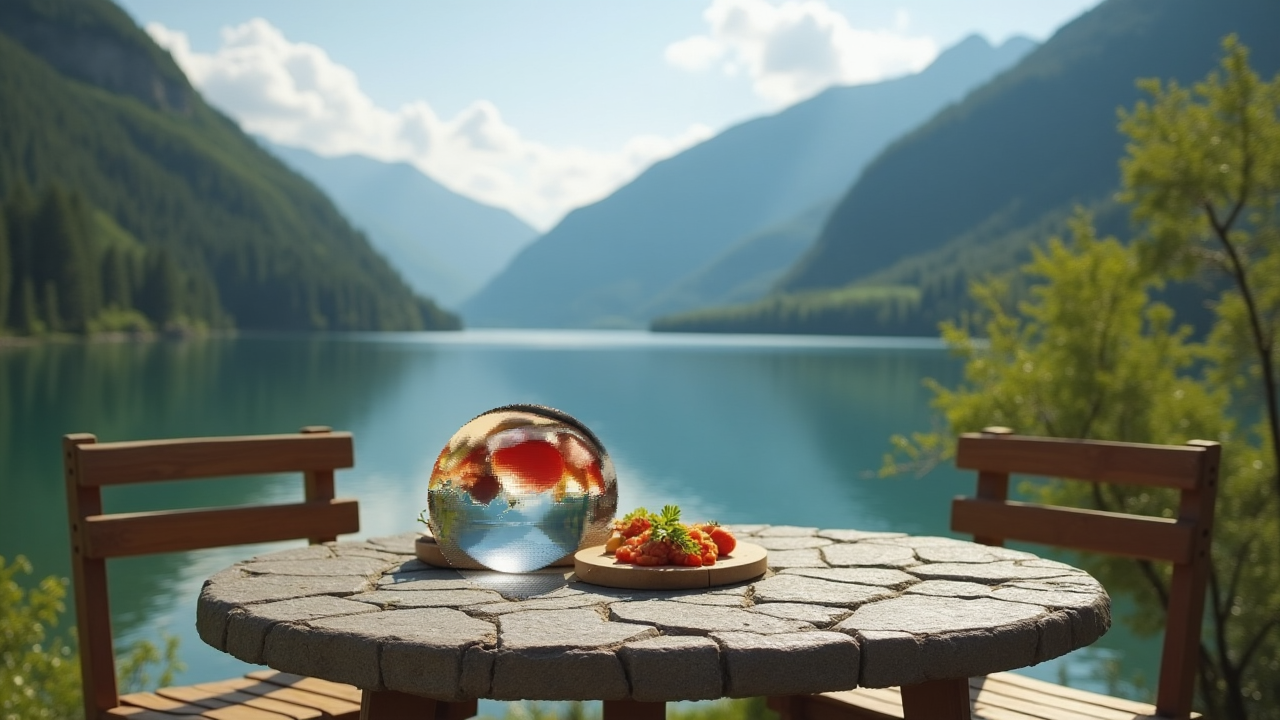}
            \vspace{-13pt}
            \subcaption{Ours w/o panorama branch}%
    \end{minipage}%
    \vspace{-5pt}
    \caption{Qualitative ablation on a landscape scene comparing the Blender reference, full method, w/o reflection modeling, and w/o panorama synchronization.
    Removing panorama synchronization leads to missing or hallucinated content in out-of-view regions, showing its importance for global scene consistency.}
    \vspace{-3pt}
    \label{fig:ablation}
\end{figure}

In this work, we have presented a training-free approach for generating images containing a transparent object, where the physical principles of refraction and reflection are respected.
It employs cross-view image synchronization to ensure that the generated perspective image $I$, the object-free perspective image $I^-$, and the panoramic image from the perspective of the transparent object $I^{360}$ are compatible, and that the refractive regions receive the correct color.
The panorama synchronization further ensures that occluded and out-of-frame regions are plausibly completed, so that the refracted rays that do not hit an observed surface receive a plausible color, and similarly for the reflected rays.

There are many promising directions for future work.
Minor extensions include
modeling absorption (\eg, tinting) of the transparent object,
modeling multiple material types (\eg, a glass of water with a straw),
modeling birefringent materials (\eg, Iceland spar),
and text-controllable object placement.
More significant directions include
native lighting and shadow handling via light source estimation and ray tracing,
improving the accuracy and consistency of the seen--unseen surface boundary by synchronizing RGBD images and panoramas,
and extending the approach to video generation with optically-accurate refractions and reflections.
The latter is currently a major failure case of video generation models and acts counter to an immersive suspension of disbelief.

\newpage
\paragraph{Acknowledgments.}
Dr Campbell is the recipient of an Australian Research Council Discovery Early Career Award (project number DE250100542) funded by the Australian Government.

{
    \small
    \bibliographystyle{ieeenat_fullname}
    \bibliography{main}
}

\setcounter{page}{1}
\twocolumn[{
\begin{center}
    {\Large \textbf{Supplementary Material}}
\end{center}
\vspace{1em}
}]

\appendix

\section{Additional Methodological Details}

\paragraph{Laplacian pyramid warping.}
When an image is warped from one view to another, different regions of the source image may undergo varying amounts of stretching or compression.
This leads to frequency mismatch: highly stretched regions tend to lose high frequency details while other regions may retain them.
We address this issue by performing Laplacian pyramid warping, as used in LookingGlass \cite{chang2025lookingglass}.
Given a source image, we construct its Laplacian pyramid, where each level isolates a specific frequency band.
We then apply the geometric warp independently to every pyramid level.
Since each level contains only a narrow range of frequencies, the warp produces smoother and more stable results with fewer large scale distortions.
After all levels are warped, we collapse the warped pyramid to obtain the final aligned image.
In our implementation, we use a five level Laplacian pyramid for all warping operations.

\paragraph{Time travel.}
Similar to LookingGlass \cite{chang2025lookingglass}, we also adopt the time travel strategy \cite{wangzero} to allow the model to blend different views better.
The idea is to travel by $l$ steps to a noisier timestep and then let the cleaner state $z_{0|t}$ guide the update, since using the cleaner prediction $z_{0|t}$ from the current timestep provides a more reliable estimate than $z_{0|t+l}$.
At a chosen timestep $t$, we first sample a latent state at a noisier timestep $t + l$ using the transition distribution $q(z_{t+l}\,|\,z_t)$.
We then restart the reverse process from $t + l$ and continue denoising to $t - 1$.
This operation effectively rewrites the recent sampling history, using a more reliable clean prediction at timestep $t$ to guide subsequent updates.
Repeating this procedure improves global consistency around challenging regions.
In our implementation, we follow LookingGlass and apply time travel only between $20\%$ and $80\%$ of the sampling steps, with a travel length of one, and we repeat the procedure three times.

\paragraph{Foreground object relighting.}
To enhance realism, we relight the foreground object using a relighting (harmonization) model \cite{labs2025flux1kontextflowmatching}.
We provide a text description of the object and its desired illumination, and the model creates the soft shadows and caustics accordingly.
We repeat 20 times of this process and choose the one that minimally changes the appearance w.r.t. PSNR in the transparent object region.
For instance, for a glass sphere on the table of an artroom scene, we use a prompt such as ``\textit{Creating soft shadows and caustics under the glass sphere on the table.}"
This relighting step improves the coherence between the inserted transparent object and the surrounding generated scene.

\begin{figure*}[!t]
\centering
\resizebox{\textwidth}{!}{
\begin{tikzpicture}[
    font=\footnotesize,
    fullbox/.style={
        rectangle,
        rounded corners,
        draw=black!20,
        fill=blue!3,
        inner sep=4pt,
        text width=0.53\textwidth,
        align=left
    },
    panobox/.style={
        rectangle,
        rounded corners,
        draw=black!20,
        fill=blue!3,
        inner sep=4pt,
        text width=0.3\textwidth,
        align=left
    },
    title/.style={
        font=\bfseries\footnotesize,
        align=center
    },
    scenelabel/.style={
        font=\bfseries\scriptsize,
        align=center
    }
]

\def\xfull{1.9}
\def\xpano{9.5}

\def\xlabel{-3.2}

\newcommand{\rowy}[1]{-#1}

\node[title, text width=0.53\textwidth] at (\xfull,0) {Full Scene Prompt $p$};
\node[title, text width=0.3\textwidth] at (\xpano,0) {Panorama Prompt $p^{360}$};

\node[scenelabel, rotate=90] at (\xlabel,\rowy{1.0})
    {\shortstack[l]{Living\\[-3pt]Room}};
\node[scenelabel, rotate=90] at (\xlabel,\rowy{2.27})
    {\shortstack[l]{Dining\\[-3pt]Room}};
\node[scenelabel, rotate=90] at (\xlabel,\rowy{3.54}) {Office};
\node[scenelabel, rotate=90] at (\xlabel,\rowy{4.81}) {Kitchen};
\node[scenelabel, rotate=90] at (\xlabel,\rowy{6.08}) {Artroom};
\node[scenelabel, rotate=90] at (\xlabel,\rowy{7.35}) {Café};
\node[scenelabel, rotate=90] at (\xlabel,\rowy{8.62}) {Cave};
\node[scenelabel, rotate=90] at (\xlabel,\rowy{9.89}) {Desert};
\node[scenelabel, rotate=90] at (\xlabel,\rowy{11.16}) {Karaoke};
\node[scenelabel, rotate=90] at (\xlabel,\rowy{12.43}) {Landscape};

\node[fullbox] at (\xfull,\rowy{1.0}) {
\textit{A cozy living room with a polished wooden coffee table close to the camera, {\color{red}\textbf{with a big glass sphere on top,}} surrounded by a beige sofa, a patterned rug, plants, bookshelves, framed wall art, and sunlight through sheer curtains.}
};
\node[panobox] at (\xpano,\rowy{1.0}) {
\textit{A high resolution equirectangular 360 degree panorama captured on top of a polished wooden coffee table in a cozy living room.}
};

\node[fullbox] at (\xfull,\rowy{2.27}) {
\textit{A bright dining room with a wooden dining table in the center, {\color{red}\textbf{with a big glass sphere on top,}} surrounded by upholstered chairs, a pendant lamp, fruit bowls, paintings, and daylight through tall windows.}
};
\node[panobox] at (\xpano,\rowy{2.27}) {
\textit{A high resolution equirectangular 360 degree panorama captured on top of a wooden dining table in a bright dining room.}
};

\node[fullbox] at (\xfull,\rowy{3.54}) {
\textit{A minimalist home office with a smooth wooden desk closer to the camera, {\color{red}\textbf{with a big glass sphere on top,}} surrounded by a black office chair, bookshelves with plants, framed posters, a side table with a monitor.}
};
\node[panobox] at (\xpano,\rowy{3.54}) {
\textit{A high resolution equirectangular 360 degree panorama captured on top of a smooth wooden desk in a minimalist home office.}
};

\node[fullbox] at (\xfull,\rowy{4.81}) {
\textit{A modern kitchen with a large marble island in the center, {\color{red}\textbf{with a big glass sphere on top,}} surrounded by wooden cabinetry, bar stools, hanging lights, utensils, and reflections from stainless steel appliances under morning light.}
};
\node[panobox] at (\xpano,\rowy{4.81}) {
\textit{A high resolution equirectangular 360 degree panorama captured on top of a marble island in a modern kitchen.}
};

\node[fullbox] at (\xfull,\rowy{6.08}) {
\textit{An art classroom with a rectangular wooden worktable near the camera, {\color{red}\textbf{with a big glass sphere on top,}} surrounded by easels, color splattered stools, sketches, jars of brushes, and warm daylight through wide windows.}
};
\node[panobox] at (\xpano,\rowy{6.08}) {
\textit{A high resolution equirectangular 360 degree panorama captured on top of a wooden worktable in an artroom.}
};

\node[fullbox] at (\xfull,\rowy{7.35}) {
\textit{A minimalist café interior with a square wooden table in the foreground, {\color{red}\textbf{with a big glass sphere on top,}} surrounded by metal framed chairs, plants, hanging lights, a pastry counter, and sunlight on the tiled floor.}
};
\node[panobox] at (\xpano,\rowy{7.35}) {
\textit{A high resolution equirectangular 360 degree panorama captured on top of a square wooden table in a minimalist café interior.}
};

\node[fullbox] at (\xfull,\rowy{8.62}) {
\textit{A rocky cave lit by a bright campfire, warm flickering light casting shadows, {\color{red}\textbf{a big glass sphere on the ground,}} with scattered camping gear, tents, sleeping bags, backpacks, lanterns, and cooking pots, with smoke and embers in the air.}
};
\node[panobox] at (\xpano,\rowy{8.62}) {
\textit{A high resolution equirectangular 360 degree panorama captured from the ground of a rocky cave, camping gear scattering around.}
};

\node[fullbox] at (\xfull,\rowy{9.89}) {
\textit{A high-noon desert scene with blinding sunlight and hard shadows, heat haze over sand and rocks, {\color{red}\textbf{a big glass sphere on the ground,}} and camping gear in the foreground, tent, backpacks, and a small stove.}
};
\node[panobox] at (\xpano,\rowy{9.89}) {
\textit{A high resolution equirectangular 360 degree panorama captured from the ground of a desert scene, with camping gear around.}
};

\node[fullbox] at (\xfull,\rowy{11.16}) {
\textit{A karaoke room with colorful lights, TV on the wall displaying music videos, a coffee table {\color{red}\textbf{with a big glass sphere on top,}} in front of the TV, and a sofa around the coffee table.}
};
\node[panobox] at (\xpano,\rowy{11.16}) {
\textit{A high resolution equirectangular 360 degree panorama captured from on top of a coffee table in a colorful karaoke room.}
};

\node[fullbox] at (\xfull,\rowy{12.43}) {
\textit{A beautiful landscape with a river and mountains, viewed from a camera directly in front of a stone table and chairs in the foreground, filling the lower frame, {\color{red}\textbf{a big glass sphere on the table,}} on the left of colorful food.}
};
\node[panobox] at (\xpano,\rowy{12.43}) {
\textit{A high resolution equirectangular 360 degree panorama captured from on top of a stone table of a beautiful landscape.}
};

\end{tikzpicture}\unskip
}
\vspace{-18pt}
\caption{
Prompts used for generating scenes.
The full prompts $p$ are shown on the left, where the transparent object phrases are highlighted in red bold text.
The object free prompts $p^{-}$ are obtained by replacing the transparent object description in each full prompt with wording that states the surface is clean and empty, while keeping all other scene details unchanged.
The right column shows the panorama prompts $p^{360}$.
The relighting prompt $p^{r}$ is: \textit{Create soft shadows and caustics under $p^{\mathrm{obj}}$}.
}
\label{fig:prompts}
\end{figure*}

\section{From Text to Object Parameters}
\begin{figure}[!t]
\centering
\resizebox{\linewidth}{!}{
\input{figures/suppl/prompt23D.tikz}\unskip
}
\caption{An example of extracting the object specific prompt $p^{\mathrm{obj}}$ using an LLM and generating a 3D mesh with TRELLIS. From the full scene prompt  
``\textit{An art classroom with a rectangular wooden worktable near the camera, with a glass fox on top, surrounded by easels, color splattered stools, sketches, jars of brushes, and warm daylight through wide windows}"  
the LLM identifies ``\textit{a glass fox on top}" and distills the object description ``\textit{a glass fox}" as $p^{\mathrm{obj}}$, which is then provided to TRELLIS to produce the 3D fox mesh.
}
\label{fig:prompt23D}
\end{figure}
For each scene category, we use the prompts $p$, $p^{-}$, $p^{360}$, and $p^{r}$ to guide the generation process, as illustrated in \cref{fig:prompts}. 
The full prompt $p$ describes the scene together with the transparent object, while the object free prompt $p^{-}$ removes the object and instead specifies that the supporting surface is clean and empty. 
The panorama prompt $p^{360}$ describes a high resolution equirectangular 360 degree panorama captured from the center of the transparent object, providing additional scene context beyond the perspective view. 
The relighting prompt $p^{r}$ adds physically plausible shadows and caustics using the instruction ``\textit{Create soft shadows and caustics under $p^{\mathrm{obj}}$}".

Given a text prompt $p$ that describes a scene containing a transparent object, we derive the three components required by our method: a 3D model of the object, its refractive index, and its pose relative to the camera.
These are obtained automatically using a sequence of lightweight language model queries together with a text to 3D generator.

\paragraph{Prompt decomposition.}
Starting from the full prompt $p$, we ask the large language model Gemma \cite{team2024gemma} to identify the span of text that refers to the transparent object.
For the first prompt describing the living room in \cref{fig:prompts}, the LLM identifies the removable phrase
``\textit{with a big glass sphere on top}".
From this phrase, we extract the core object description
``\textit{a glass sphere}",
which we denote as $p^{\mathrm{obj}}$.  
To construct the object free prompt $p^{-}$, we replace the entire identified phrase ``\textit{a big glass sphere on top}" with ``\textit{surface clean and empty}", while keeping all other scene details unchanged.

\paragraph{Refractive index extraction.}
We obtain the refractive index by querying the LLM directly with the object description $p^{\mathrm{obj}}$.  
For common materials such as glass or water, the returned values are well defined and stable across queries.  
In the example above, asking ``\textit{What is the refractive index of a glass sphere}" typically returns a value of $1.5$.

\paragraph{3D model generation.}
In the main paper, we use a text-to-3D generator TRELLIS \cite{xiang2025structured} to produce high quality meshes that follow $p^{\mathrm{obj}}$.
The additional examples given in the supplement come from the RefRef dataset \cite{yin2025refref}, which provide clean and watertight meshes.
However, we have verified that TRELLIS can also generate suitable meshes for the associated prompts. 
An example is given in~\cref{fig:prompt23D}.

\paragraph{Object pose estimation.}
To place the transparent object into the scene, 
we apply SAM 3 \cite{carion2025sam3segmentconcepts} to segment the horizontal surface referred to in the text prompt, and place the bottom center of the 3D model generated by TRELLIS~\cite{xiang2025structured} at the center of the surface.

\section{Extended Experimental Results}
\label{sec:extended_results}

\subsection{Qualitative Results}
\begin{figure*}[!t]
    \centering

    \begin{minipage}[c]{0.025\linewidth}\centering
        \rotatebox{90}{Reference}
    \end{minipage}%
    \begin{subfigure}[]{0.237\linewidth}\centering
        \includegraphics[width=\linewidth]{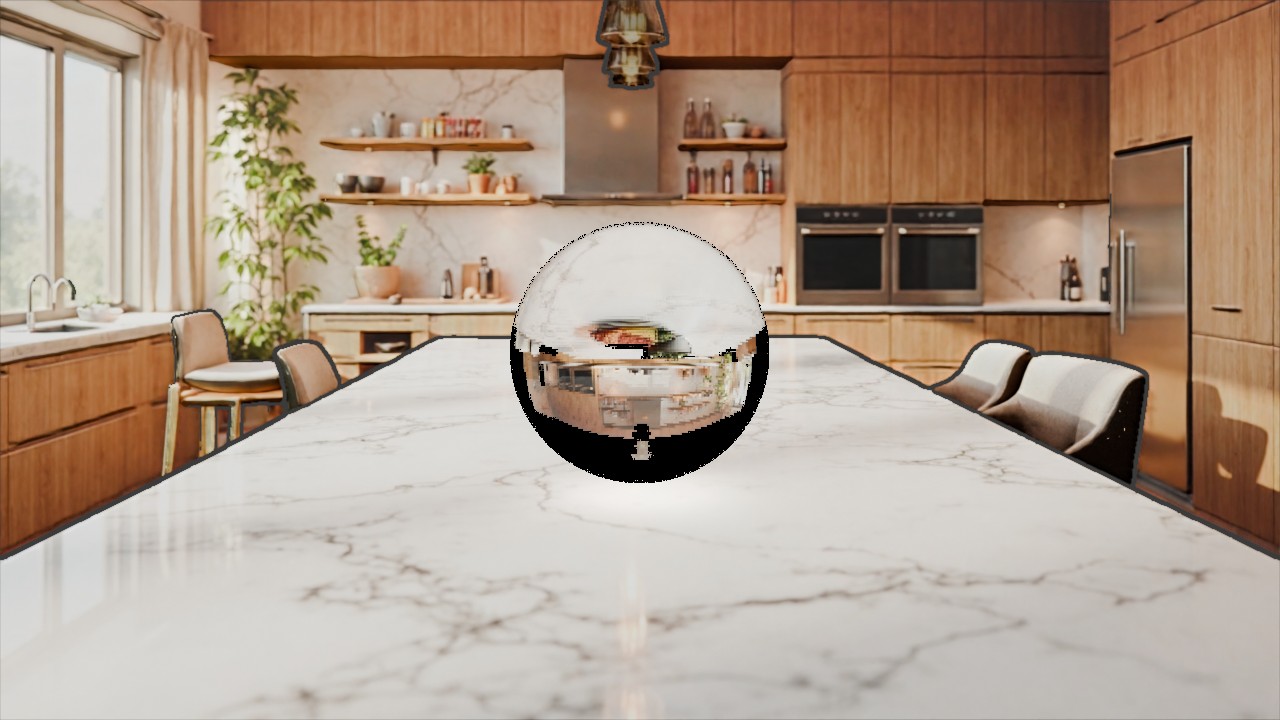}
    \end{subfigure}\hspace{2pt}%
    \begin{subfigure}[]{0.237\linewidth}\centering
        \includegraphics[width=\linewidth]{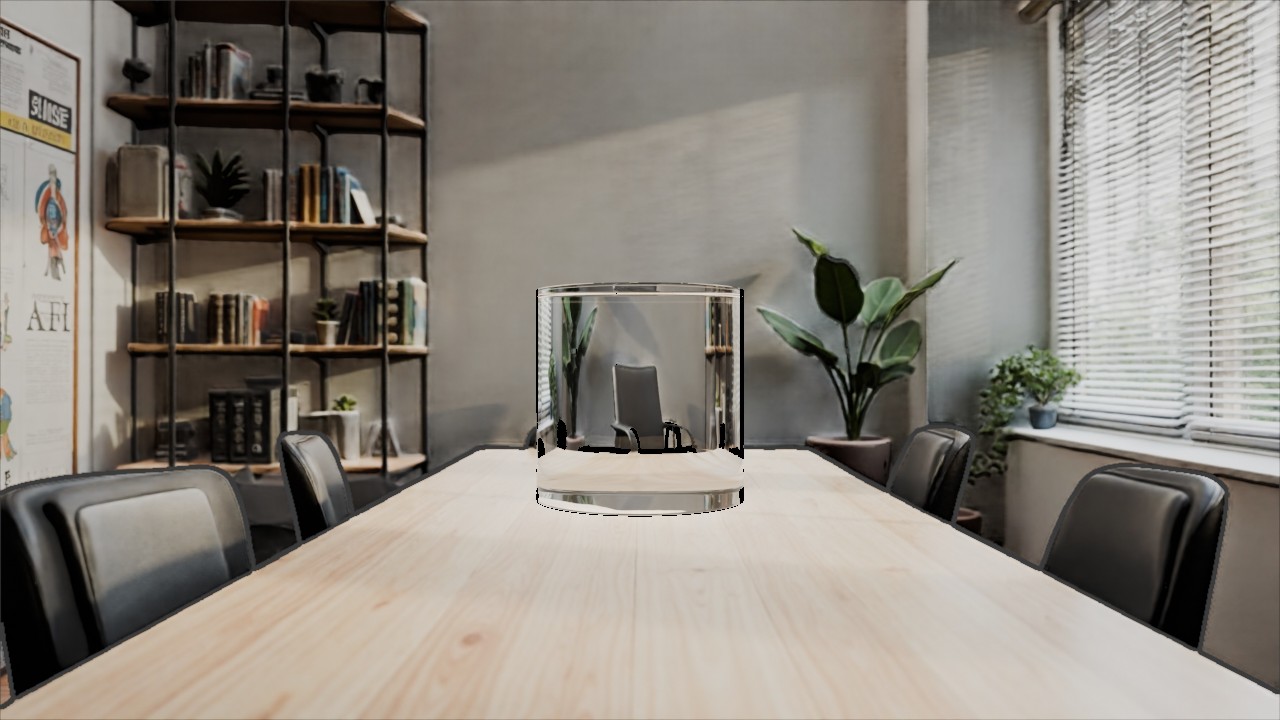}
    \end{subfigure}\hspace{2pt}%
    \begin{subfigure}[]{0.237\linewidth}\centering
        \includegraphics[width=\linewidth]{figures/suppl/generated_other_shapes/artroom_9577/artroom_9577_maskedgt.jpg}
    \end{subfigure}\hspace{2pt}%
    \begin{subfigure}[]{0.237\linewidth}\centering
        \includegraphics[width=\linewidth]{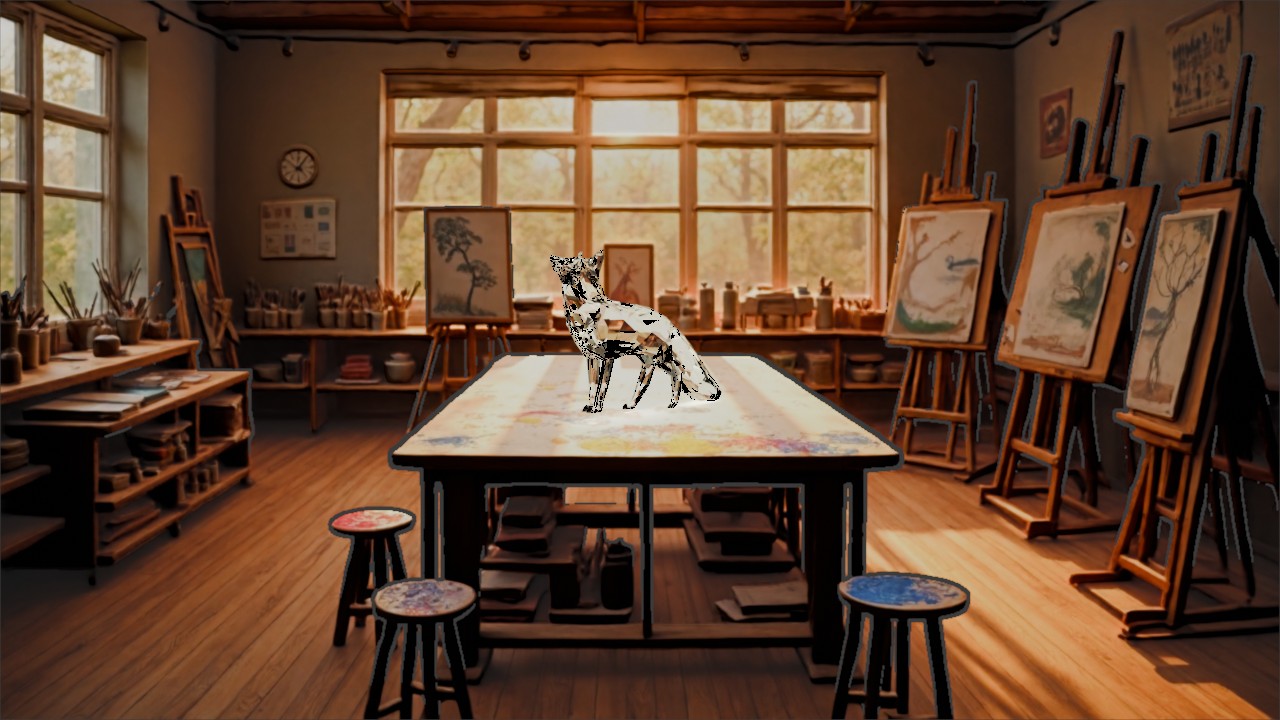}
    \end{subfigure}\\[2pt]

    \begin{minipage}[c]{0.025\linewidth}\centering
        \rotatebox{90}{Ours}
    \end{minipage}%
    \begin{subfigure}[]{0.237\linewidth}\centering
        \includegraphics[width=\linewidth]{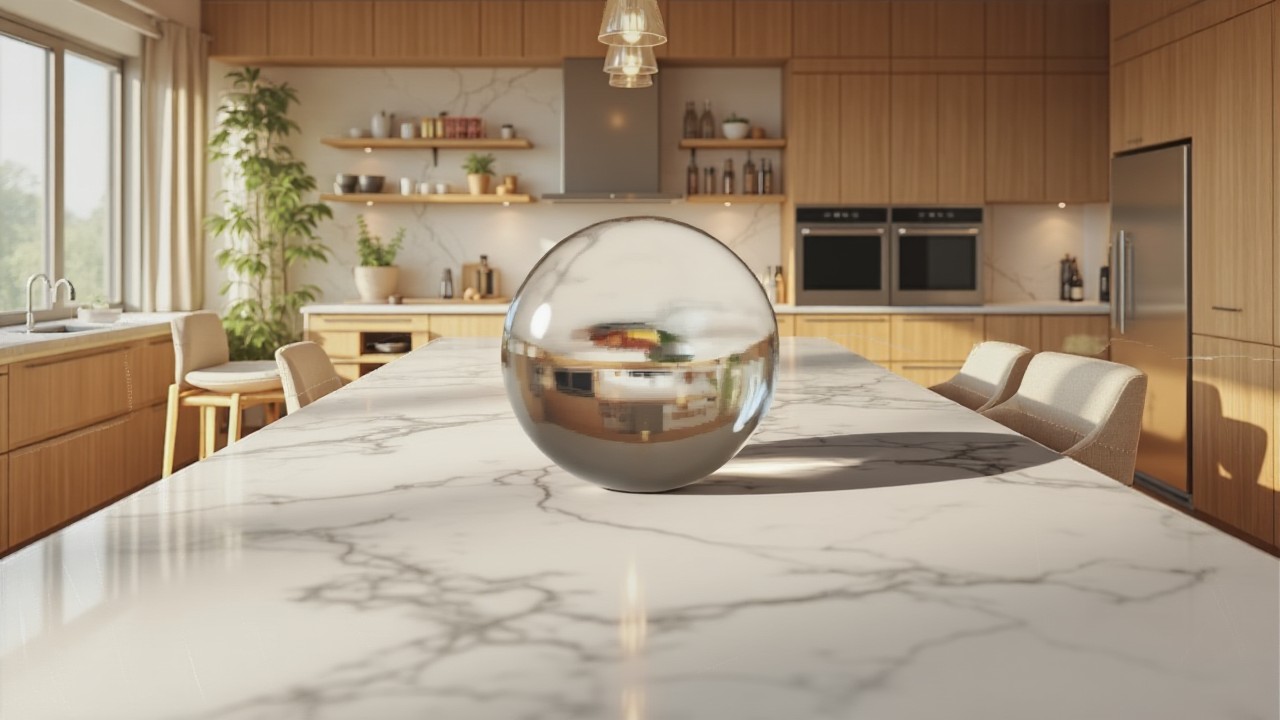}
    \end{subfigure}\hspace{2pt}%
    \begin{subfigure}[]{0.237\linewidth}\centering
        \includegraphics[width=\linewidth]{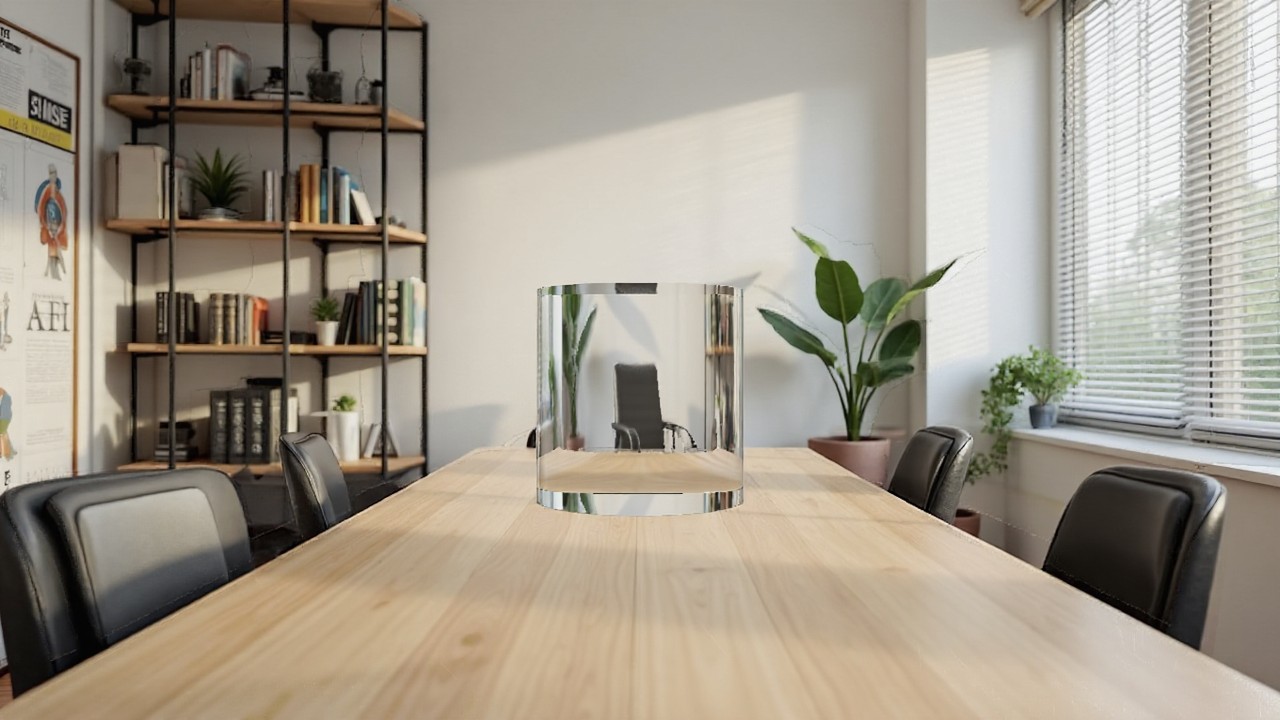}
    \end{subfigure}\hspace{2pt}%
    \begin{subfigure}[]{0.237\linewidth}\centering
        \includegraphics[width=\linewidth]{figures/suppl/generated_other_shapes/artroom_9577/artroom_9577_main.jpg}
    \end{subfigure}\hspace{2pt}%
    \begin{subfigure}[]{0.237\linewidth}\centering
        \includegraphics[width=\linewidth]{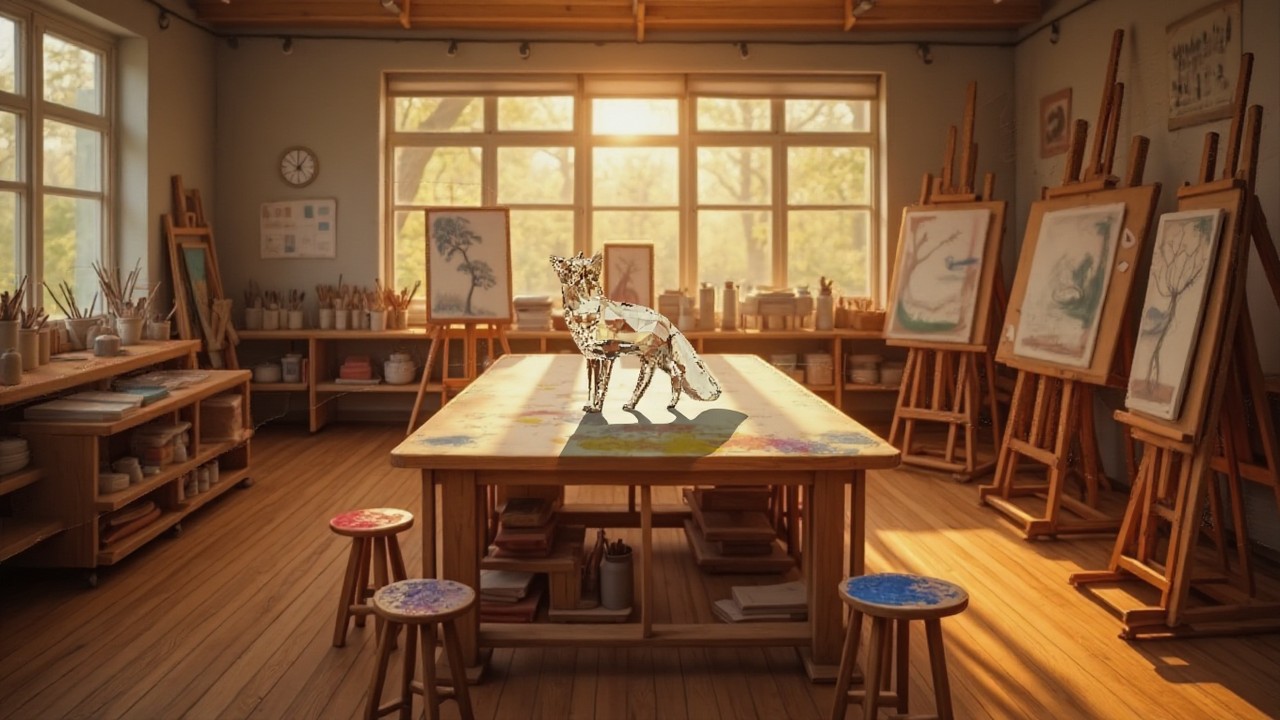}
    \end{subfigure}\\[2pt]

    \begin{minipage}[c]{0.025\linewidth}\centering
        \rotatebox{90}{FLUX Inpaint}
    \end{minipage}%
    \begin{subfigure}[]{0.237\linewidth}\centering
        \includegraphics[width=\linewidth]{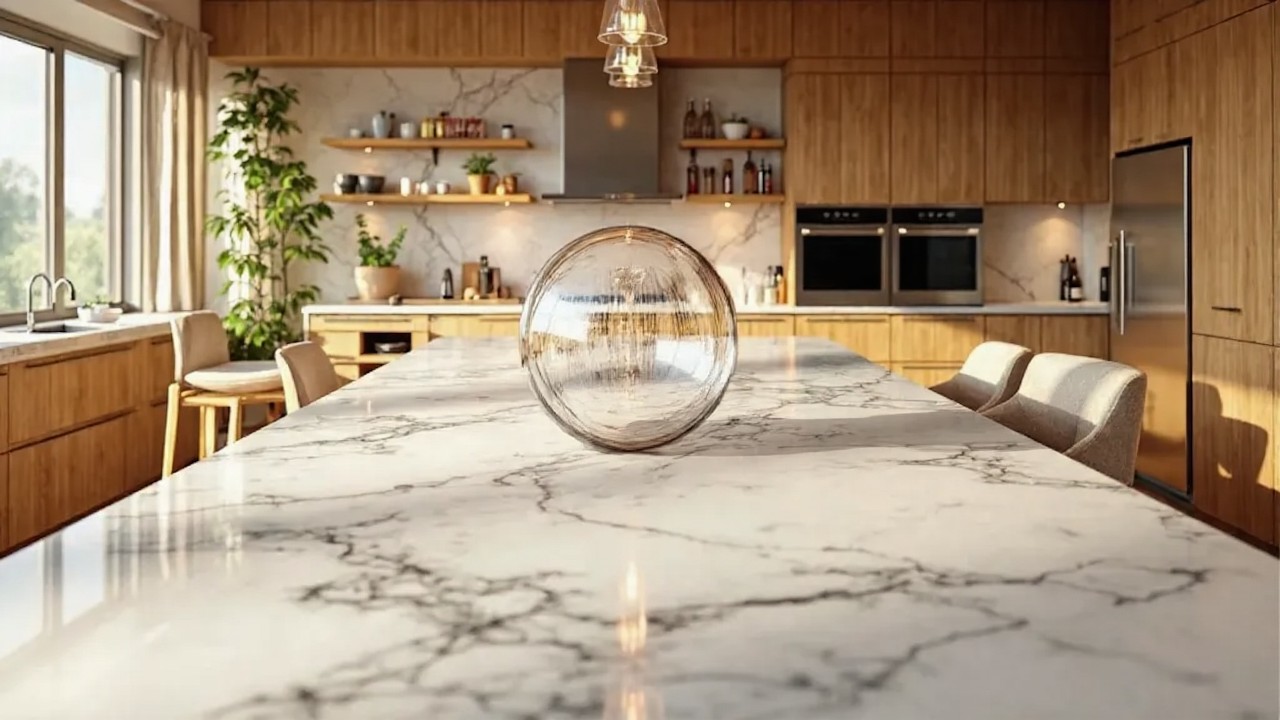}
    \end{subfigure}\hspace{2pt}%
    \begin{subfigure}[]{0.237\linewidth}\centering
        \includegraphics[width=\linewidth]{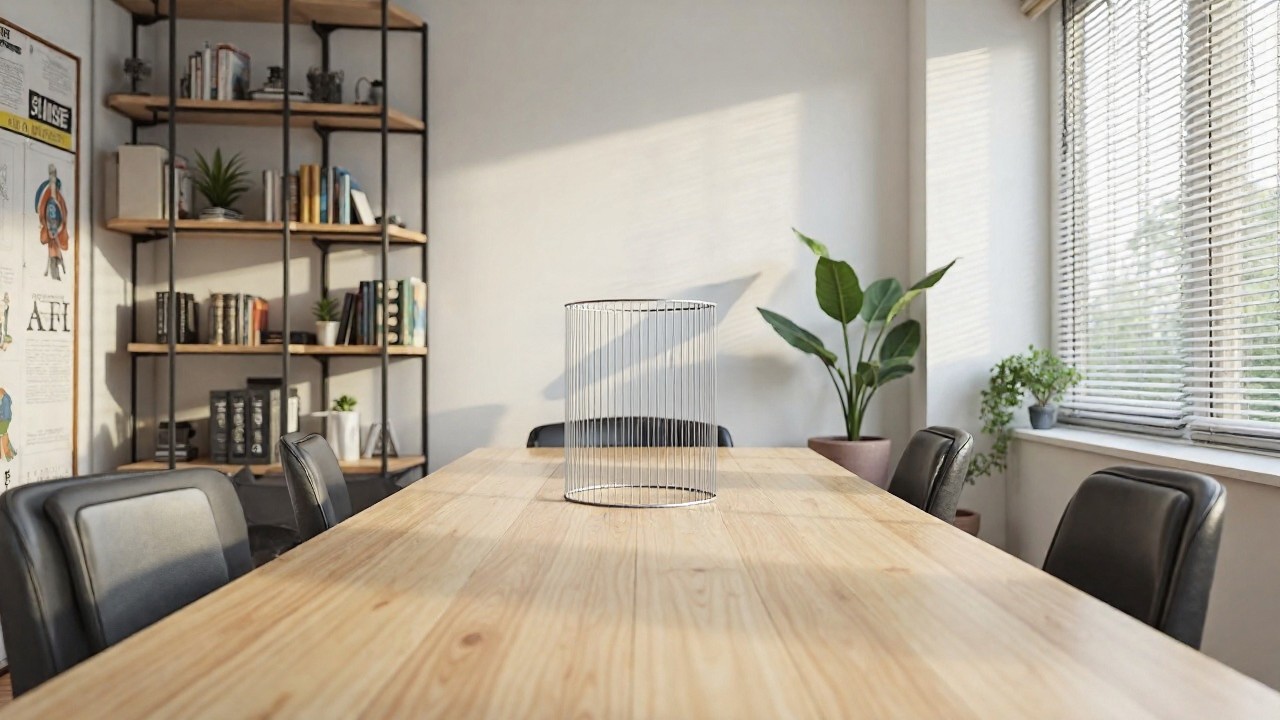}
    \end{subfigure}\hspace{2pt}%
    \begin{subfigure}[]{0.237\linewidth}\centering
        \includegraphics[width=\linewidth]{figures/suppl/generated_other_shapes/artroom_9577/artroom_9577_flux_fill.jpg}
    \end{subfigure}\hspace{2pt}%
    \begin{subfigure}[]{0.237\linewidth}\centering
        \includegraphics[width=\linewidth]{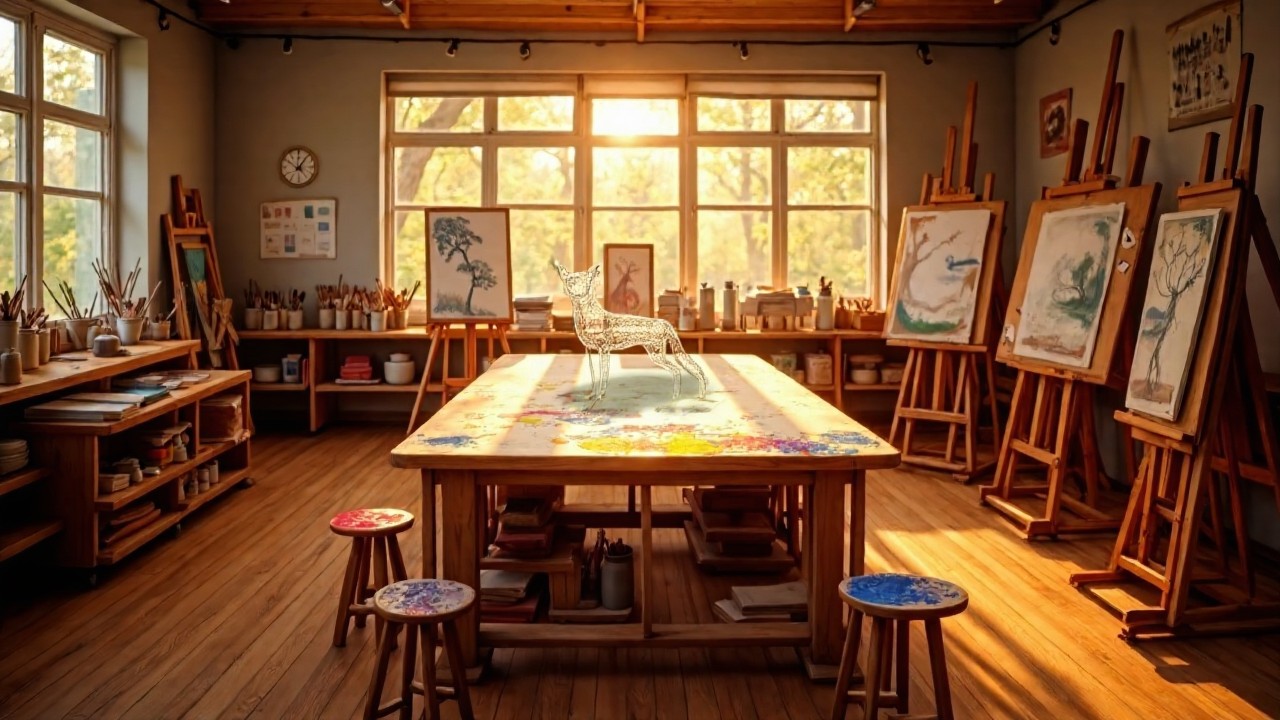}
    \end{subfigure}\\[2pt]

    \begin{minipage}[c]{0.025\linewidth}\centering
        \rotatebox{90}{FLUX-dev}
    \end{minipage}%
    \begin{subfigure}[]{0.237\linewidth}\centering
        \includegraphics[width=\linewidth]{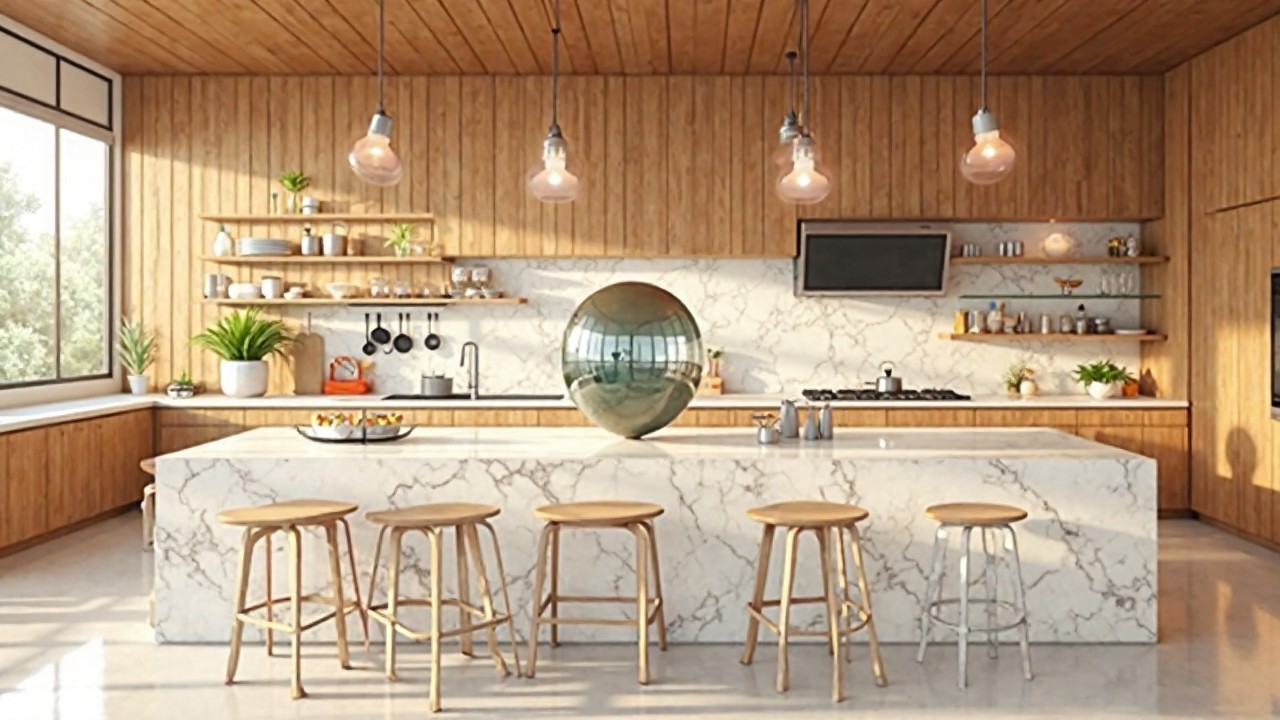}
    \end{subfigure}\hspace{2pt}%
    \begin{subfigure}[]{0.237\linewidth}\centering
        \includegraphics[width=\linewidth]{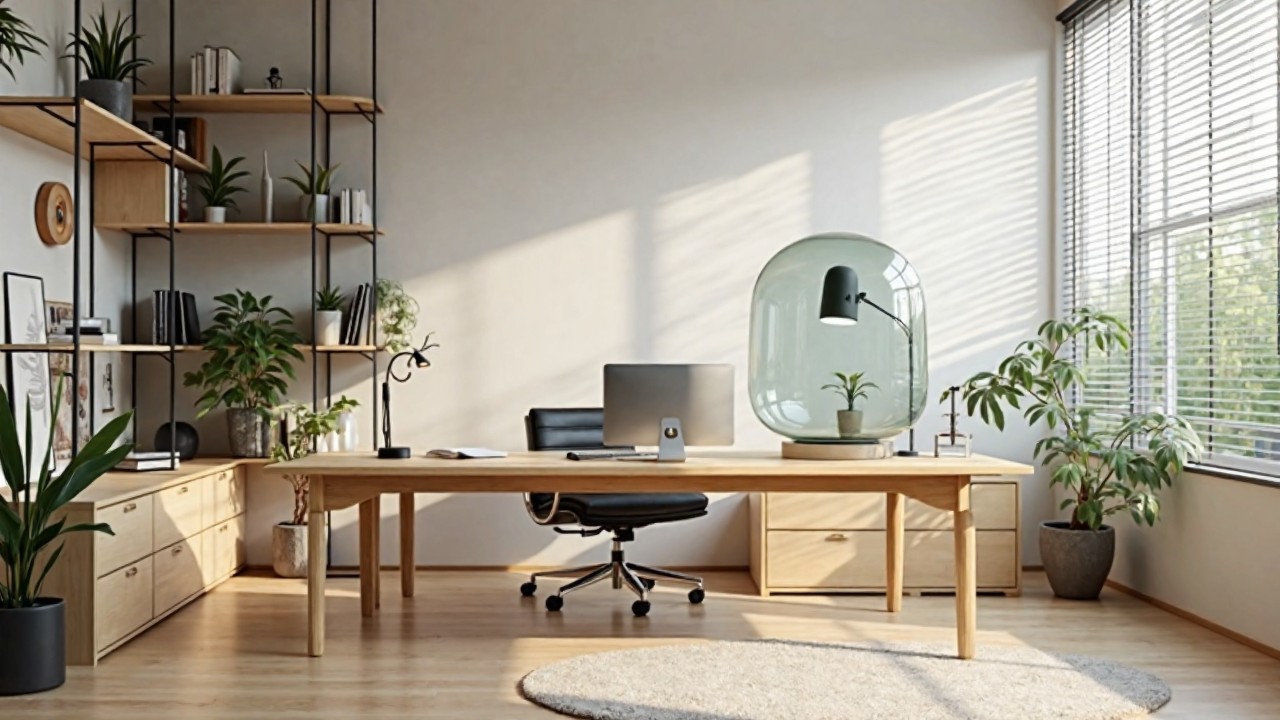}
    \end{subfigure}\hspace{2pt}%
    \begin{subfigure}[]{0.237\linewidth}\centering
        \includegraphics[width=\linewidth]{figures/suppl/generated_other_shapes/artroom_9577/artroom_9577_flux.jpg}
    \end{subfigure}\hspace{2pt}%
    \begin{subfigure}[]{0.237\linewidth}\centering
        \includegraphics[width=\linewidth]{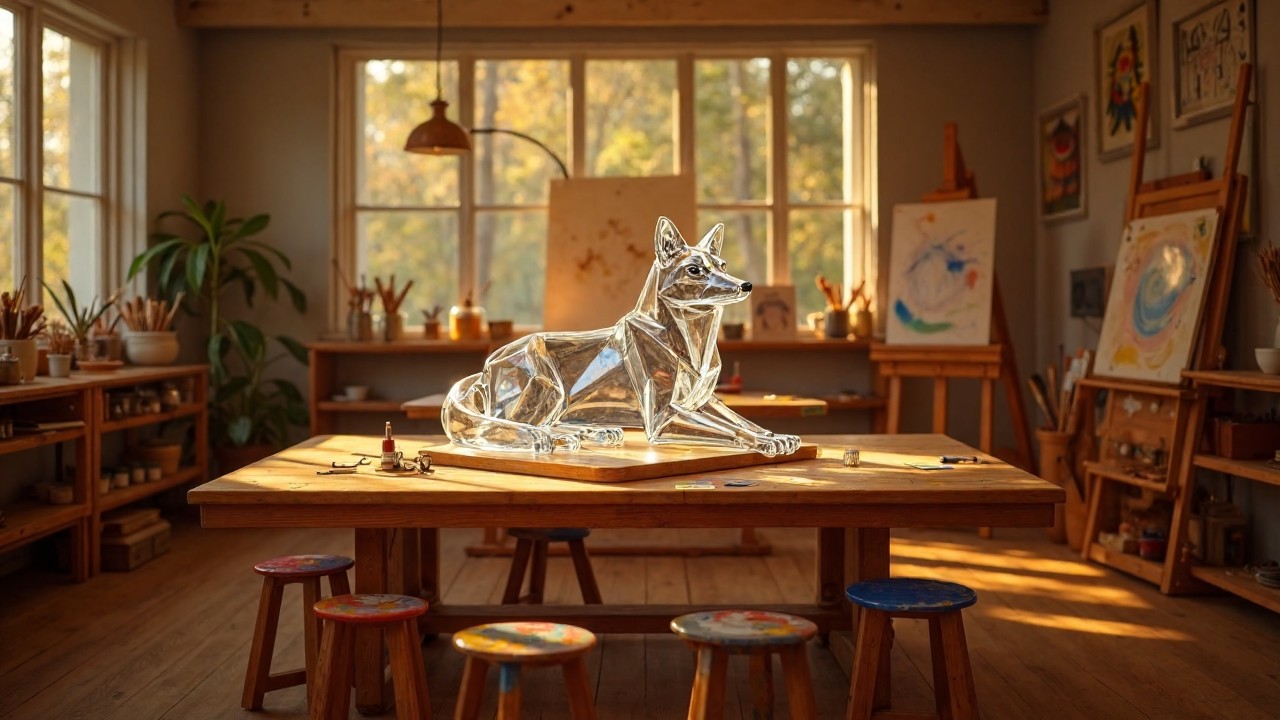}
    \end{subfigure}\\[2pt]

    \begin{minipage}[c]{0.025\linewidth}\centering
        \rotatebox{90}{FLUX.2-dev}
    \end{minipage}%
    \begin{subfigure}[]{0.237\linewidth}\centering
        \includegraphics[width=\linewidth]{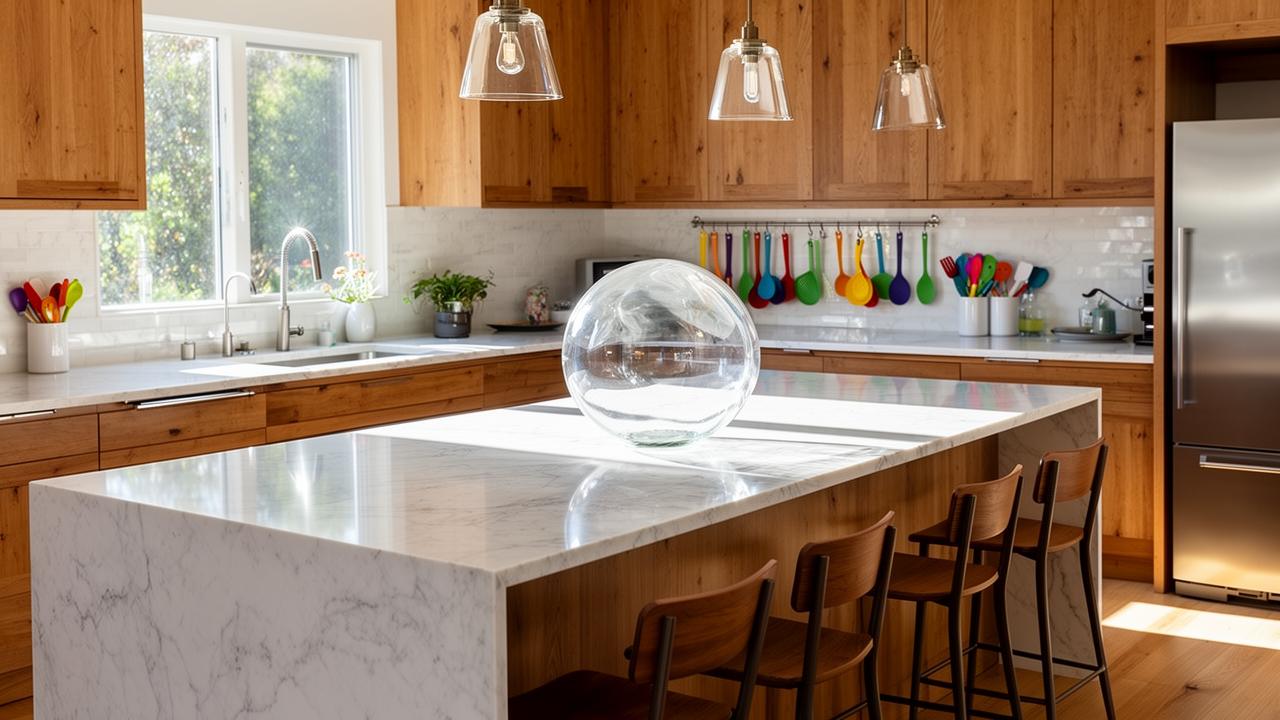}
    \end{subfigure}\hspace{2pt}%
    \begin{subfigure}[]{0.237\linewidth}\centering
        \includegraphics[width=\linewidth]{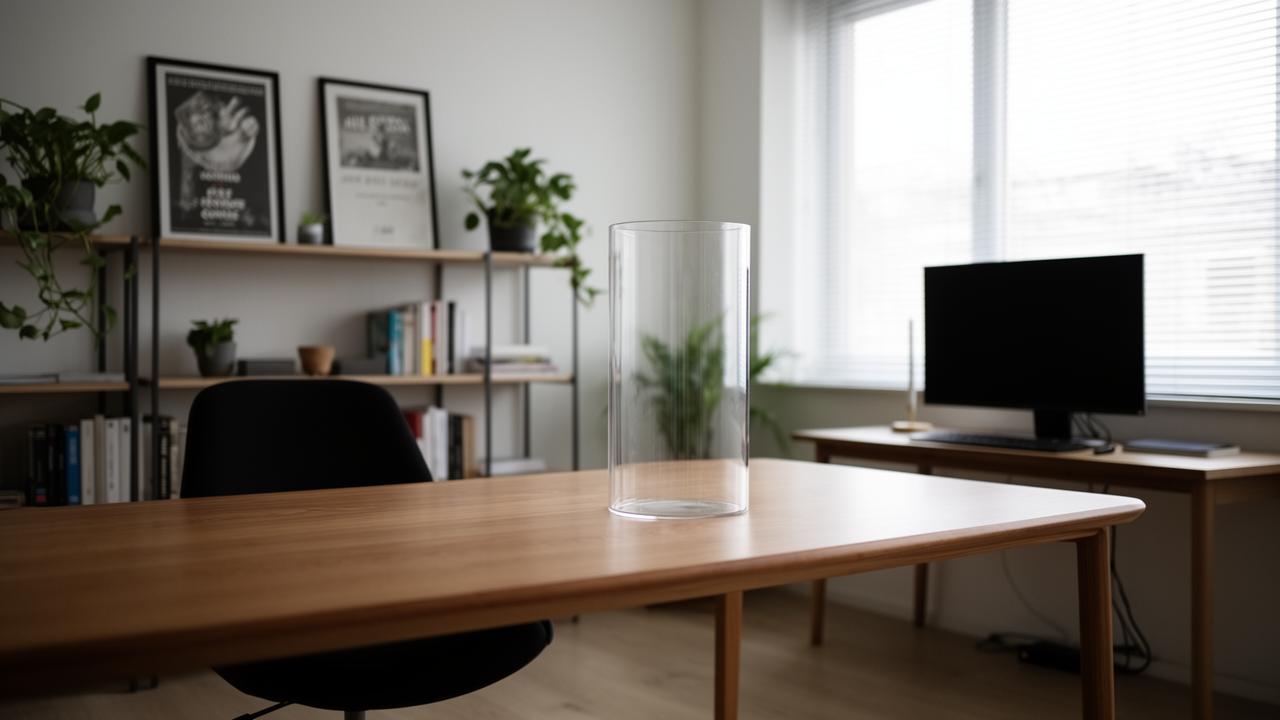}
    \end{subfigure}\hspace{2pt}%
    \begin{subfigure}[]{0.237\linewidth}\centering
        \includegraphics[width=\linewidth]{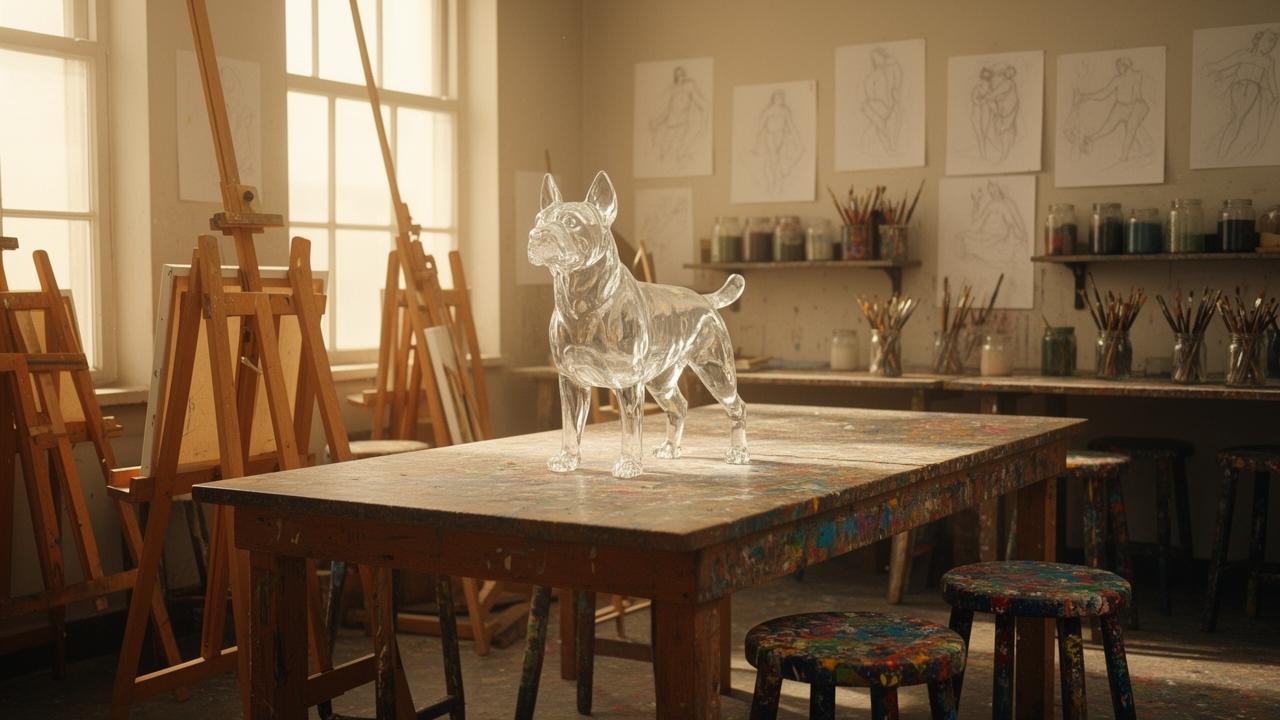}
    \end{subfigure}\hspace{2pt}%
    \begin{subfigure}[]{0.237\linewidth}\centering
        \includegraphics[width=\linewidth]{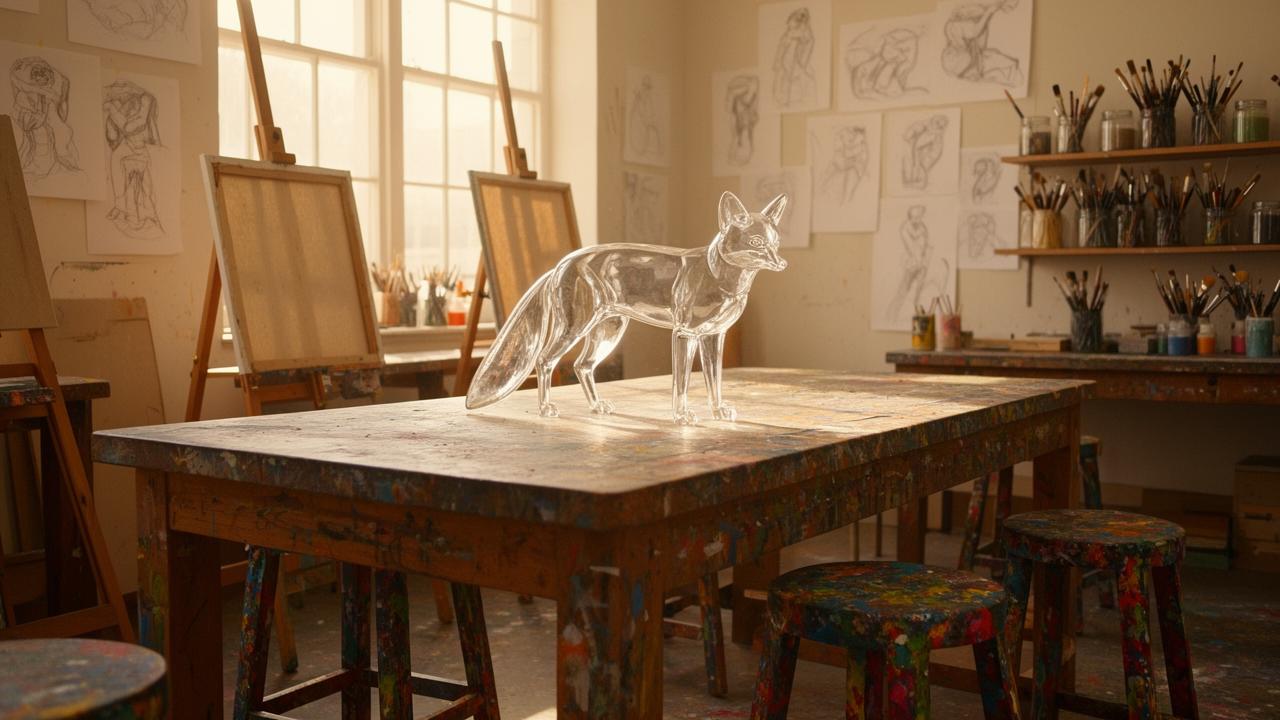}
    \end{subfigure}\\[2pt]

    \begin{minipage}[c]{0.025\linewidth}\centering
        \rotatebox{90}{SD3.5}
    \end{minipage}%
    \begin{subfigure}[]{0.237\linewidth}\centering
        \includegraphics[width=\linewidth]{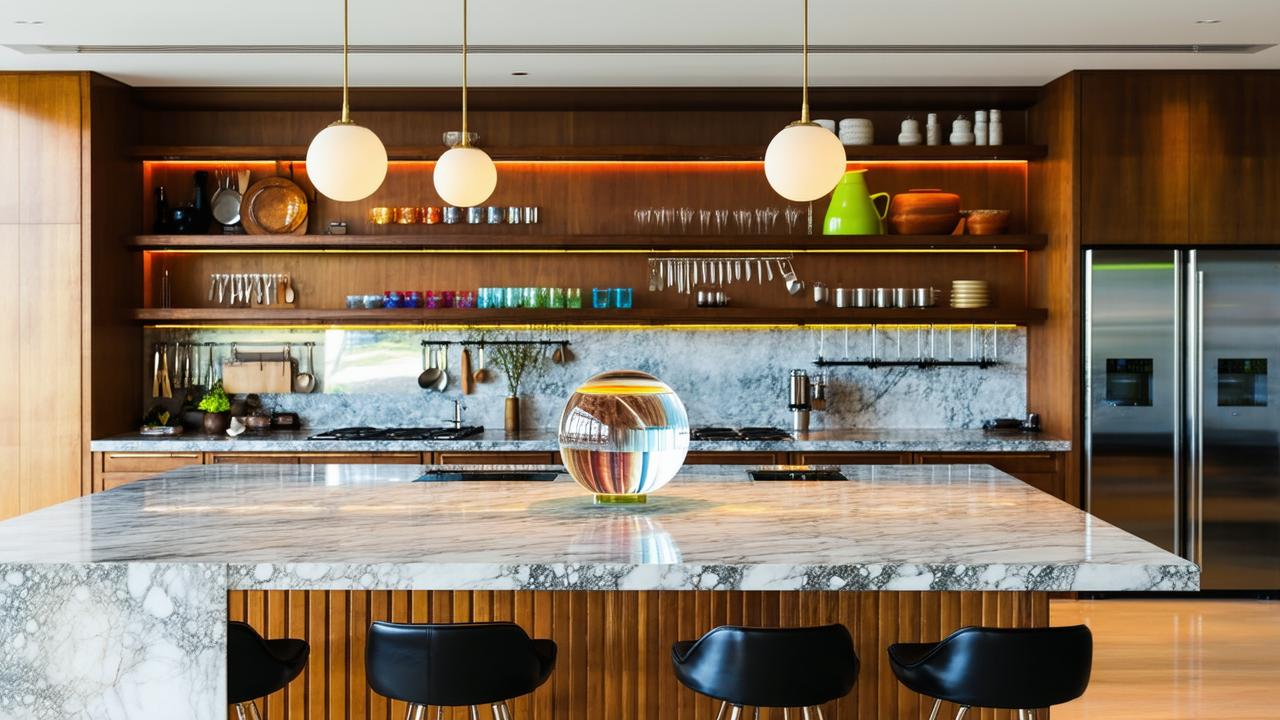}
    \end{subfigure}\hspace{2pt}%
    \begin{subfigure}[]{0.237\linewidth}\centering
        \includegraphics[width=\linewidth]{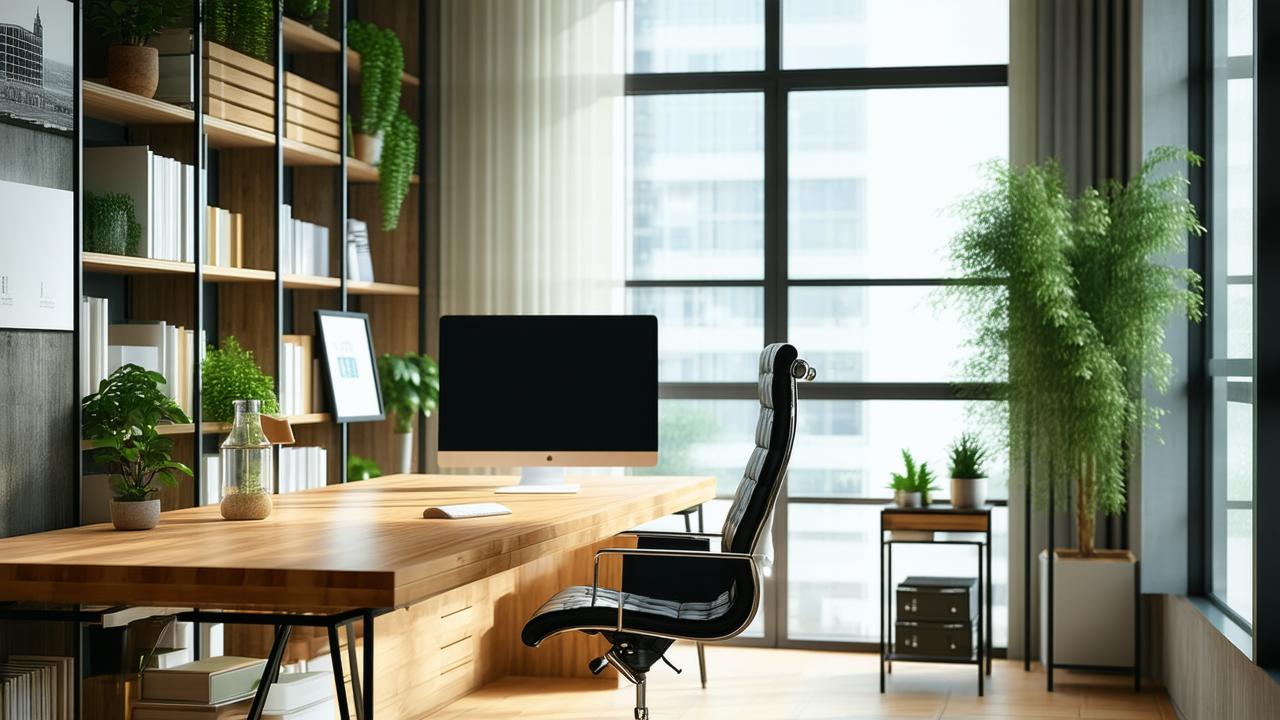}
    \end{subfigure}\hspace{2pt}%
    \begin{subfigure}[]{0.237\linewidth}\centering
        \includegraphics[width=\linewidth]{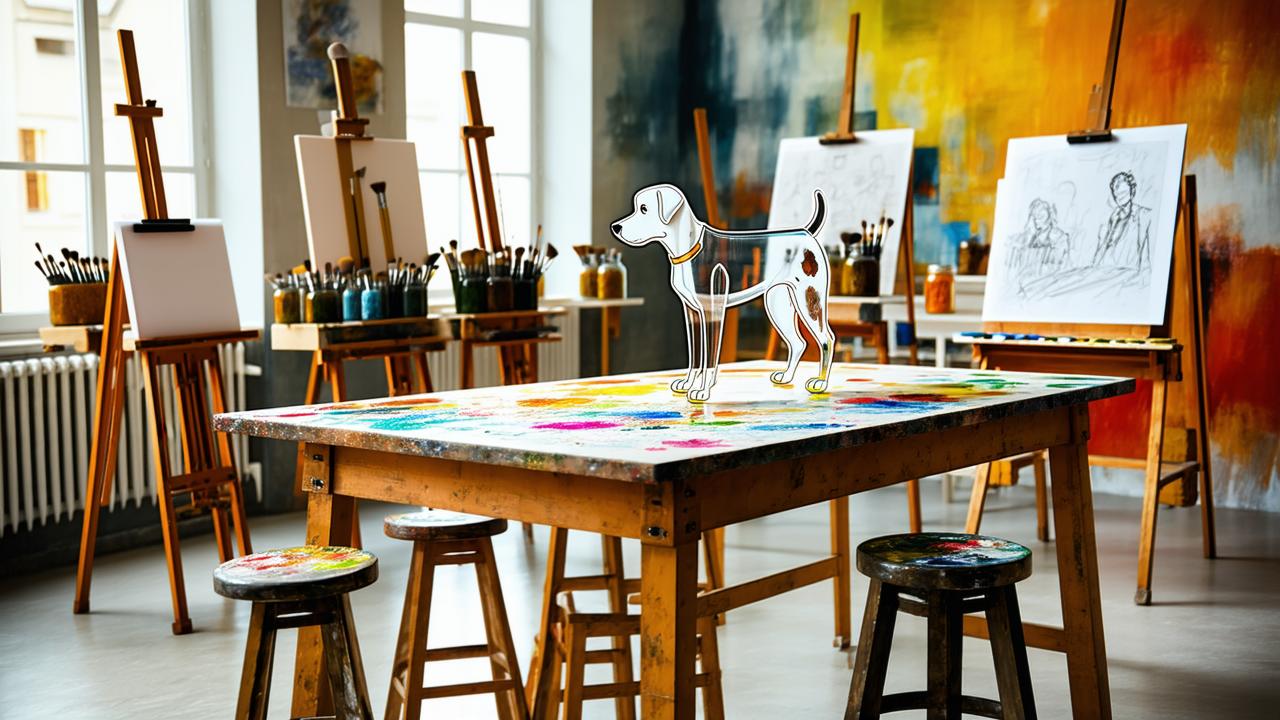}
    \end{subfigure}\hspace{2pt}%
    \begin{subfigure}[]{0.237\linewidth}\centering
        \includegraphics[width=\linewidth]{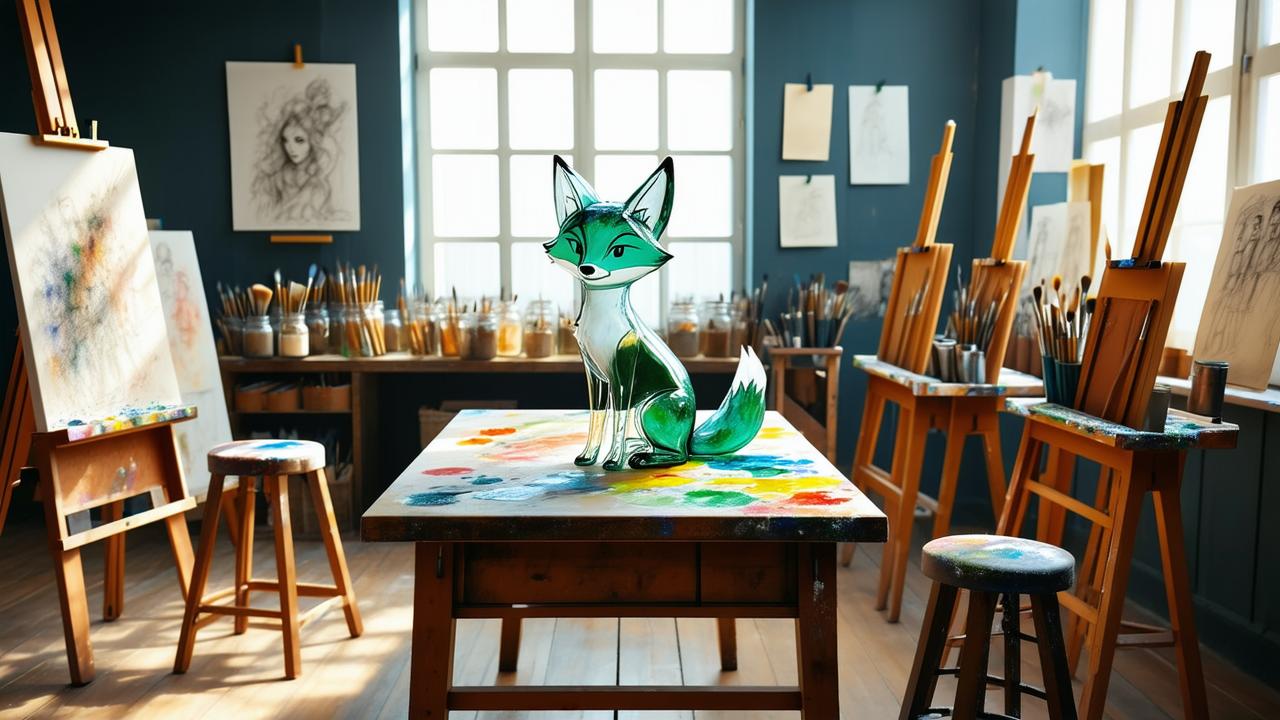}
    \end{subfigure}\\[2pt]

    \begin{minipage}[c]{0.025\linewidth}\centering
        \rotatebox{90}{Qwen}
    \end{minipage}%
    \begin{subfigure}[]{0.237\linewidth}\centering
        \includegraphics[width=\linewidth]{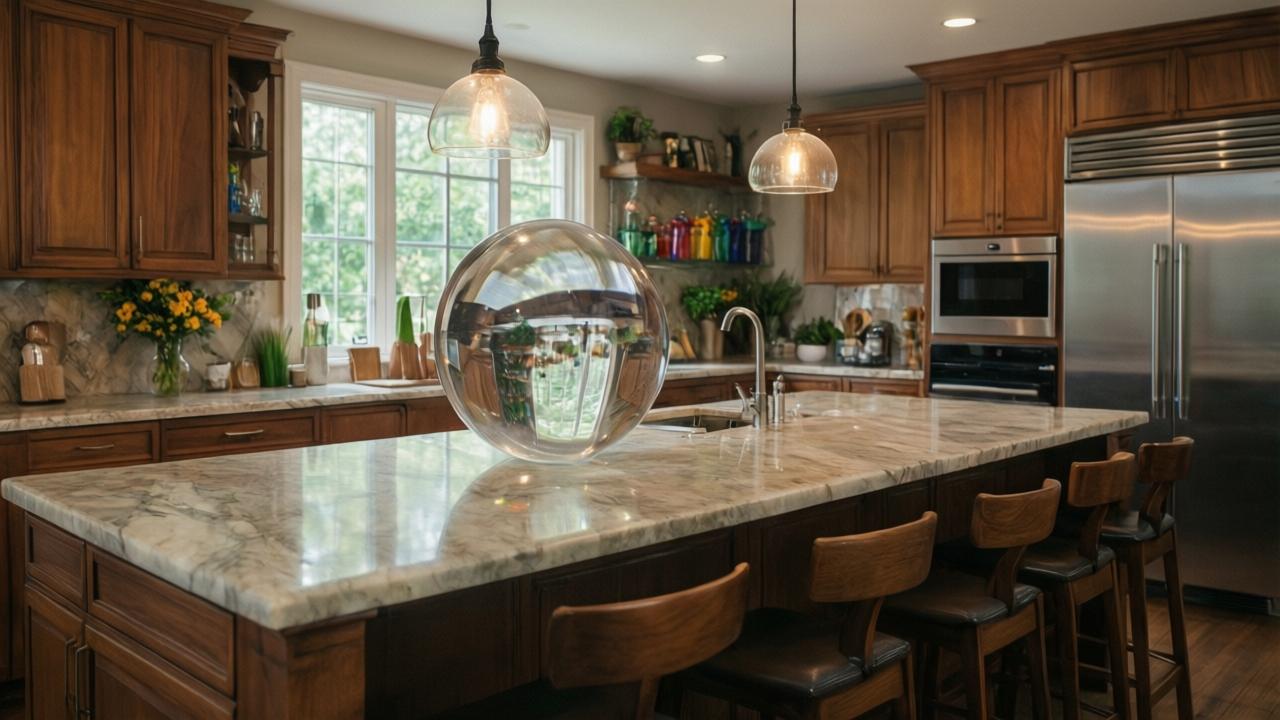}
    \end{subfigure}\hspace{2pt}%
    \begin{subfigure}[]{0.237\linewidth}\centering
        \includegraphics[width=\linewidth]{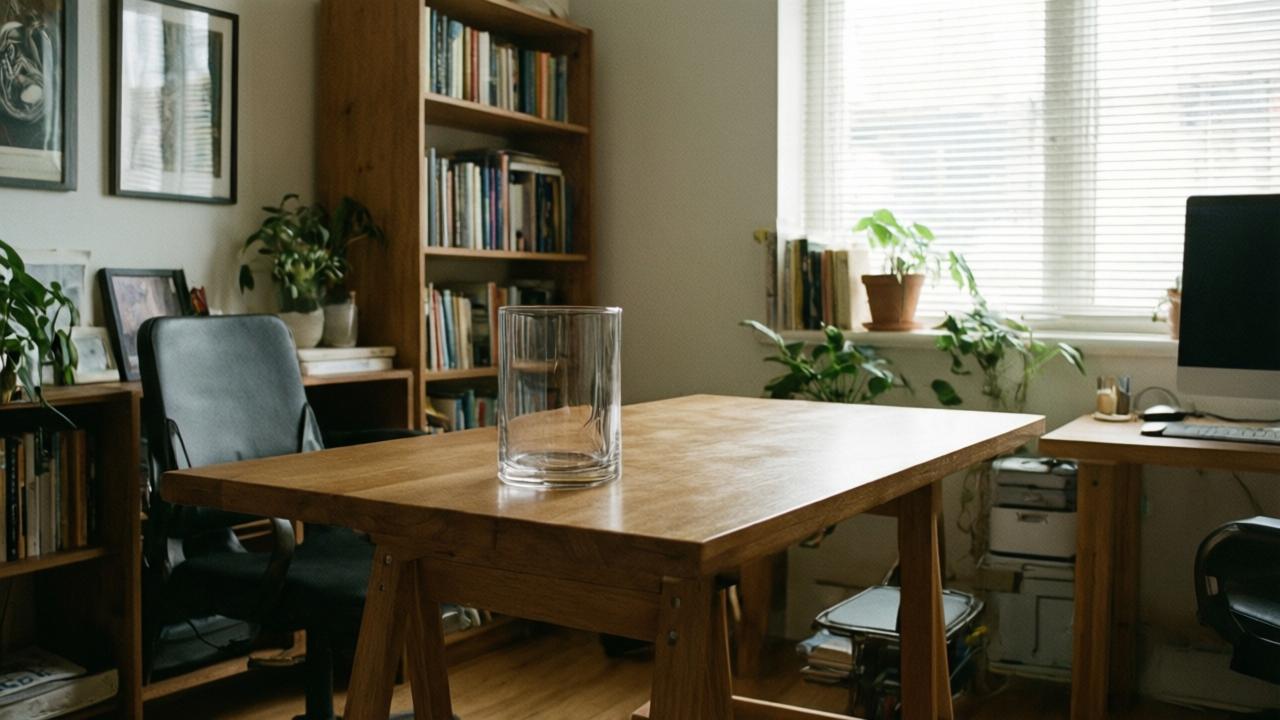}
    \end{subfigure}\hspace{2pt}%
    \begin{subfigure}[]{0.237\linewidth}\centering
        \includegraphics[width=\linewidth]{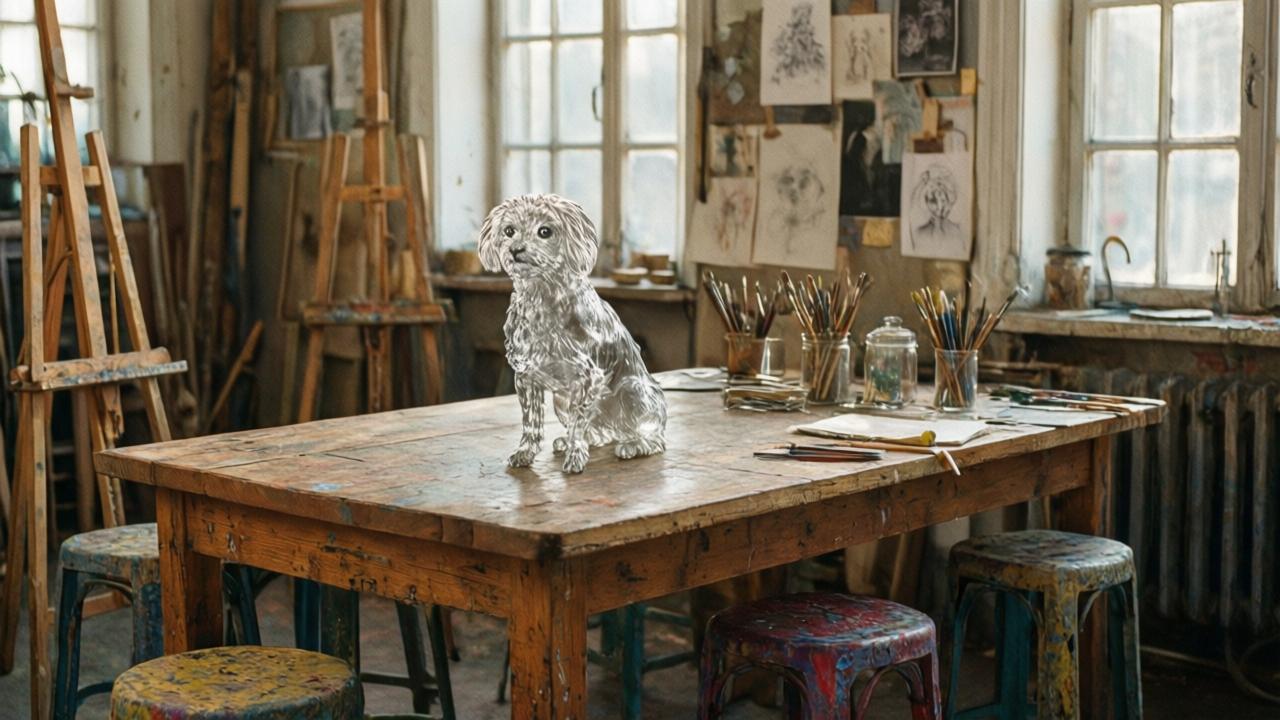}
    \end{subfigure}\hspace{2pt}%
    \begin{subfigure}[]{0.237\linewidth}\centering
        \includegraphics[width=\linewidth]{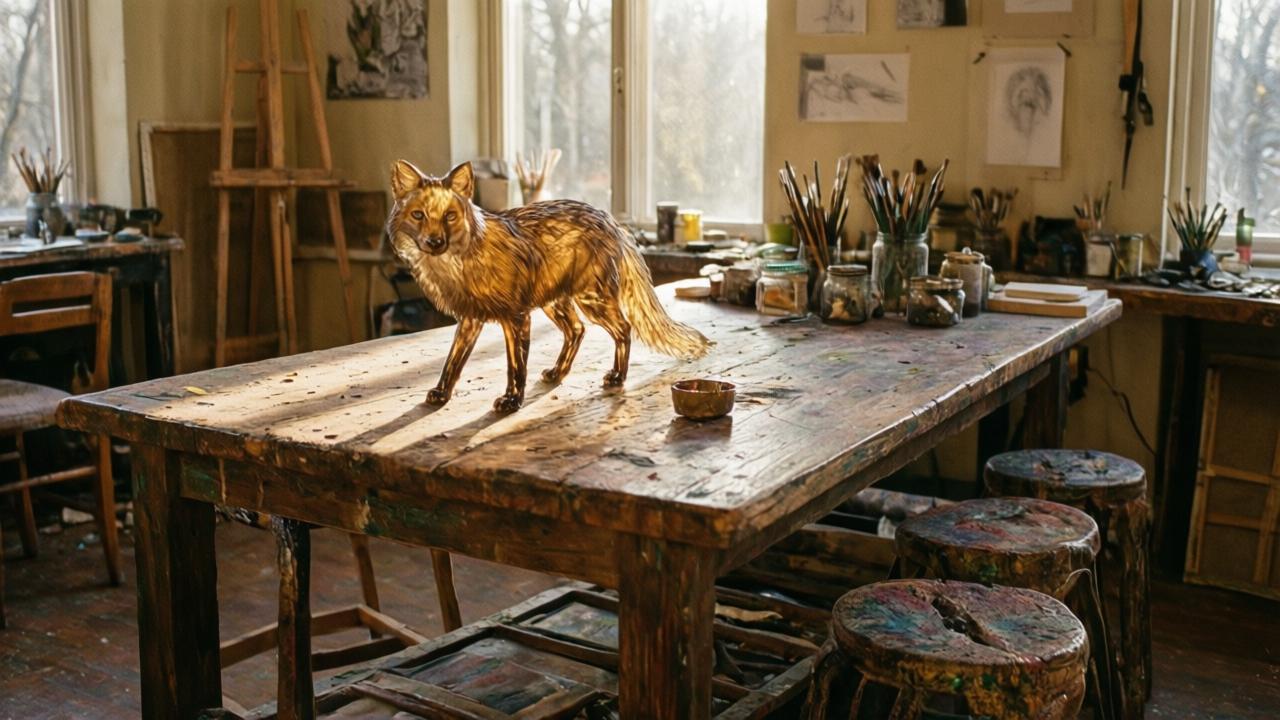}
    \end{subfigure}
    \caption{Additional qualitative results of our method against FLUX-dev~\cite{labs2025flux1kontextflowmatching}, FLUX.2-dev~\cite{flux-2-2025}, Qwen-Image~\cite{wu2025qwen}, Stable Diffusion 3.5 (Large)~\cite{sd3}, and FLUX-based inpainting model~\cite{labs2025flux1kontextflowmatching}, on a diversity of shapes.
    These examples illustrate the robustness of our method across diverse object geometries.
    This figure is the first part of a two page layout.}
    \label{fig:more_objects1}
\end{figure*}
\begin{figure*}[!t]
    \centering

    \makebox[0.025\linewidth]{} %
    \begin{minipage}[]{0.237\linewidth}\centering
        Blender Reference
    \end{minipage}\hspace{2pt}%
    \begin{minipage}[]{0.237\linewidth}\centering
        Snellcaster (Ours)
    \end{minipage}\hspace{2pt}%
    \begin{minipage}[]{0.237\linewidth}\centering
        FLUX Inpainting
    \end{minipage}\hspace{2pt}%
    \begin{minipage}[]{0.237\linewidth}\centering
        FLUX-dev
    \end{minipage}\\[1.0pt]
    \begin{minipage}[c]{0.025\linewidth}\centering
        \rotatebox{90}{Sculpture}
    \end{minipage}%
    \begin{subfigure}[]{0.237\linewidth}\centering
        \includegraphics[width=\linewidth]{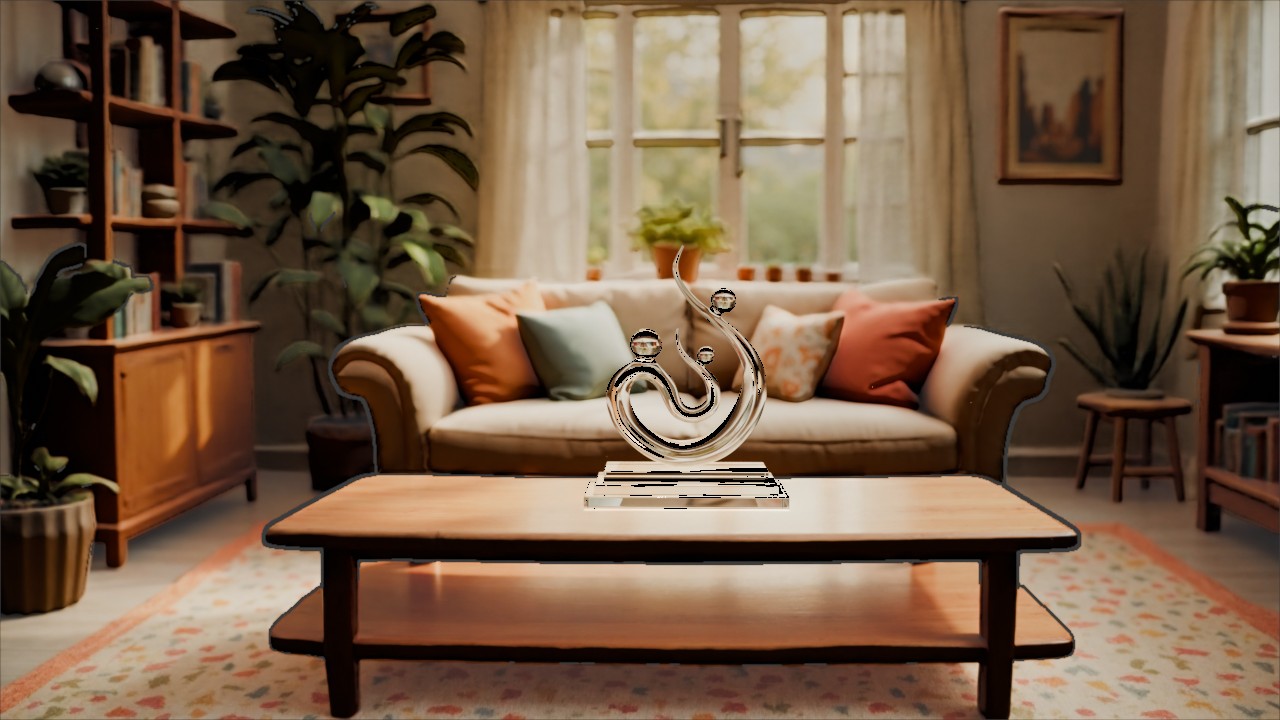}
    \end{subfigure}\hspace{2pt}%
    \begin{subfigure}[]{0.237\linewidth}\centering
        \includegraphics[width=\linewidth]{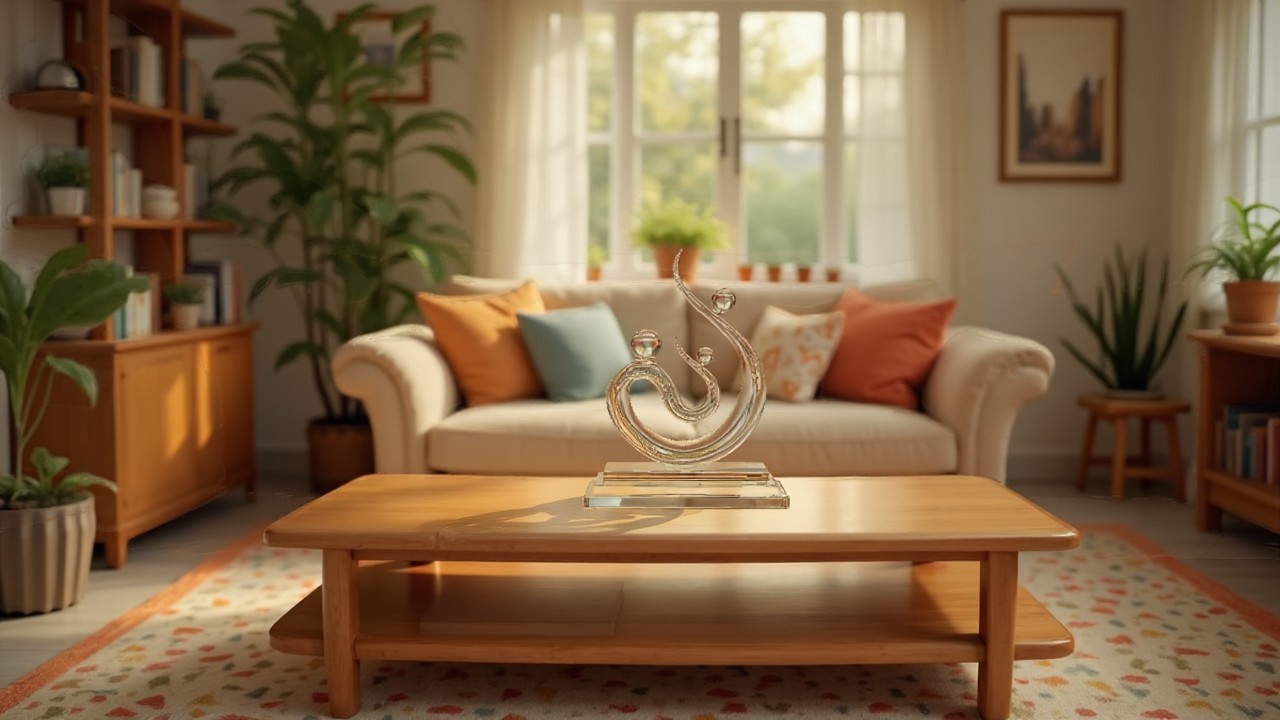}
    \end{subfigure}\hspace{2pt}%
    \begin{subfigure}[]{0.237\linewidth}\centering
        \includegraphics[width=\linewidth]{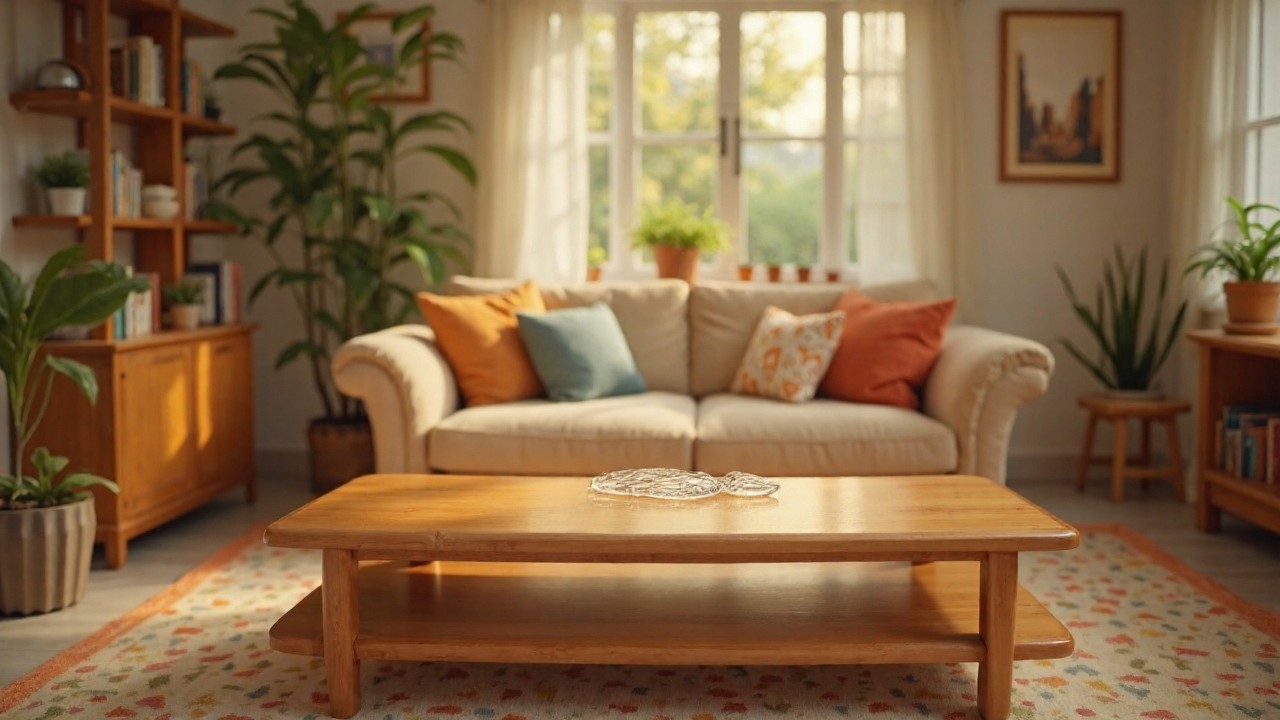}
    \end{subfigure}\hspace{2pt}%
    \begin{subfigure}[]{0.237\linewidth}\centering
        \includegraphics[width=\linewidth]{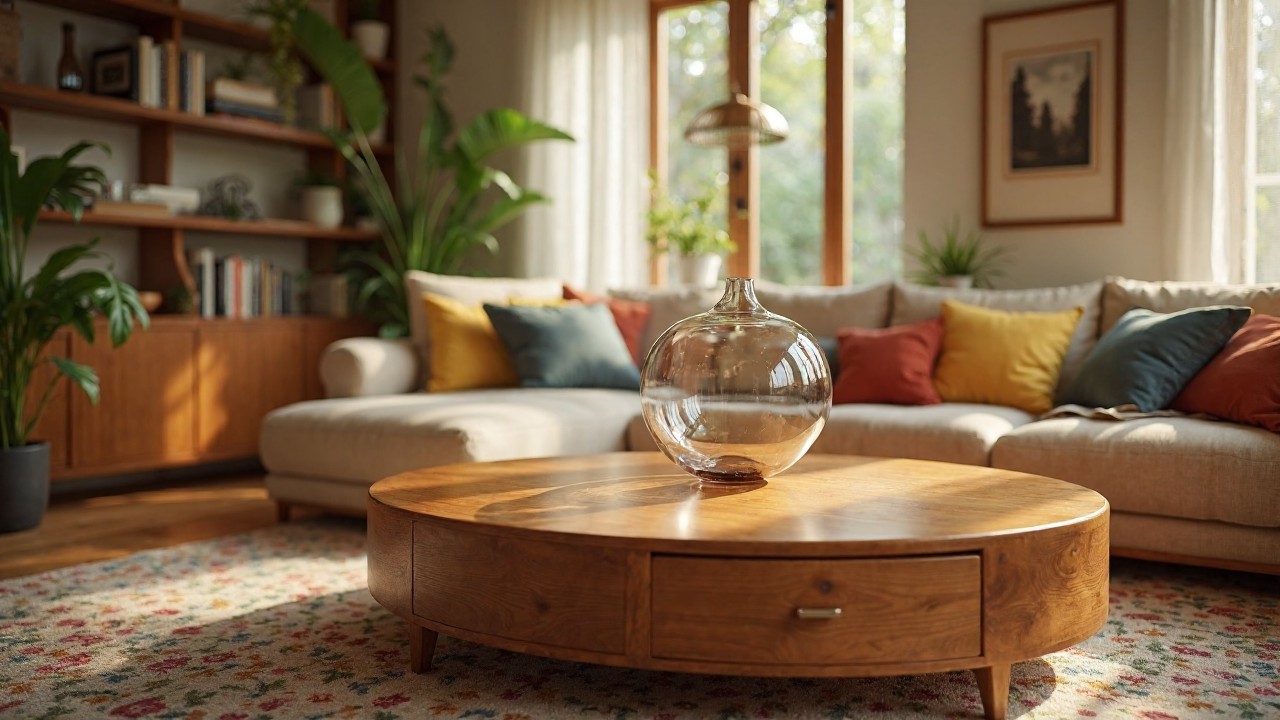}
    \end{subfigure}\\[2pt]

    \begin{minipage}[c]{0.025\linewidth}\centering
        \rotatebox{90}{Sculpture (inset)}
    \end{minipage}%
    \begin{subfigure}[]{0.237\linewidth}\centering
        \includegraphics[
            width=\linewidth,
            trim=550 200 450 220,
            clip
        ]{figures/suppl/generated_other_shapes/living_room_409/living_room_409_maskedgt.jpg}
    \end{subfigure}\hspace{2pt}%
    \begin{subfigure}[]{0.237\linewidth}\centering
        \includegraphics[
            width=\linewidth,
            trim=550 200 450 220,
            clip
        ]{figures/suppl/generated_other_shapes/living_room_409/living_room_409_main.jpg}
    \end{subfigure}\hspace{2pt}%
    \begin{subfigure}[]{0.237\linewidth}\centering
        \includegraphics[
            width=\linewidth,
            trim=550 200 450 220,
            clip
        ]{figures/suppl/generated_other_shapes/living_room_409/living_room_409_flux_fill.jpg}
    \end{subfigure}\hspace{2pt}%
    \begin{subfigure}[]{0.237\linewidth}\centering
        \includegraphics[
            width=\linewidth,
            trim=600 200 400 220,
            clip
        ]{figures/suppl/generated_other_shapes/living_room_409/living_room_409_flux.jpg}
    \end{subfigure}\\[4pt]

    \begin{minipage}[c]{0.025\linewidth}\centering
        \rotatebox{90}{Pyramid}
    \end{minipage}%
    \begin{subfigure}[]{0.237\linewidth}\centering
        \includegraphics[width=\linewidth]{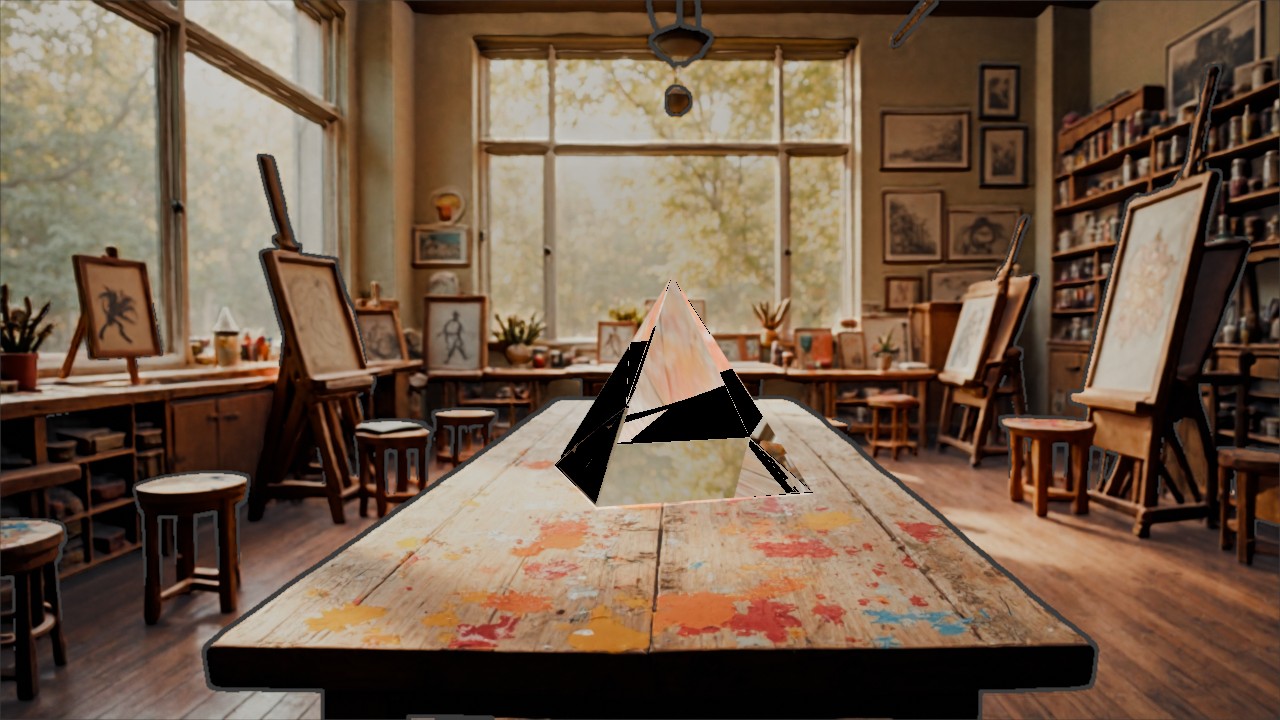}
    \end{subfigure}\hspace{2pt}%
    \begin{subfigure}[]{0.237\linewidth}\centering
        \includegraphics[width=\linewidth]{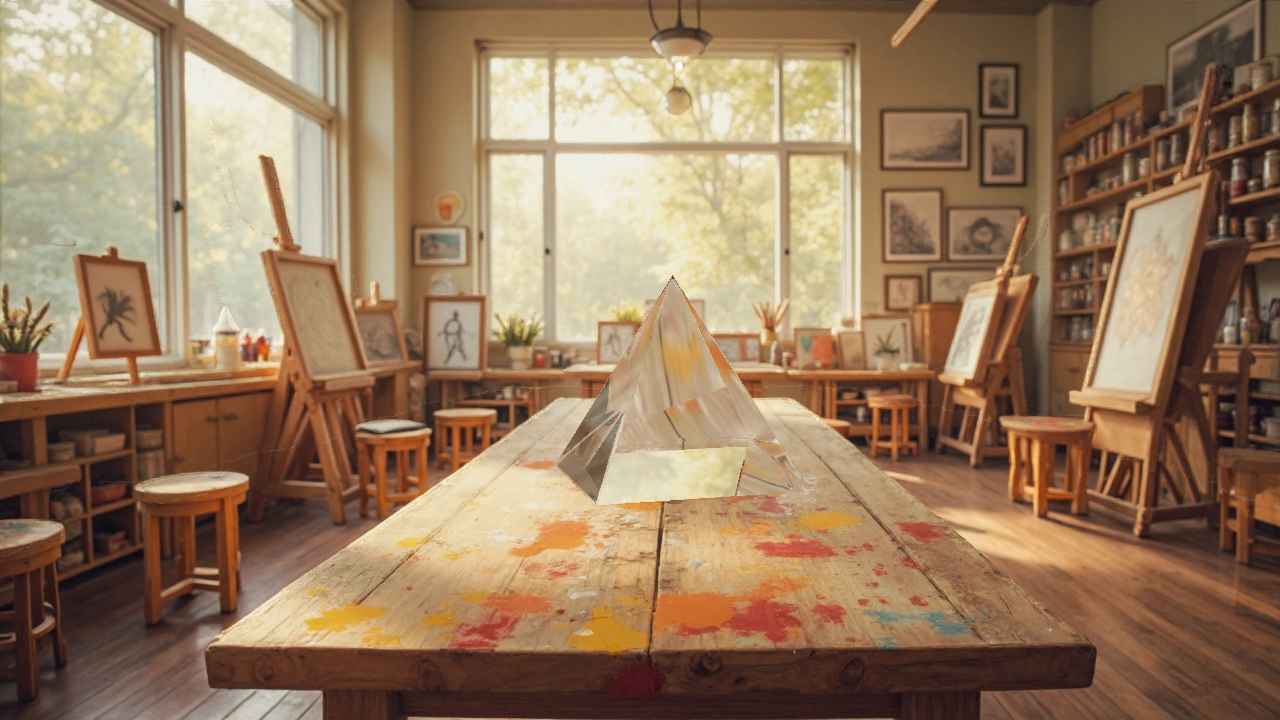}
    \end{subfigure}\hspace{2pt}%
    \begin{subfigure}[]{0.237\linewidth}\centering
        \includegraphics[width=\linewidth]{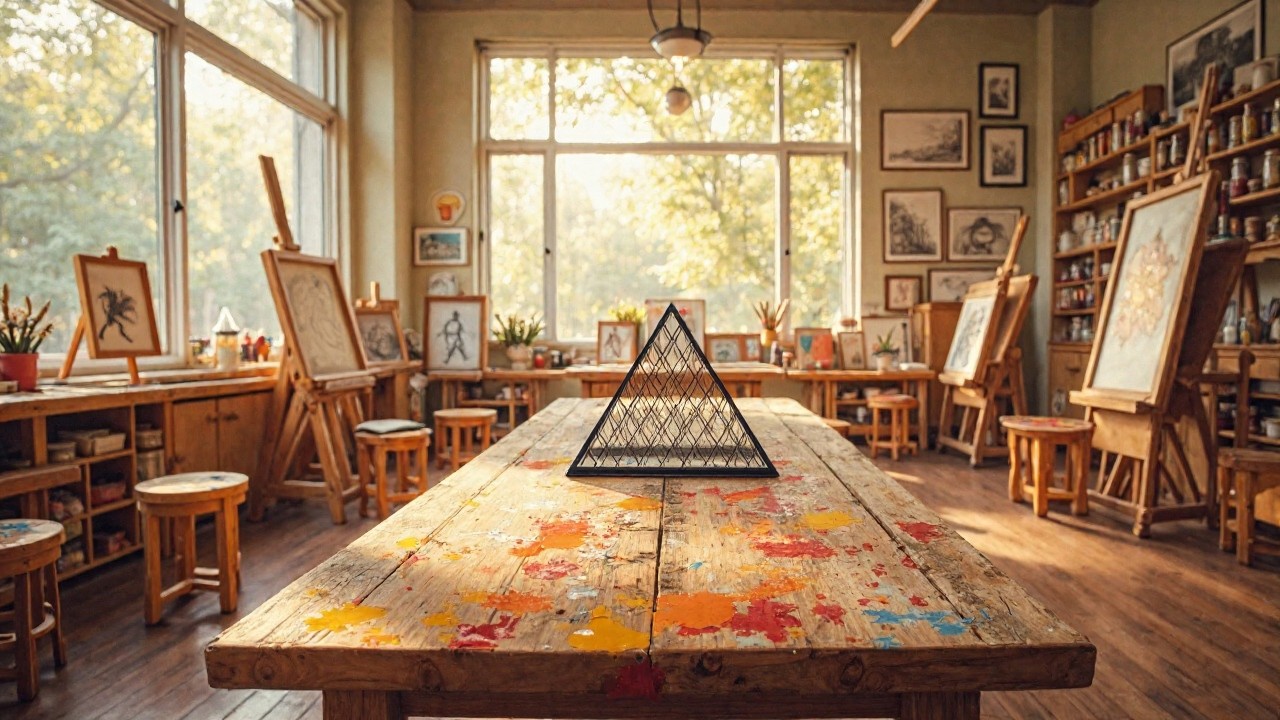}
    \end{subfigure}\hspace{2pt}%
    \begin{subfigure}[]{0.237\linewidth}\centering
        \includegraphics[width=\linewidth]{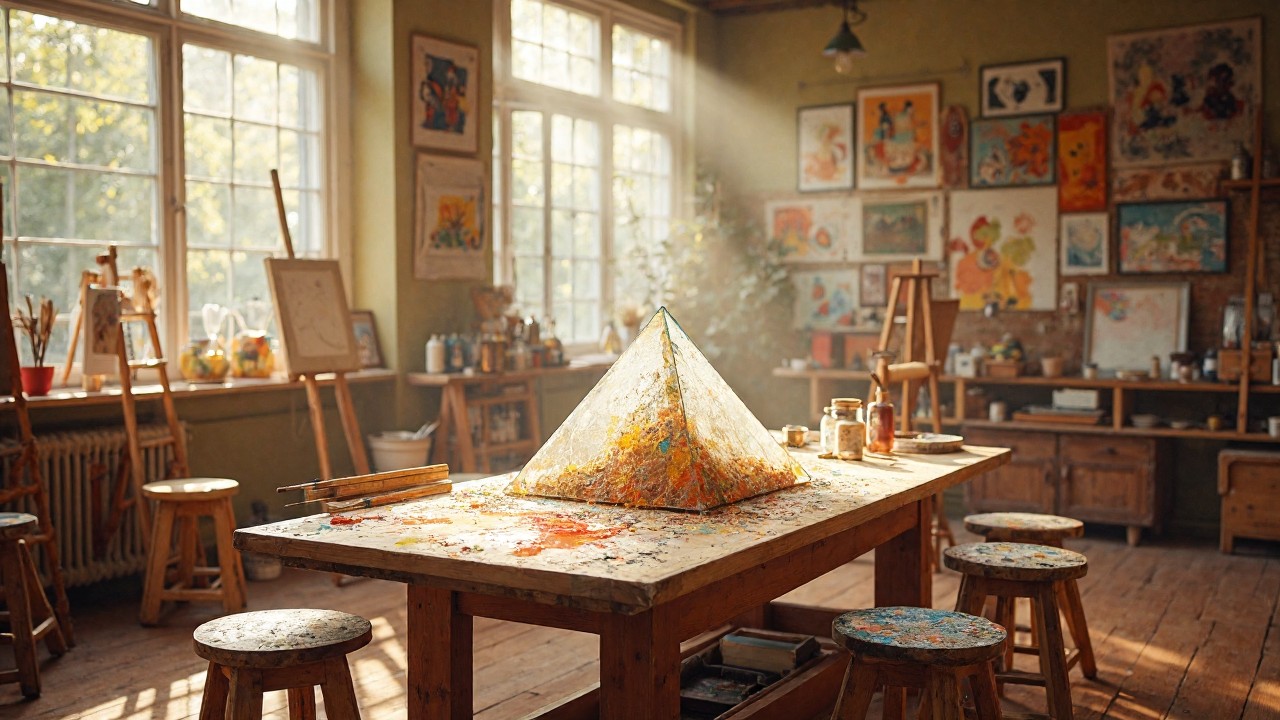}
    \end{subfigure}\\[2pt]

    \begin{minipage}[c]{0.025\linewidth}\centering
        \rotatebox{90}{Pyramid (inset)}
    \end{minipage}%
    \begin{subfigure}[]{0.237\linewidth}\centering
        \includegraphics[
            width=\linewidth,
            trim=500 200 420 250,
            clip
        ]{figures/suppl/generated_other_shapes/artroom_8830/artroom_8830_maskedgt.jpg}
    \end{subfigure}\hspace{2pt}%
    \begin{subfigure}[]{0.237\linewidth}\centering
        \includegraphics[
            width=\linewidth,
            trim=500 200 420 250,
            clip
        ]{figures/results/pyramid.jpg}
    \end{subfigure}\hspace{2pt}%
    \begin{subfigure}[]{0.237\linewidth}\centering
        \includegraphics[
            width=\linewidth,
            trim=500 200 420 250,
            clip
        ]{figures/suppl/generated_other_shapes/artroom_8830/artroom_8830_flux_fill.jpg}
    \end{subfigure}\hspace{2pt}%
    \begin{subfigure}[]{0.237\linewidth}\centering
        \includegraphics[
            width=\linewidth,
            trim=480 170 440 280,
            clip
        ]{figures/suppl/generated_other_shapes/artroom_8830/artroom_8830_flux.jpg}
    \end{subfigure}
    \vspace{-6pt}
    \caption{Additional qualitative results on complex transparent objects (continued).
    This second part supplements \cref{fig:more_objects1} and is separated only for page layout.
    The results further demonstrate that our method maintains physically plausible refraction across different shapes, while existing FLUX variants fail to handle challenging geometries.}
    \label{fig:more_objects2}
\end{figure*}
\begin{figure*}[!t]
    \centering

    \setlength{\tabcolsep}{2pt}     %
    \renewcommand{\arraystretch}{1} %

    \begin{tabular}{c c c c}
        Water (1.333) & Plastic (1.45) & Glass (1.5) & Diamond (2.418) \\[2pt]

        \includegraphics[width=0.235\linewidth]{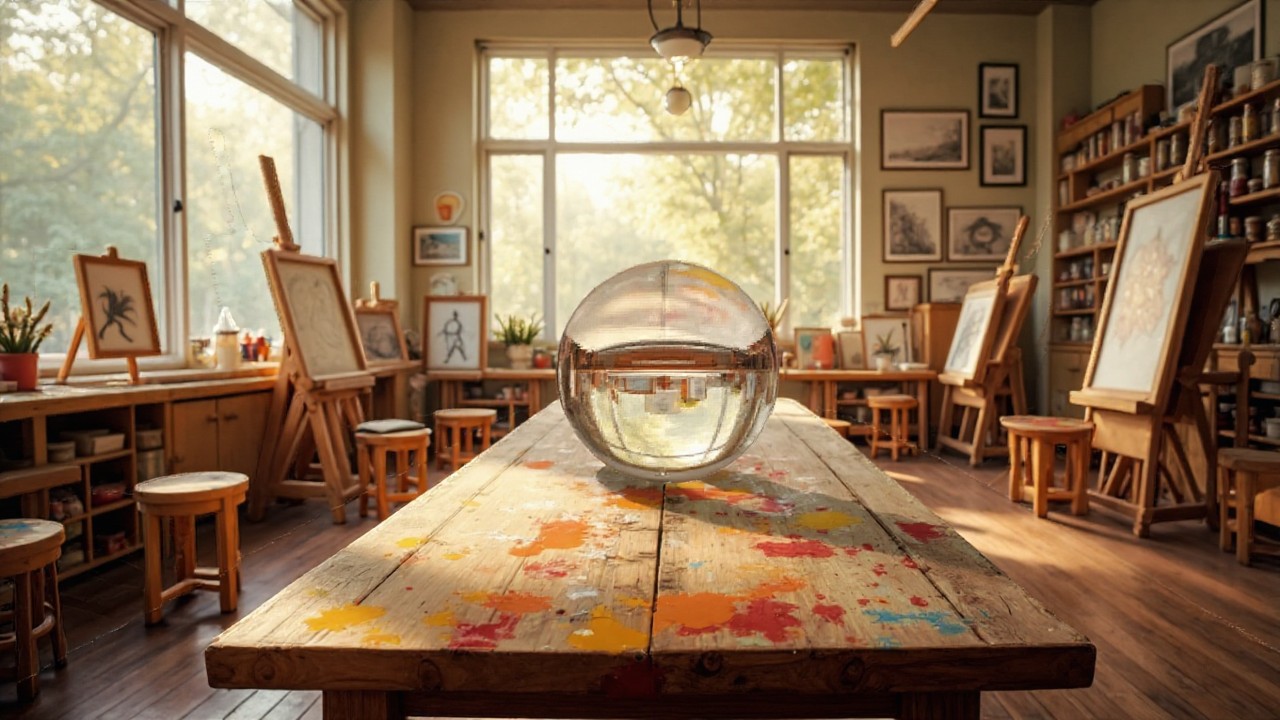} &
        \includegraphics[width=0.235\linewidth]{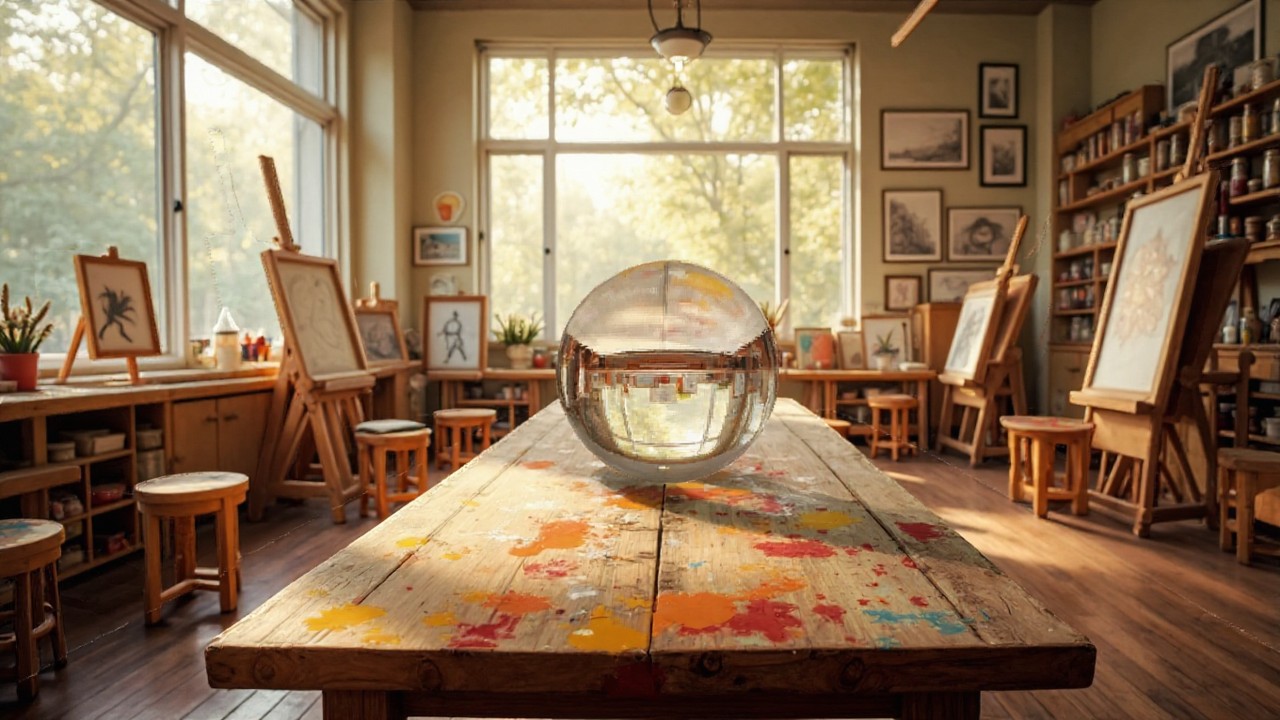} &
        \includegraphics[width=0.235\linewidth]{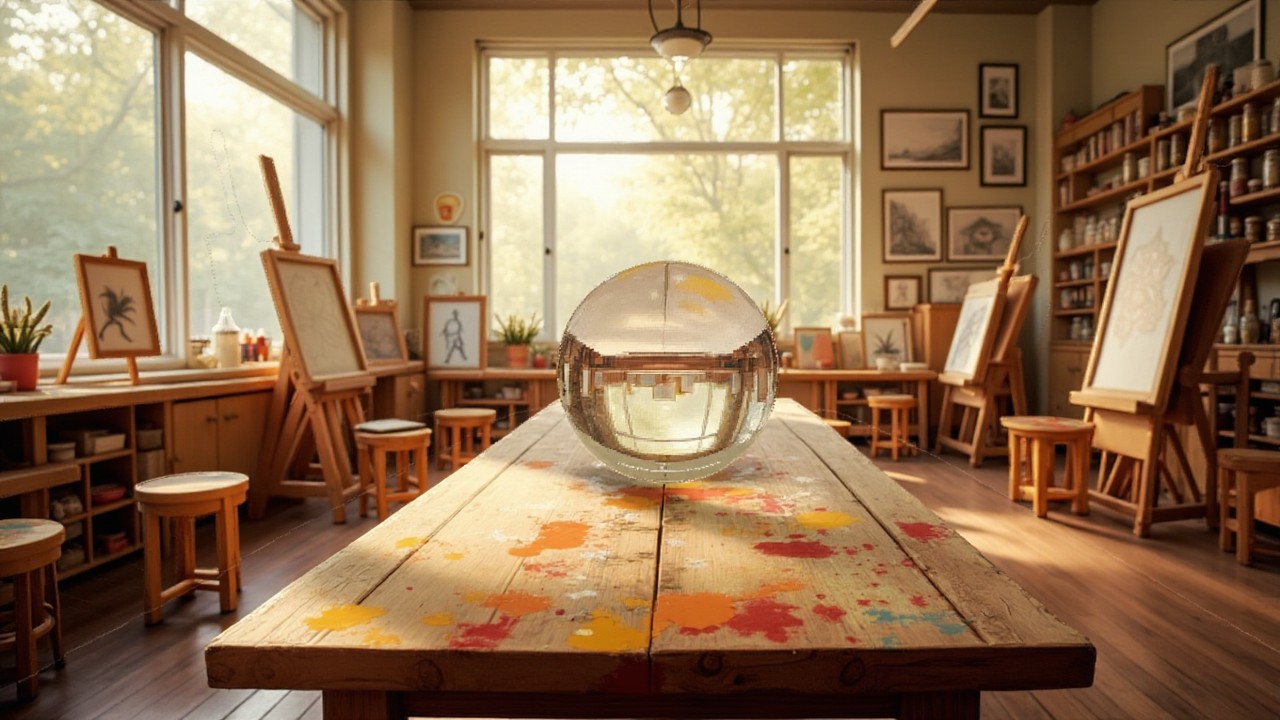} &
        \includegraphics[width=0.235\linewidth]{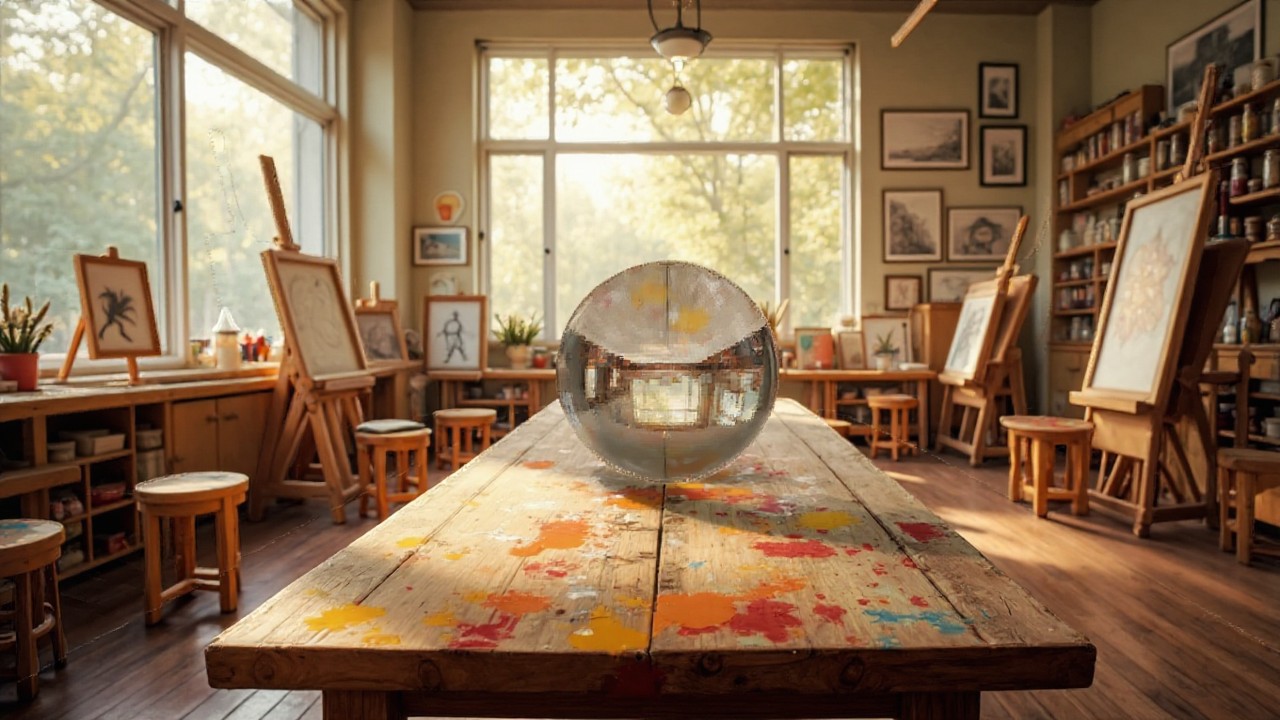} \\[1.5pt]

        \includegraphics[width=0.235\linewidth,
            trim={450pt 200pt 400pt 200pt}, clip]{figures/suppl/other_materials/water.jpg} &
        \includegraphics[width=0.235\linewidth,
            trim={450pt 200pt 400pt 200pt}, clip]{figures/suppl/other_materials/plastic.jpg} &
        \includegraphics[width=0.235\linewidth,
            trim={450pt 200pt 400pt 200pt}, clip]{figures/suppl/other_materials/glass.jpg} &
        \includegraphics[width=0.235\linewidth,
            trim={450pt 200pt 400pt 200pt}, clip]{figures/suppl/other_materials/diamond.jpg}
    \end{tabular}
    \vspace{-6pt}
    \caption{Appearance variation under different refractive indices for a glass sphere.
    We render the same scene while sweeping the refractive index and show the corresponding outputs from Snellcaster (ours). 
    The results demonstrate the strong sensitivity of transparent object appearance to the refractive index and highlight the importance of accurate material estimation for physically consistent generation.}
    \label{fig:ior_sweep}
\end{figure*}
\begin{figure*}[!t]
    \centering

    \begin{minipage}{0.244\linewidth}\centering
        $R_{\text{main}} = 1, R_{\text{pano}} = 1$
    \end{minipage}
    \hfill
    \begin{minipage}{0.244\linewidth}\centering
        $R_{\text{main}} = 3, R_{\text{pano}} = 3$
    \end{minipage}
    \hfill
    \begin{minipage}{0.244\linewidth}\centering
        $R_{\text{main}} = 5, R_{\text{pano}} = 5$
    \end{minipage}
    \hfill
    \begin{minipage}{0.244\linewidth}\centering
        $R_{\text{main}} = 8, R_{\text{pano}} = 8$
    \end{minipage}
    
    \begin{minipage}{0.244\linewidth}\centering
        \includegraphics[width=\linewidth,trim=500 260 460 270,clip]{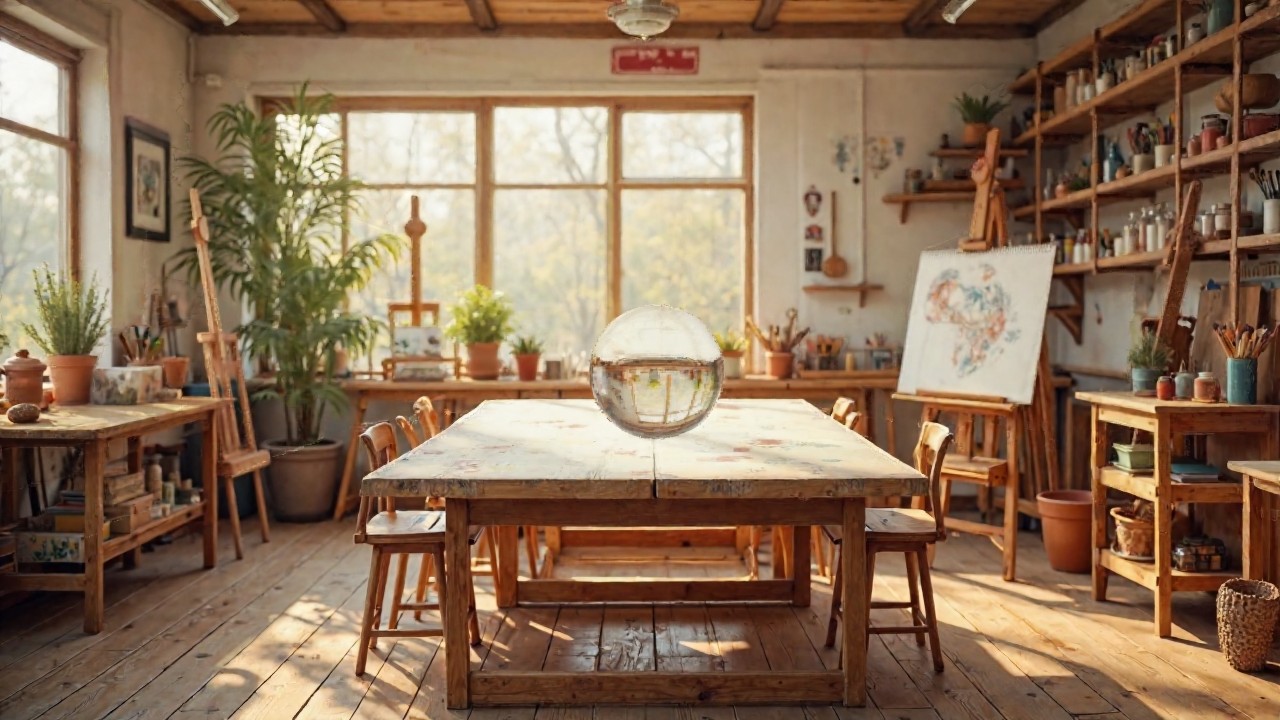}
    \end{minipage}
    \hfill
    \begin{minipage}{0.244\linewidth}\centering
        \includegraphics[width=\linewidth,trim=500 260 460 270,clip]{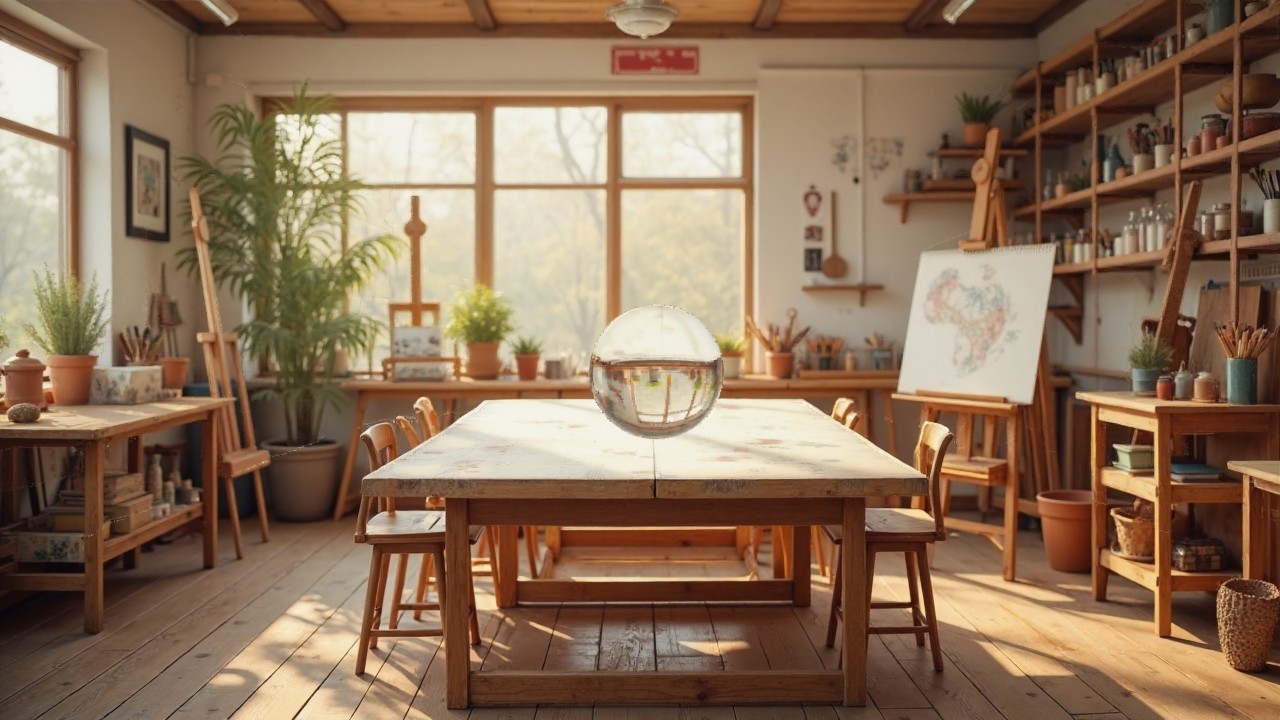}
    \end{minipage}
    \hfill
    \begin{minipage}{0.244\linewidth}\centering
        \includegraphics[width=\linewidth,trim=500 260 460 270,clip]{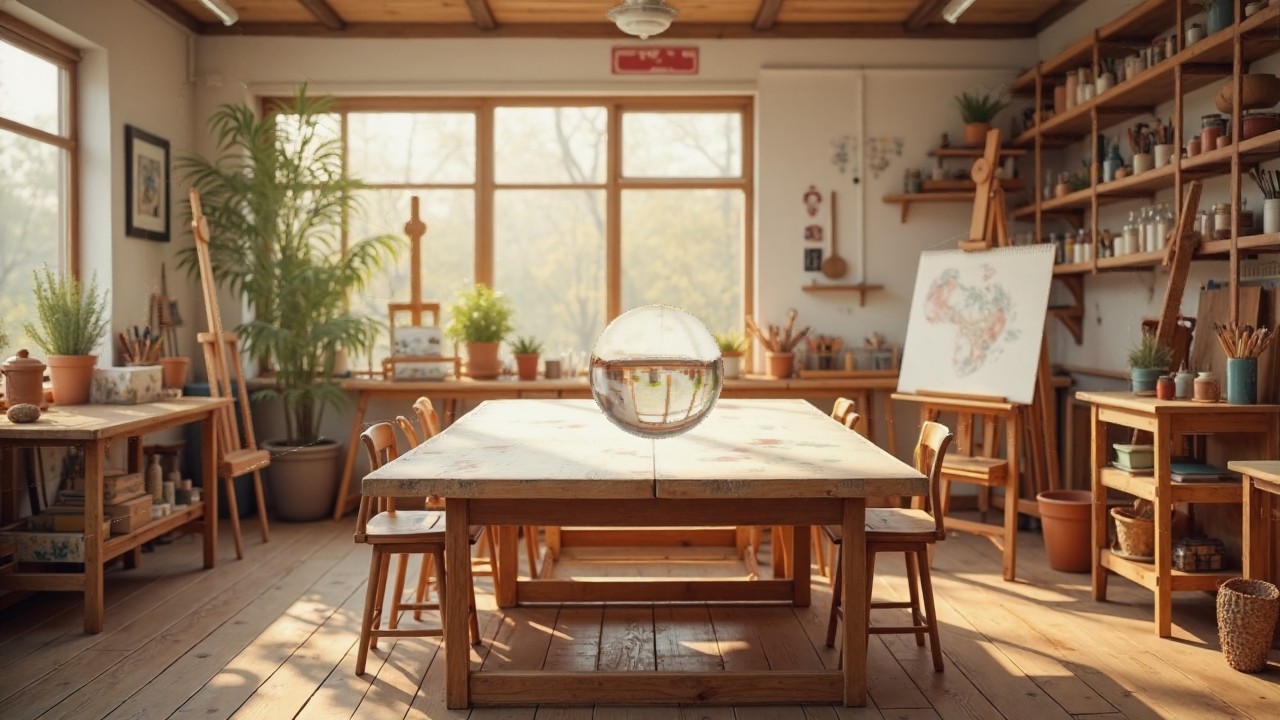}
    \end{minipage}
    \hfill
    \begin{minipage}{0.244\linewidth}\centering
        \includegraphics[width=\linewidth,trim=500 260 460 270,clip]{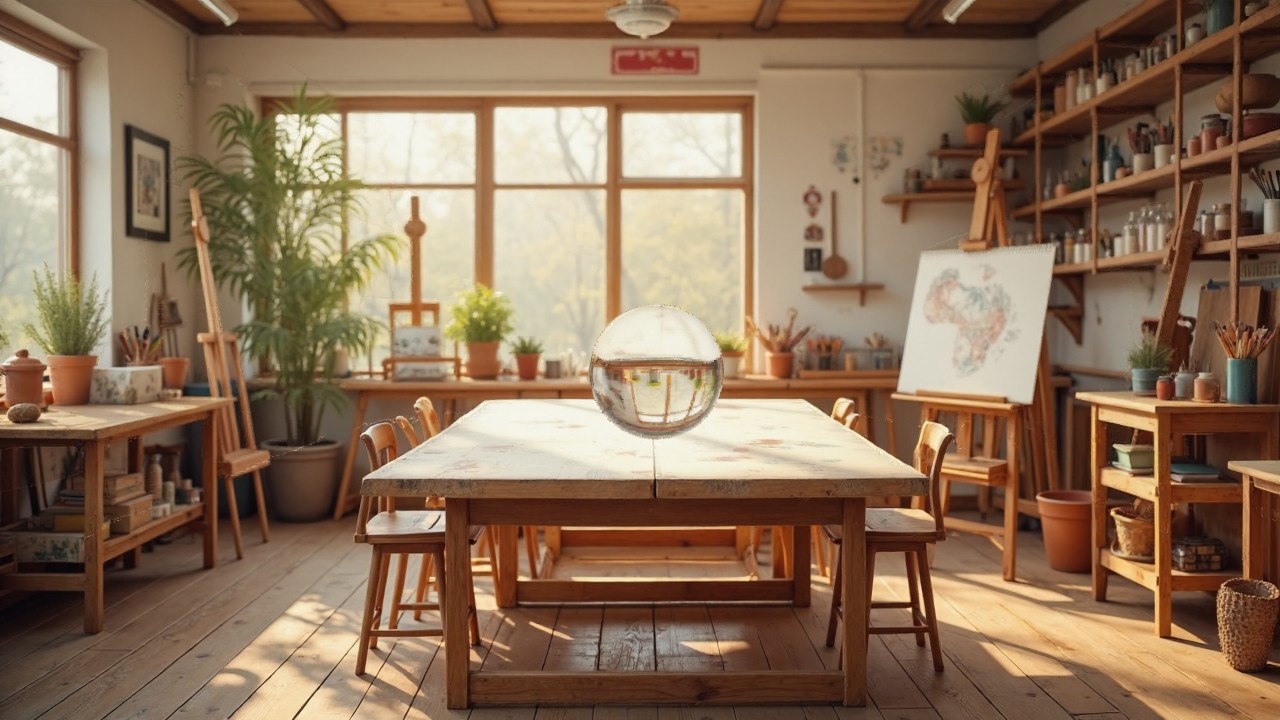}
    \end{minipage}
    \vspace{2pt}
    \begin{minipage}{0.244\linewidth}\centering
        \includegraphics[width=\linewidth]{figures/suppl/tt_analysis/artroom_main1.jpg}
    \end{minipage}
    \hfill
    \begin{minipage}{0.244\linewidth}\centering
        \includegraphics[width=\linewidth]{figures/suppl/tt_analysis/artroom_main33.jpg}
    \end{minipage}
    \hfill
    \begin{minipage}{0.244\linewidth}\centering
        \includegraphics[width=\linewidth]{figures/suppl/tt_analysis/artroom_main55.jpg}
    \end{minipage}
    \hfill
    \begin{minipage}{0.244\linewidth}\centering
        \includegraphics[width=\linewidth]{figures/suppl/tt_analysis/artroom_main88.jpg}
    \end{minipage}
    \vspace{2pt}
    \begin{minipage}{0.244\linewidth}\centering
        \includegraphics[width=\linewidth]{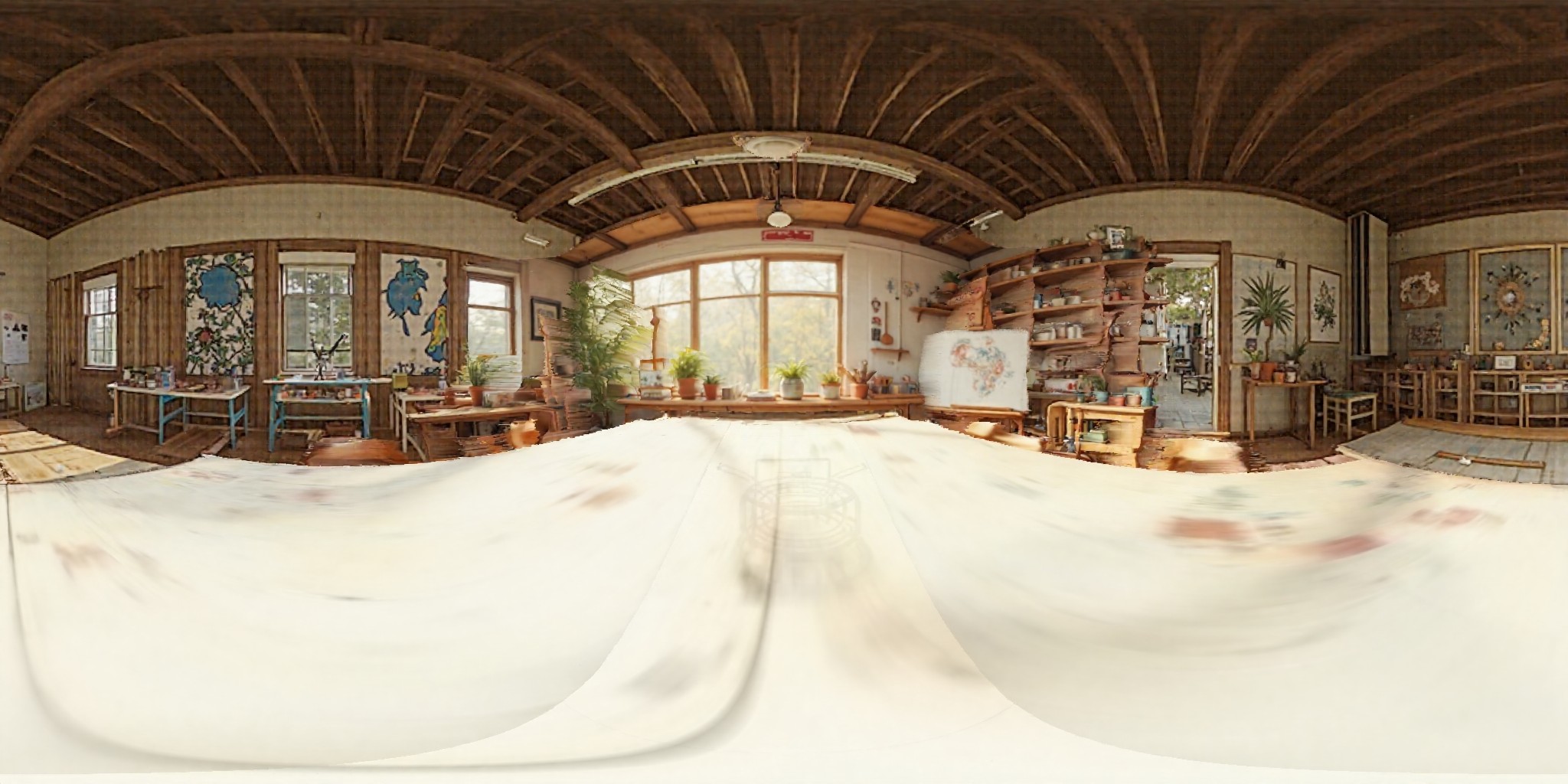}
    \end{minipage}
    \hfill
    \begin{minipage}{0.244\linewidth}\centering
        \includegraphics[width=\linewidth]{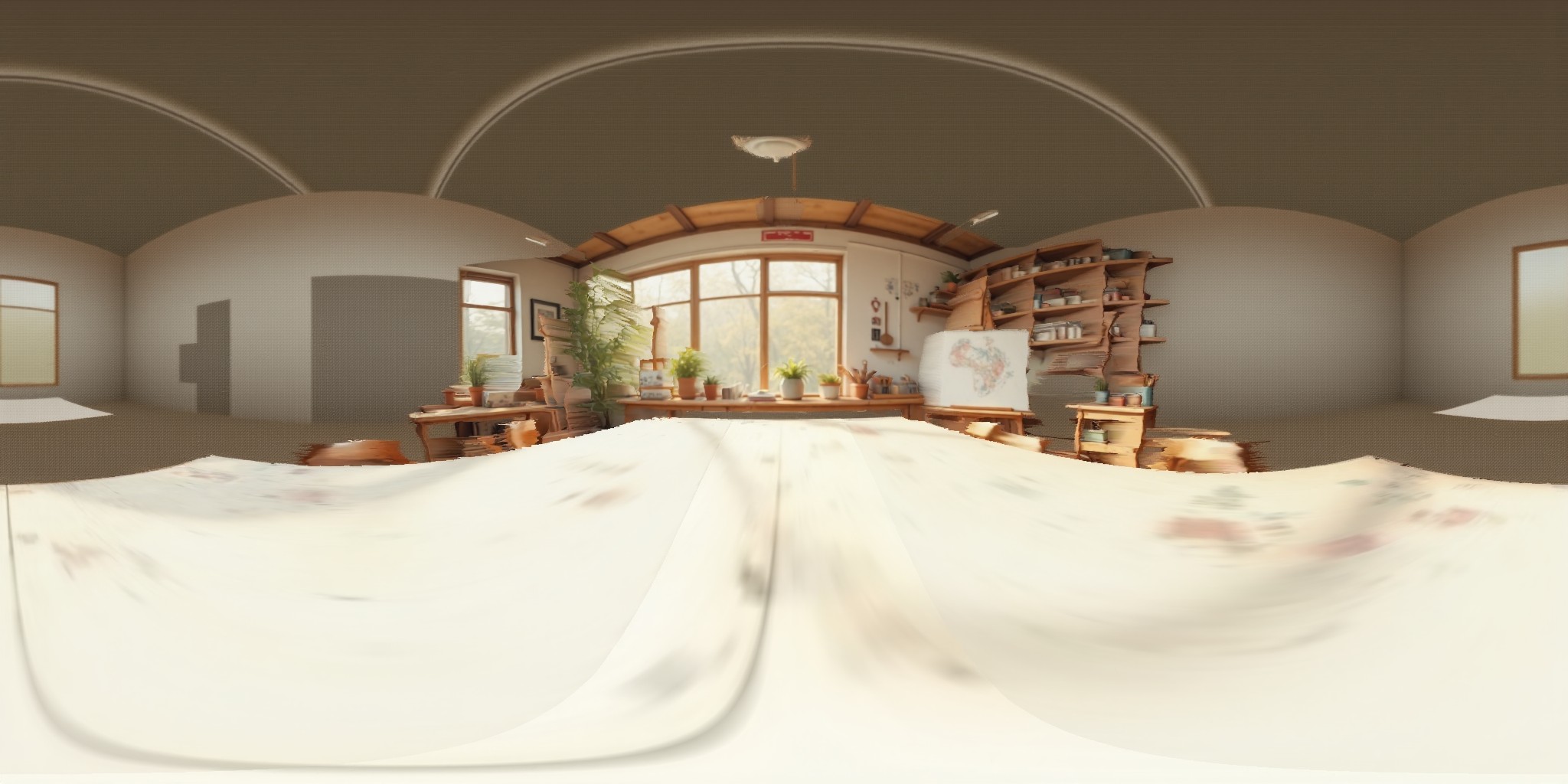}
    \end{minipage}
    \hfill
    \begin{minipage}{0.244\linewidth}\centering
        \includegraphics[width=\linewidth]{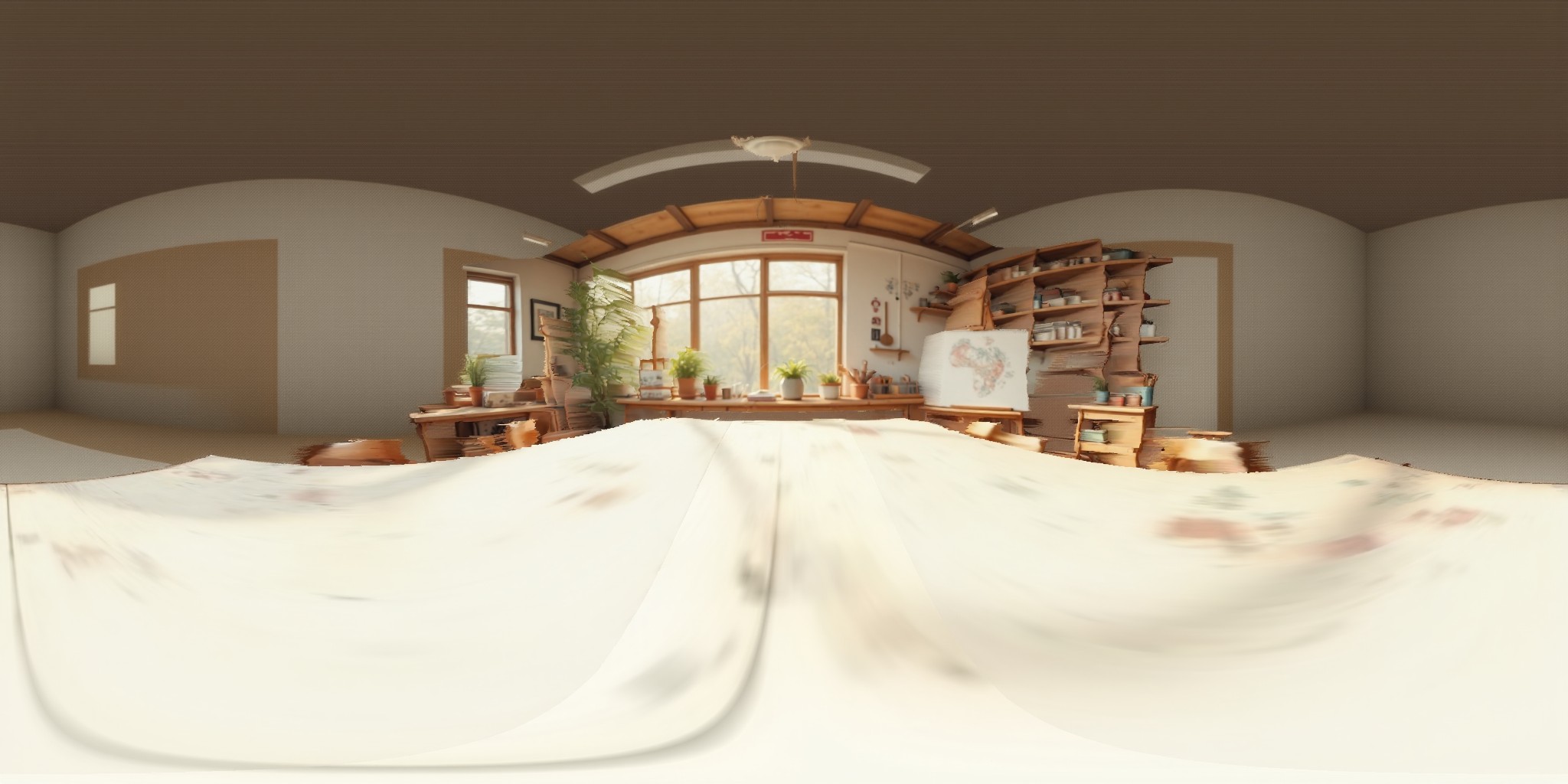}
    \end{minipage}
    \hfill
    \begin{minipage}{0.244\linewidth}\centering
        \includegraphics[width=\linewidth]{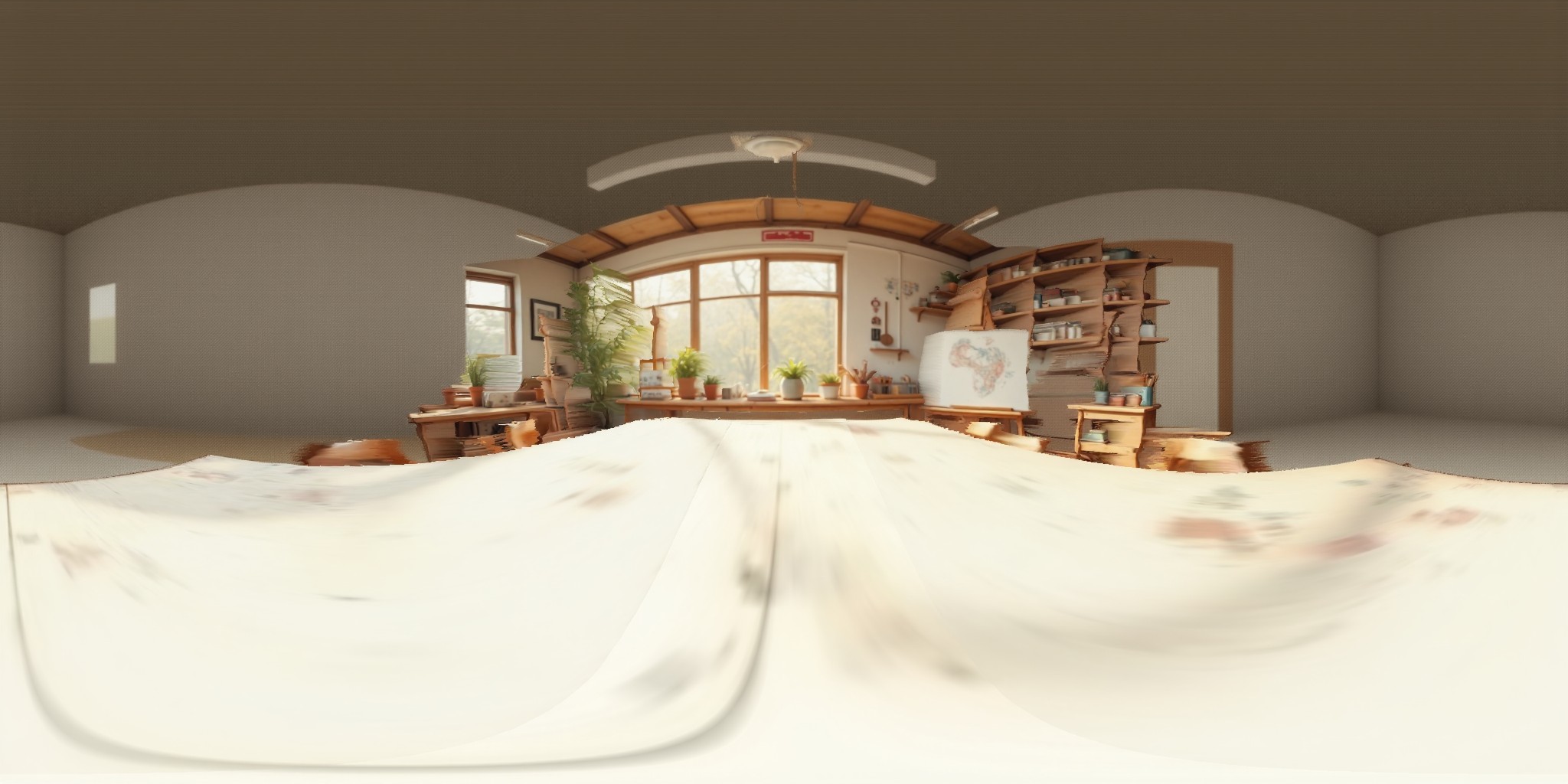}
    \end{minipage}
    \vspace{2pt}
    \begin{minipage}{0.244\linewidth}\centering
        $R_{\text{main}} = 1, R_{\text{pano}} = 1$
    \end{minipage}
    \hfill
    \begin{minipage}{0.244\linewidth}\centering
        $R_{\text{main}} = 3, R_{\text{pano}} = 1$
    \end{minipage}
    \hfill
    \begin{minipage}{0.244\linewidth}\centering
        $R_{\text{main}} = 5, R_{\text{pano}} = 1$
    \end{minipage}
    \hfill
    \begin{minipage}{0.244\linewidth}\centering
        $R_{\text{main}} = 8, R_{\text{pano}} = 1$
    \end{minipage}
    
    \begin{minipage}{0.244\linewidth}\centering
        \includegraphics[width=\linewidth,trim=500 260 460 270,clip]{figures/suppl/tt_analysis/artroom_main1.jpg}
    \end{minipage}
    \hfill
    \begin{minipage}{0.244\linewidth}\centering
        \includegraphics[width=\linewidth,trim=500 260 460 270,clip]{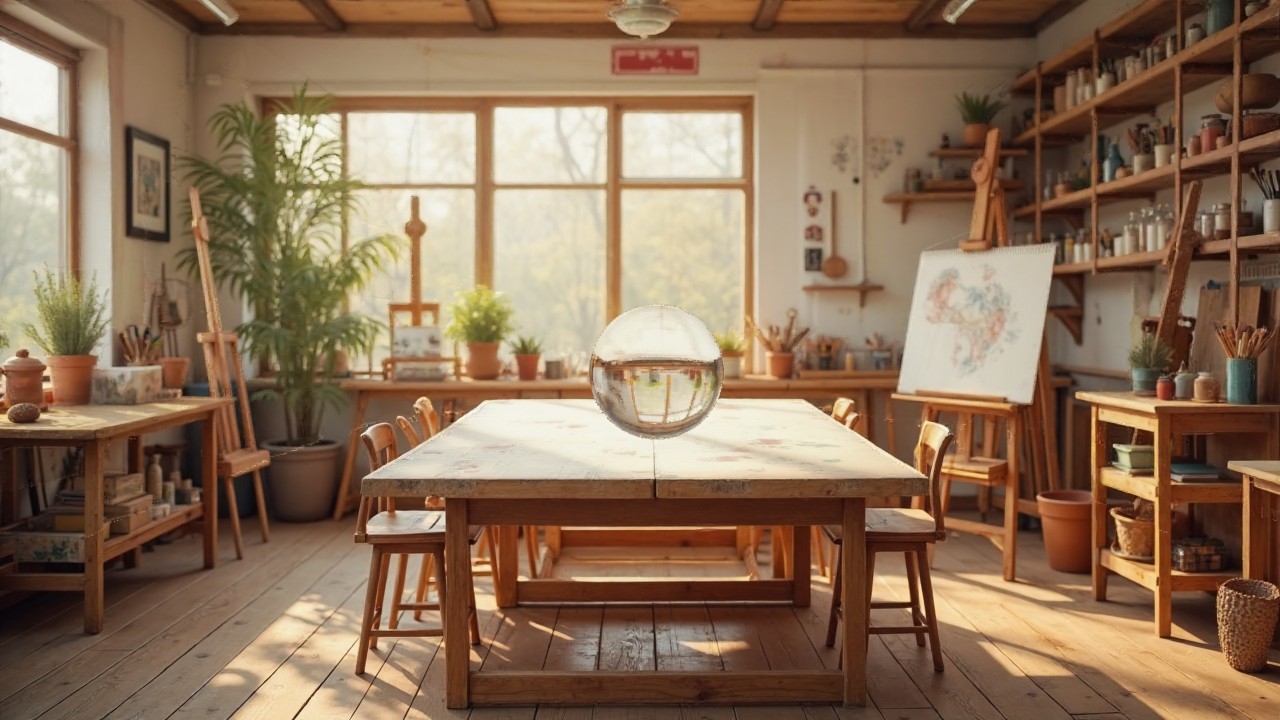}
    \end{minipage}
    \hfill
    \begin{minipage}{0.244\linewidth}\centering
        \includegraphics[width=\linewidth,trim=500 260 460 270,clip]{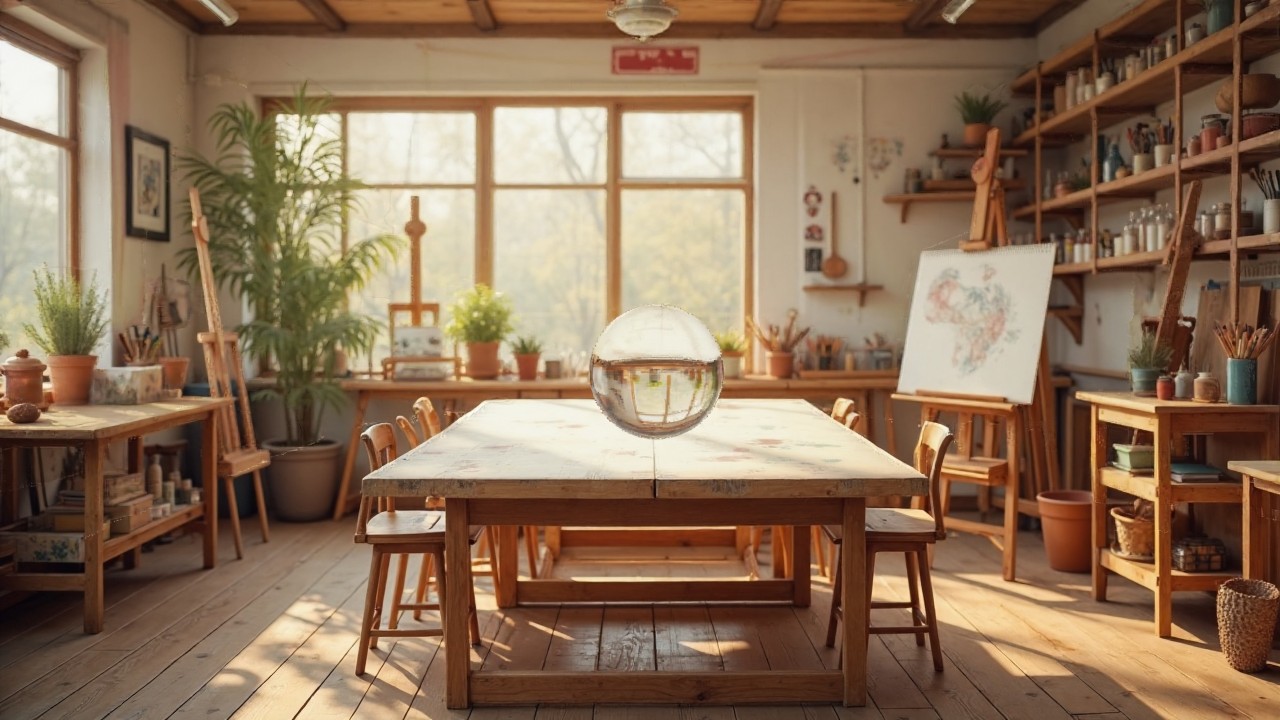}
    \end{minipage}
    \hfill
    \begin{minipage}{0.244\linewidth}\centering
        \includegraphics[width=\linewidth,trim=500 260 460 270,clip]{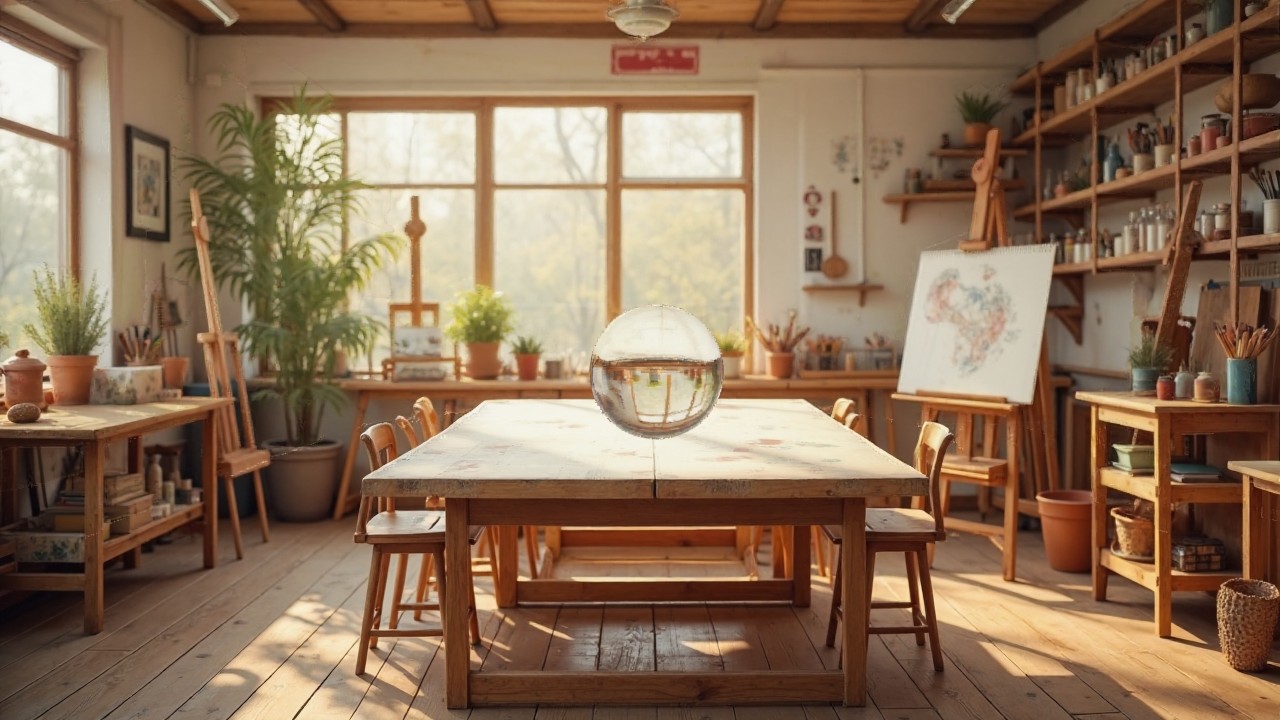}
    \end{minipage}
    \vspace{2pt}
    \begin{minipage}{0.244\linewidth}\centering
        \includegraphics[width=\linewidth]{figures/suppl/tt_analysis/artroom_main1.jpg}
    \end{minipage}
    \hfill
    \begin{minipage}{0.244\linewidth}\centering
        \includegraphics[width=\linewidth]{figures/suppl/tt_analysis/artroom_main3.jpg}
    \end{minipage}
    \hfill
    \begin{minipage}{0.244\linewidth}\centering
        \includegraphics[width=\linewidth]{figures/suppl/tt_analysis/artroom_main5.jpg}
    \end{minipage}
    \hfill
    \begin{minipage}{0.244\linewidth}\centering
        \includegraphics[width=\linewidth]{figures/suppl/tt_analysis/artroom_main8.jpg}
    \end{minipage}
    \vspace{2pt}
    \begin{minipage}{0.244\linewidth}\centering
        \includegraphics[width=\linewidth]{figures/suppl/tt_analysis/artroom_pano1.jpg}
    \end{minipage}
    \hfill
    \begin{minipage}{0.244\linewidth}\centering
        \includegraphics[width=\linewidth]{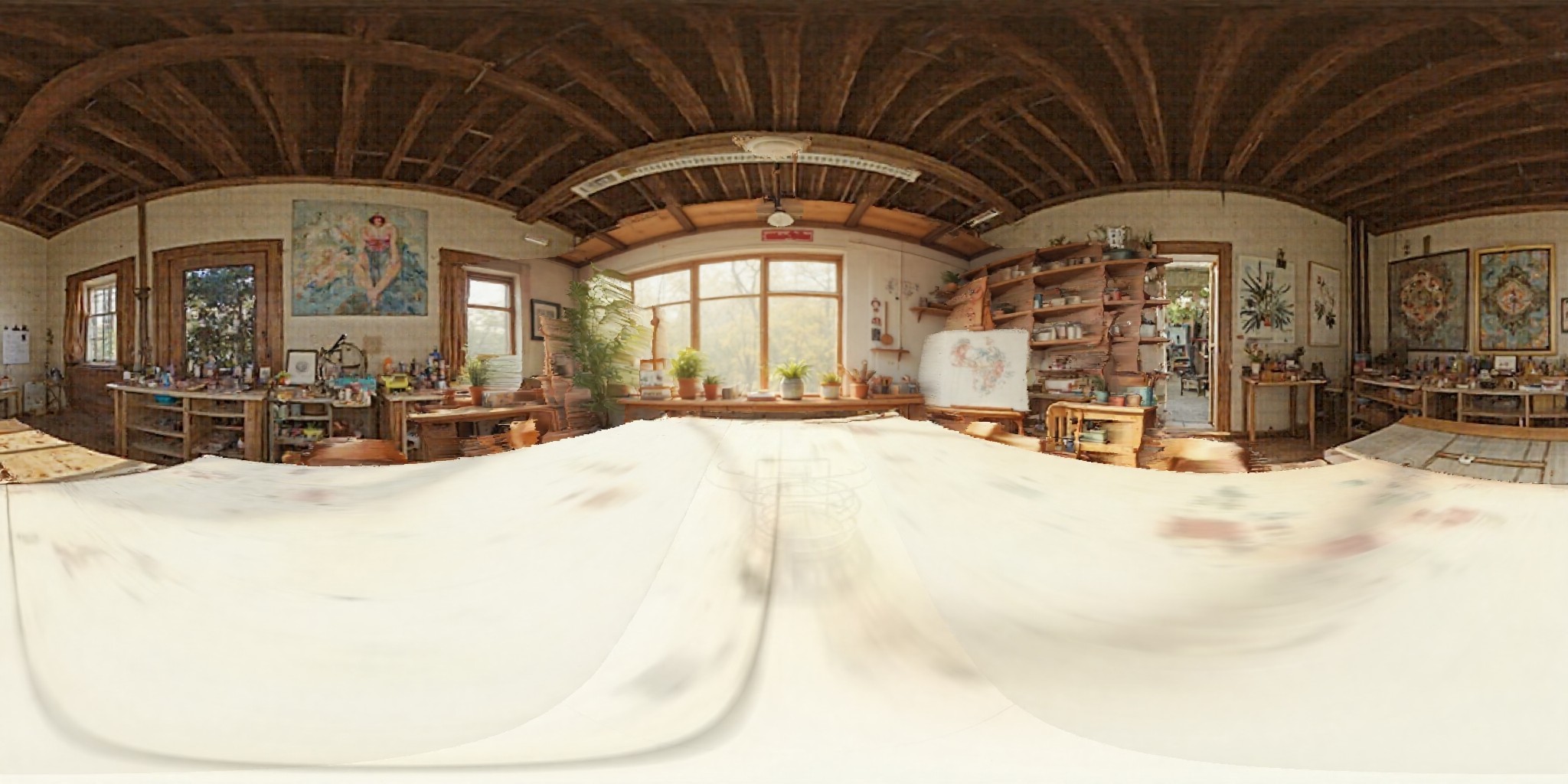}
    \end{minipage}
    \hfill
    \begin{minipage}{0.244\linewidth}\centering
        \includegraphics[width=\linewidth]{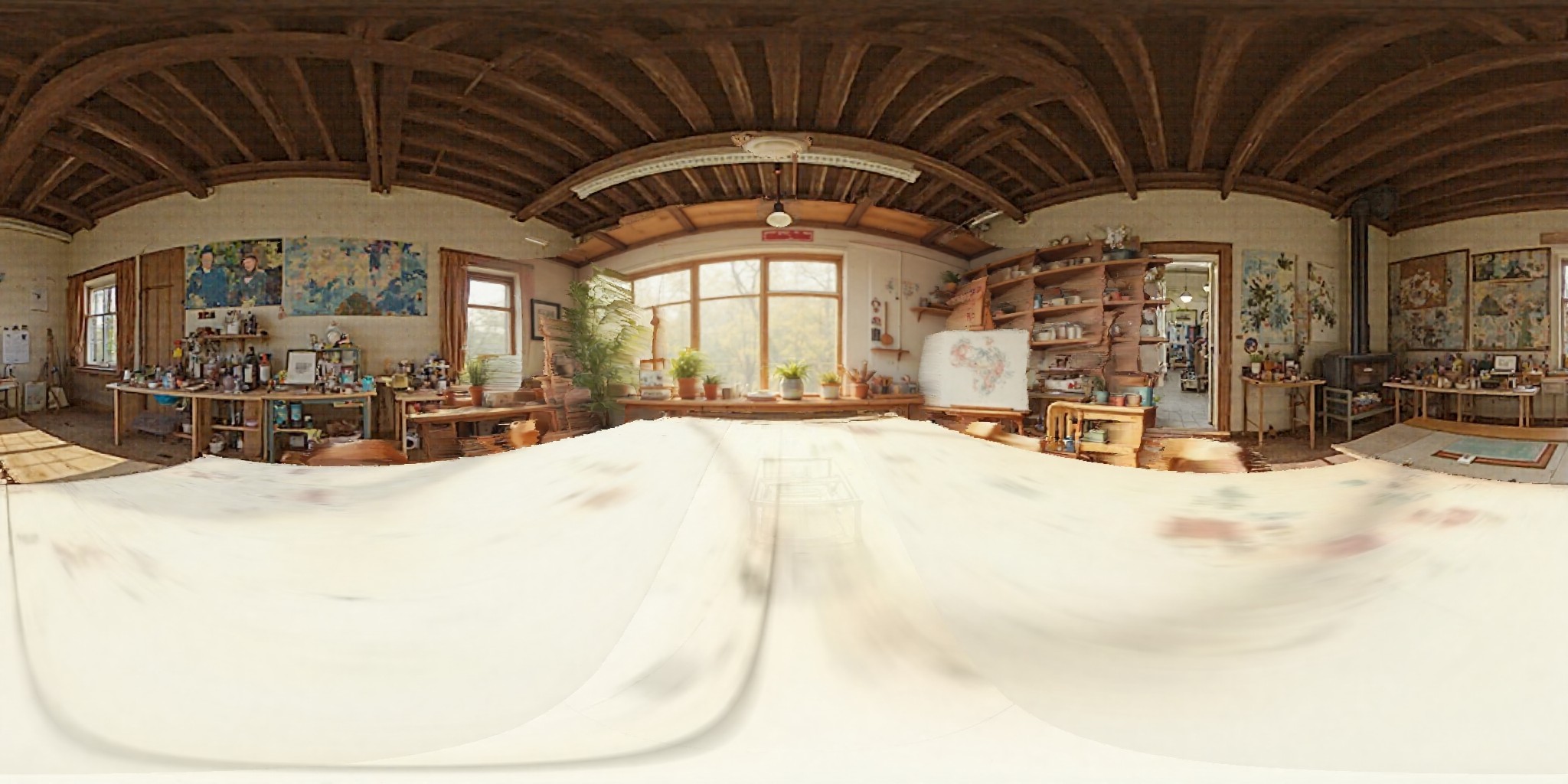}
    \end{minipage}
    \hfill
    \begin{minipage}{0.244\linewidth}\centering
        \includegraphics[width=\linewidth]{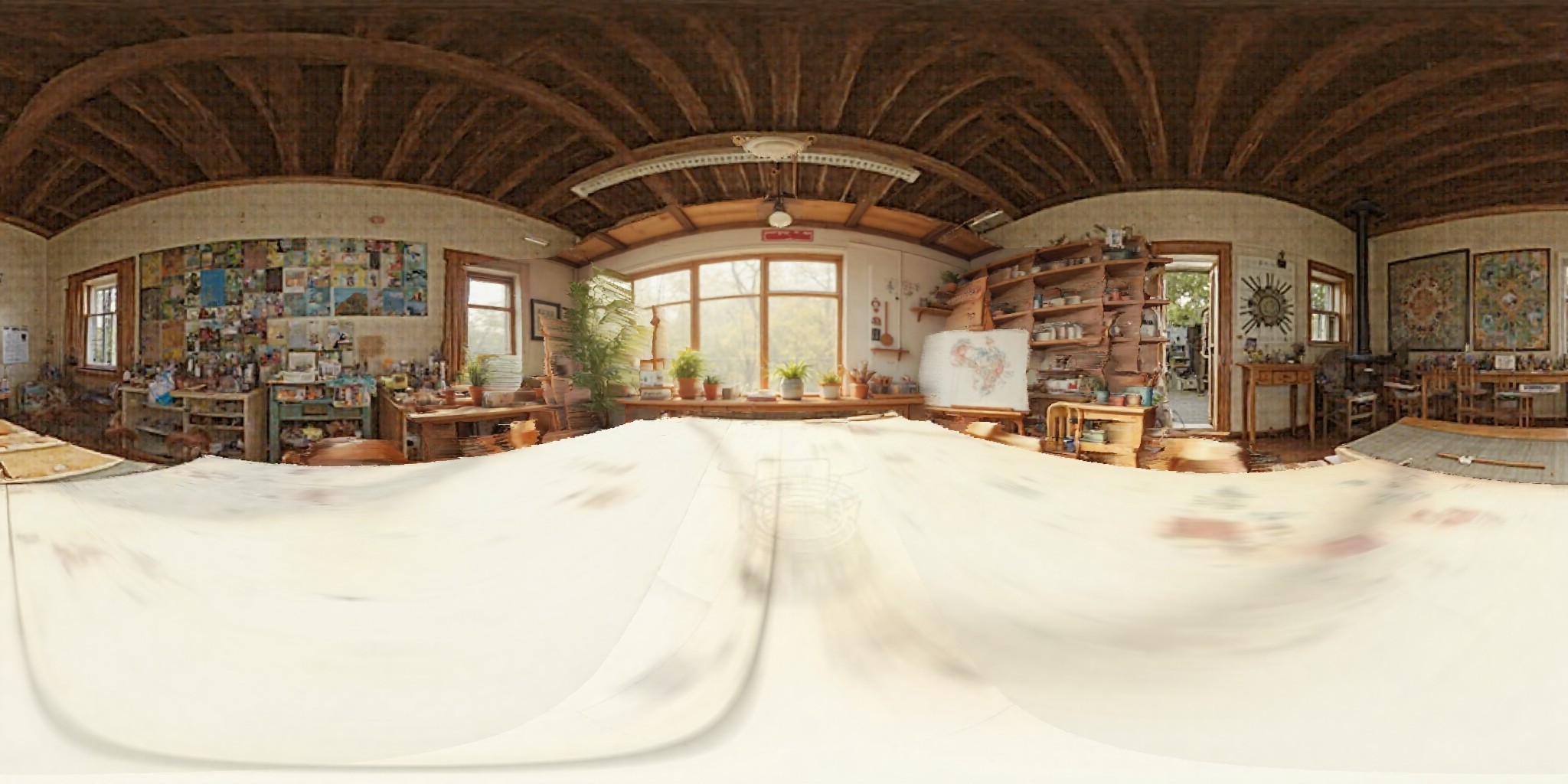}
    \end{minipage}
    \vspace{-6pt}
    \caption{
    Qualitative comparison of time travel repeat counts.
    The top half of the figure shows configurations where the same repeat count is applied to both the perspective and panorama views ($R_{\text{main}} = R_{\text{pano}}$).
    The bottom half shows a setting where time travel is applied only to the perspective view while the panorama is left unmodified ($R_{\text{main}} \in \{1,3,5,8\}, R_{\text{pano}} = 1$).
    }

    \label{fig:tt_analysis}

\end{figure*}

\begin{table*}[!t]
\centering
\small
\caption{Full per-scene quantitative comparison of our method against FLUX-based inpainting model~\cite{labs2025flux1kontextflowmatching}, FLUX-dev~\cite{labs2025flux1kontextflowmatching}, FLUX.2-dev~\cite{flux-2-2025}, Qwen-Image~\cite{wu2025qwen}, and Stable Diffusion 3.5 (Large)~\cite{sd3}. 
The table is split into two sections to maintain readability in portrait orientation. 
Higher CLIP, ImageReward, and PSNR are better, while lower LPIPS is better.
Our method consistently outperforms the baselines in terms of refraction fidelity (measured by PSNR and LPIPS), without sacrificing text alignment (CLIP).
}
\label{tab:full_results}
\vspace{-8pt}
\begin{tabularx}{\linewidth}{@{}l *{6}{C} *{6}{C}@{}}
\toprule
& \multicolumn{6}{c}{CLIP $\uparrow$} & \multicolumn{6}{c}{ImageReward $\uparrow$} \\
\cmidrule(lr){2-7} \cmidrule(lr){8-13}
Scene & Inpaint & FLUX & FLUX2 & Qwen & SD3.5 & Ours & Inpaint & FLUX & FLUX2 & Qwen & SD3.5 & Ours \\
\midrule
Artroom      & \textbf{35.06} & 31.81 & 33.44 & 34.46 & 34.37 & 34.12 & 0.14 & 0.09 & \textbf{0.80} & 0.10 & 0.52 & 0.30\\
Cafe         & 32.55 & 30.95 & 31.11 & 31.36 & \textbf{33.72} & 30.55 & -1.42 & -1.25 & -0.85 & \textbf{-0.78} & -1.00 & -1.26 \\
Cave         & 36.85 & 36.87 & 36.89 & 36.14 & \textbf{37.77} & 36.83 & -0.28 & -0.14 & \textbf{0.24} & -0.07 & -0.03 & -0.24\\
Desert       & 34.93 & 35.26 & 35.02 & 35.22 & 35.44 & \textbf{35.73} & 0.67 & 1.09 & \textbf{1.31} & 1.01 & 0.97 & 0.79\\
Dining Room  & \textbf{32.53} & 28.15 & 30.76 & 30.03 & 32.19 & 32.04 & -0.87 & -0.40 & \textbf{0.43} & -0.25 & -0.04 & -0.34 \\
Karaoke      & 36.22 & 34.01 & \textbf{37.49} & 34.64 & 37.18 & 35.87 & 0.21 & 0.36 & \textbf{1.18} & 0.75 & 0.98 & 0.43 \\
Kitchen      & 30.03 & 32.96 & \textbf{33.09} & 32.42 & 32.70 & 30.29 & -0.77 & 0.30 & \textbf{0.51} & 0.39 & -0.23 & -0.71 \\
Landscape    & 34.02 & 31.66 & 33.69 & 30.63 & \textbf{35.69} & 35.04  & 0.02 & -0.99 & -0.88 & 0.05 & \textbf{0.52} & -0.18 \\
Living Room  & 29.74 & 29.77 & 29.01 & 29.03 & \textbf{32.74} & 28.44 & -1.18 & -0.15 & -0.07 & -0.66 & \textbf{0.10} & -0.73 \\
Office       & 32.51 & 32.15 & 33.16 & 32.78 & \textbf{33.82} & 29.58 & -1.22 & -0.96 & -0.88 & \textbf{-0.47} & -0.68 & -1.29 \\
\midrule
Average      & 33.44 & 32.36 & 33.37 & 32.67 & \textbf{34.56} & 32.85 & -0.47 & -0.20 & \textbf{0.18} & 0.01 & 0.08 & -0.32 \\
\bottomrule
\end{tabularx}

\vspace{0.8em} %

\begin{tabularx}{\linewidth}{@{}l *{6}{C} *{6}{C}@{}}
\toprule
& \multicolumn{6}{c}{PSNR $\uparrow$} & \multicolumn{6}{c}{LPIPS $\downarrow$} \\
\cmidrule(lr){2-7} \cmidrule(lr){8-13}
Scene & Inpaint & FLUX & FLUX2 & Qwen & SD3.5 & Ours & Inpaint & FLUX & FLUX2 & Qwen & SD3.5 & Ours \\
\midrule
Artroom      & 13.50 & 13.43 & 12.89 & 12.93 & 13.25 & \textbf{16.67} & 0.47 & 0.47 & 0.43 & 0.45 & 0.52 & \textbf{0.27} \\
Cafe         & 11.75 & 11.81 & 11.21 & 11.69 & 11.46 & \textbf{15.65} & 0.44 & 0.45 & 0.43 & 0.45 & 0.53 & \textbf{0.23} \\
Cave         & 13.76 & 14.05 & 13.58 & 13.66 & 13.67 & \textbf{18.91} & 0.50 & 0.50 & 0.49 & 0.50 & 0.56 & \textbf{0.25}  \\
Desert       & 15.26 & 16.36 & 14.31 & 15.04 & 14.05 & \textbf{18.71} & 0.35 & 0.35 & 0.41 & 0.38 & 0.43 & \textbf{0.20} \\
Dining Room  & 14.20 & 14.09 & 14.02 & 14.26 & 14.43 & \textbf{15.14} & 0.44 & 0.43 & 0.44 & 0.43 & 0.49 & \textbf{0.21} \\
Karaoke      & 11.01 & 10.93 & 10.21 & 10.79 & 10.29 & \textbf{17.09} & 0.53 & 0.52 & 0.56 & 0.54 & 0.57 & \textbf{0.24} \\
Kitchen      & 11.79 & 11.72 & 11.51 & 11.82 & 11.63 & \textbf{15.65} & 0.51 & 0.55 & 0.55 & 0.53 & 0.58 & \textbf{0.27} \\
Landscape    & 11.28 & 11.12 & 10.70 & 12.09 & 10.85 & \textbf{15.05} & 0.47 & 0.51 & 0.49 & 0.49 & 0.53 & \textbf{0.29} \\
Living Room  & 12.83 & 12.07 & 12.17 & 12.26 & 12.00 & \textbf{16.51} & 0.48 & 0.50 & 0.53 & 0.49 & 0.55 & \textbf{0.23} \\
Office       & 11.26 & 11.24 & 10.86 & 10.98 & 10.88 & \textbf{15.71} & 0.48 & 0.49 & 0.49 & 0.54 & 0.56 & \textbf{0.23} \\
\midrule
Average      & 12.66 & 12.68 & 12.15 & 12.55 & 12.25 & \textbf{16.51} & 0.47 & 0.48 & 0.48 & 0.48 & 0.53 & \textbf{0.24} \\
\bottomrule
\end{tabularx}
\end{table*}
In this section, we provide additional qualitative and quantitative results that complement the main paper.
We present qualitative results for additional transparent objects, analyze the effect of varying refractive indices (e.g., water, plastic), provide the full per-scene quantitative comparison table, a quantitative ablation study table, and study the effect of different time travel repeat counts.
In our implementation, generating a perspective–panorama
pair requires approximately 126\,seconds on an 80\,GB NVIDIA A100 GPU (pre-processing time not included).

\paragraph{Complex transparent objects.}
We provide extended qualitative comparisons across a variety of transparent objects. 
\cref{fig:more_objects1,fig:more_objects2} show generated results for five additional objects.
From these examples, we observe that our method continues to produce refractions that are visually correct and physically plausible, closely matching the Blender reference images.
In contrast, the Flux inpainting model struggles considerably: while the main paper showed that it can sometimes produce a glass sphere, albeit with incorrect physics, it consistently fails once the object shape becomes more complex, even when provided with the exact foreground mask.
The standard Flux model is more capable of generating glass-like objects, yet its outputs often exhibit low transmittance or hollow interiors that do not correspond to the appearance of real transparent materials.
These results collectively highlight the robustness of our method across diverse geometries and the difficulty existing generative models face in handling complex refractive behaviors.

\paragraph{Effect of refractive index.}
We analyze the effect of varying the refractive index for a glass sphere in \cref{fig:ior_sweep}.
This highlights the sensitivity of appearance to the refractive index and illustrates the importance of accurate material estimation when synthesizing transparent objects.

\paragraph{Time travel analysis.}
Finally, we analyze the effect of repeat counts in time travel. 
More repeats generally produce smoother results but suppress high frequency details, making the generated image appear overly uniform and less realistic.
We first evaluate configurations where the repeat count is applied equally to both the perspective and panorama view ($R_{\text{main}} = R_{\text{pano}} \in \{1, 3, 5, 8\}$).
As shown in \cref{fig:tt_analysis}, increasing the repeat count does improve global blending and reduces artifacts like random spots.
However, we observe that it becomes overly aggressive on the panorama branch.
Since the panorama covers a much wider field of view with rich global structure, repeated smoothing removes important details and leads to blurry, unrealistic results.
To address this issue, we apply time travel only to the perspective view while disabling it for the panorama ($R_{\text{main}} \in \{1,3,5,8\}$ while $R_{\text{pano}} = 1$).
This retains the benefits of smoothing and stabilizing the perspective branch while preserving the rich high frequency information present in the panorama.
As illustrated in the second half of Figure~\ref{fig:tt_analysis}, this configuration yields cleaner and more consistent results without sacrificing panoramic detail, offering a better balance between smoothness and realism.
\subsection{Quantitative Evaluation}

\paragraph{Full scene comparison.}
We provide the full per-scene quantitative comparison in \cref{tab:full_results}, for every indoor scene prompt.
Across CLIP, PSNR, and LPIPS, our method achieves consistently stronger results than both baselines, with particularly large gains in PSNR and LPIPS, which measure physical consistency with the Blender reference images.
On these metrics, our approach performs significantly better than the baselines on every single scene.
For ImageReward, the standard Flux model obtains a slightly higher score, although the difference is negligible.
This is expected because ImageReward measures prompt alignment rather than physical plausibility, and our prompts mention many scene elements, so variations outside the object region can influence the score.
\begin{figure}[!t]
    \centering

    \setlength{\tabcolsep}{2pt}     %
    \renewcommand{\arraystretch}{1} %

    \begin{tabular}{c c c c}
        Full Model & w/o DPA & w/o LPW & w/o TT \\[2pt]

        \includegraphics[width=0.235\linewidth]{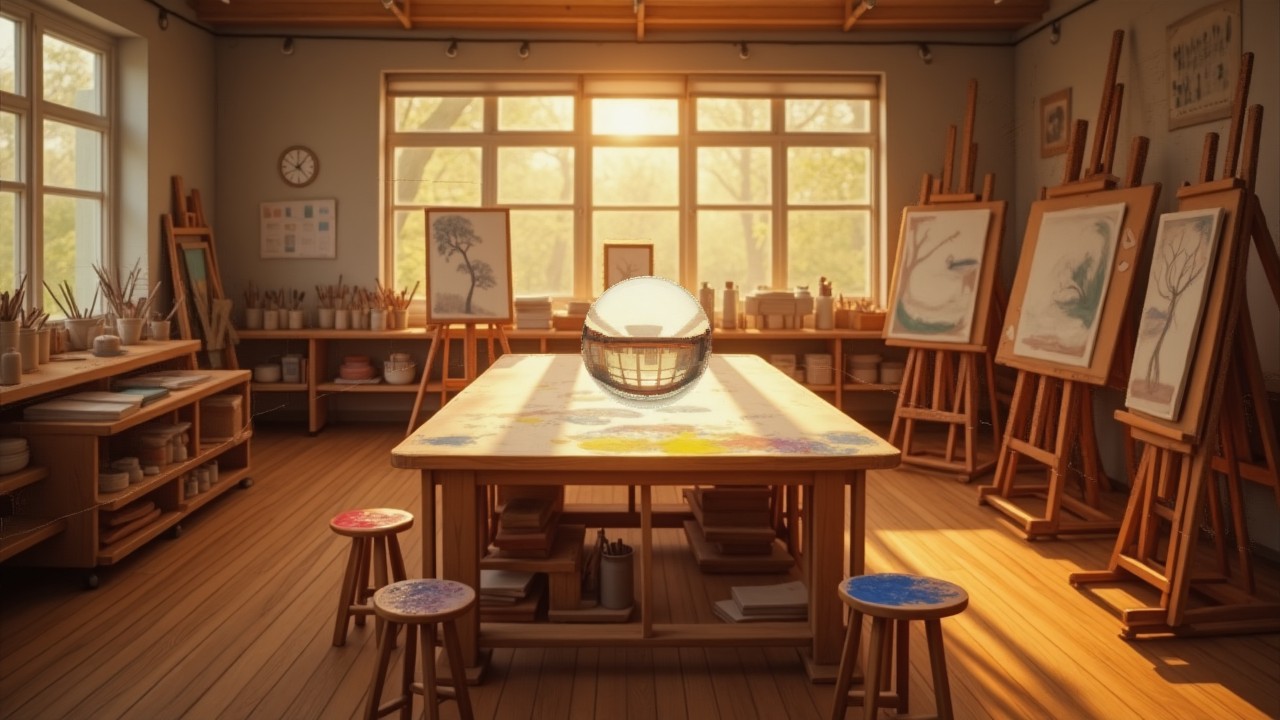} &
        \includegraphics[width=0.235\linewidth]{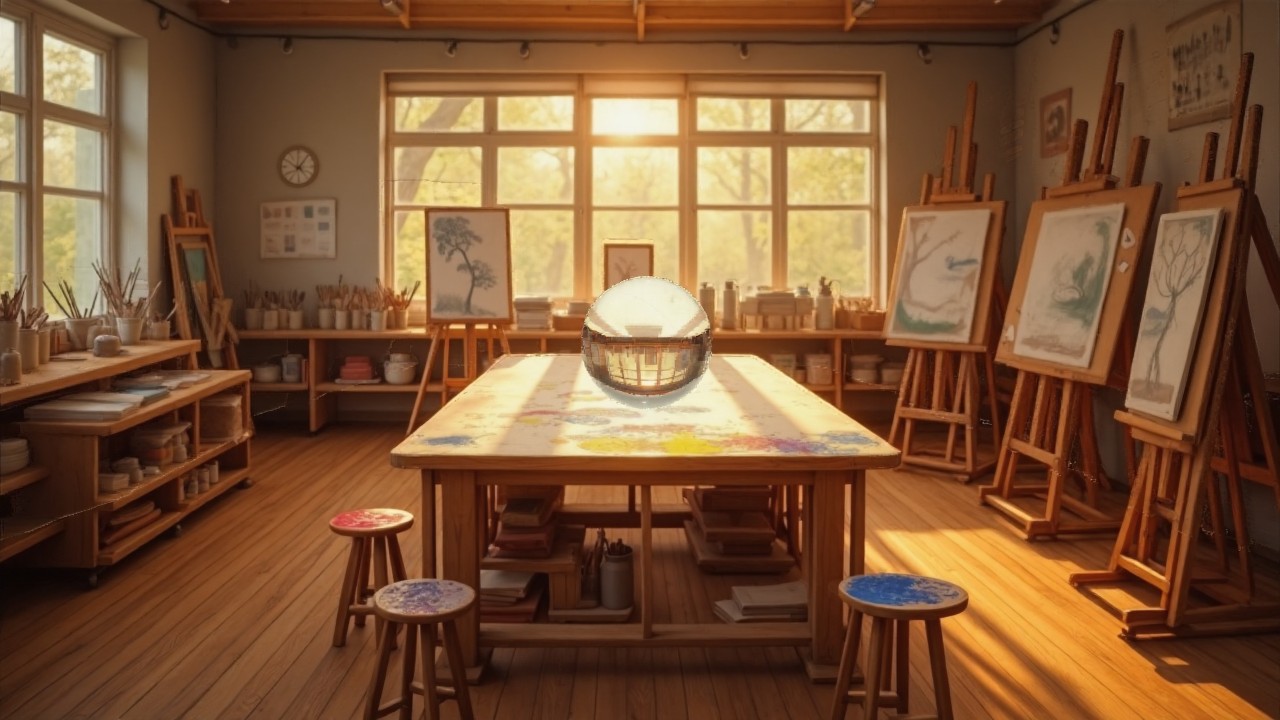} &
        \includegraphics[width=0.235\linewidth]{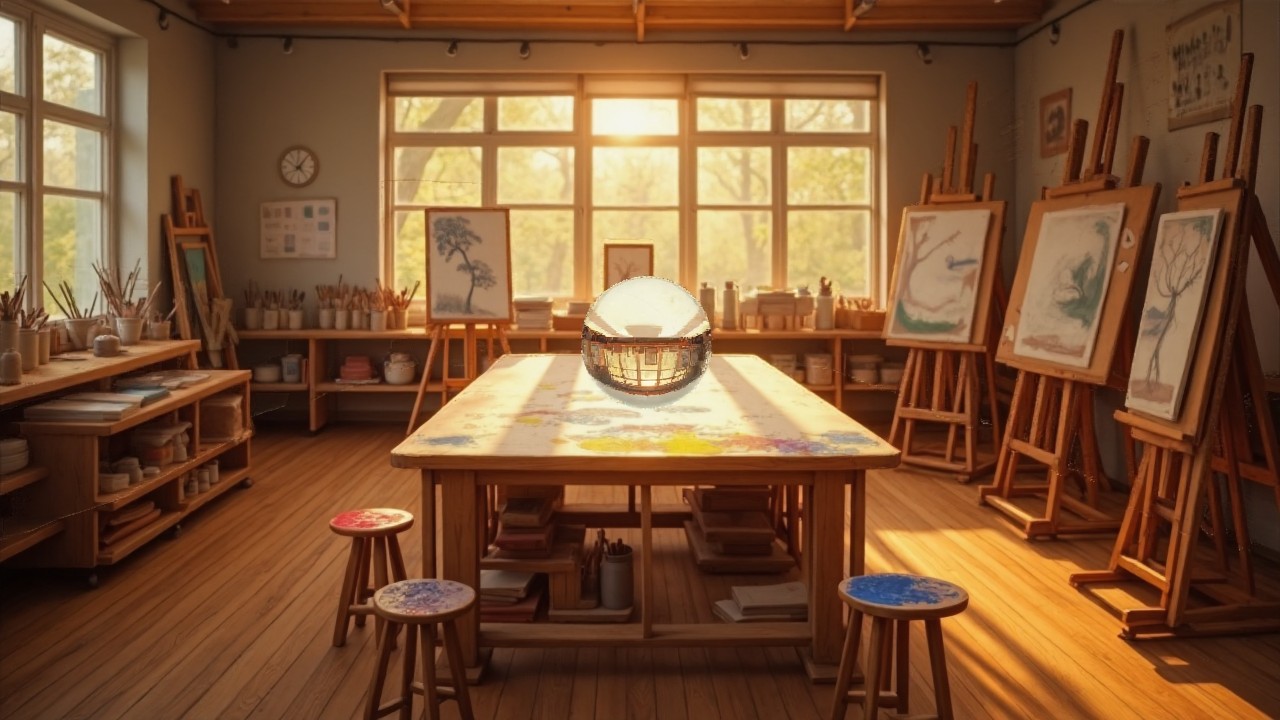} &
        \includegraphics[width=0.235\linewidth]{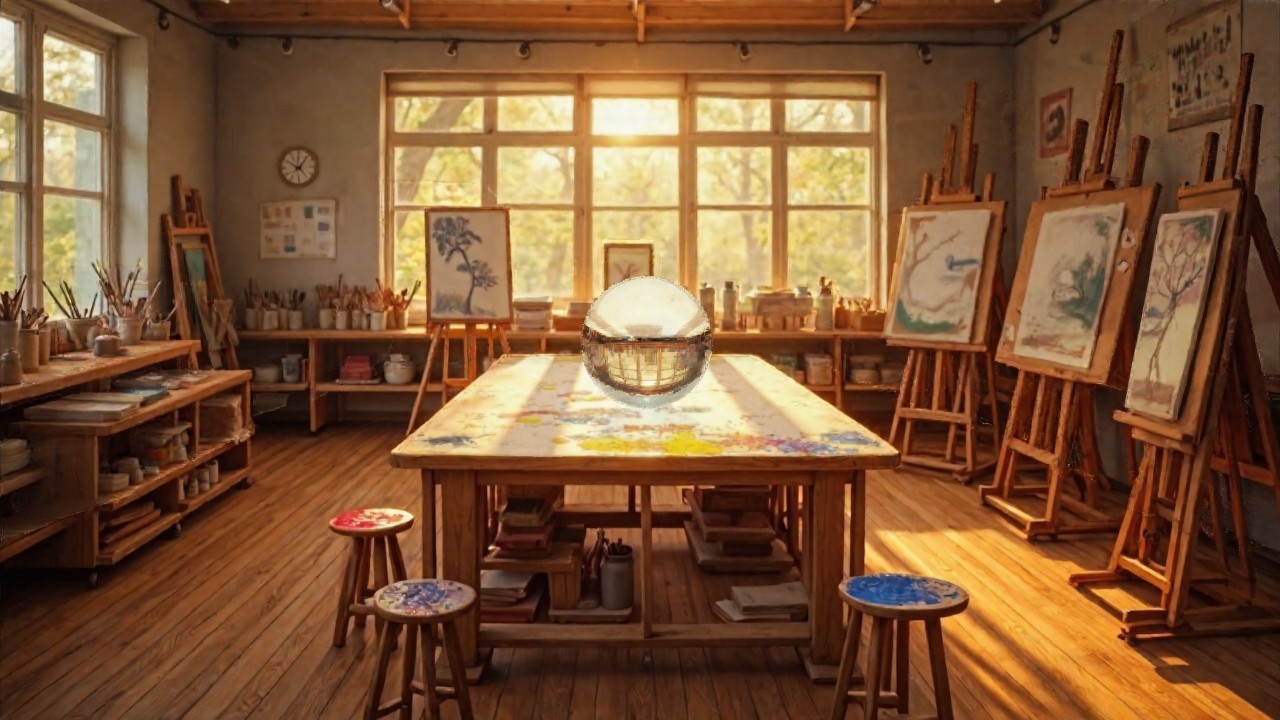} \\[1.5pt]

        \includegraphics[width=0.235\linewidth,
            trim={582pt 310pt 570pt 278pt}, clip]{figures/ablation/full.jpg} &
        \includegraphics[width=0.235\linewidth,
            trim={582pt 310pt 570pt 278pt}, clip]{figures/ablation/no_dpa.jpg} &
        \includegraphics[width=0.235\linewidth,
            trim={582pt 310pt 570pt 278pt}, clip]{figures/ablation/no_lpw.jpg} &
        \includegraphics[width=0.235\linewidth,
            trim={582pt 310pt 570pt 278pt}, clip]{figures/ablation/no_tt.jpg} \\[1.5pt]

        \includegraphics[width=0.235\linewidth,
            trim={150pt 400pt 930pt 140pt}, clip]{figures/ablation/full.jpg} &
        \includegraphics[width=0.235\linewidth,
            trim={150pt 400pt 930pt 140pt}, clip]{figures/ablation/no_dpa.jpg} &
        \includegraphics[width=0.235\linewidth,
            trim={150pt 400pt 930pt 140pt}, clip]{figures/ablation/no_lpw.jpg} &
        \includegraphics[width=0.235\linewidth,
            trim={150pt 400pt 930pt 140pt}, clip]{figures/ablation/no_tt.jpg}
    \end{tabular}
    \caption{
    Additional ablation study of the proposed method.
    We compare the full model with variants that remove individual components: detail-preserving averaging, Laplacian pyramid warping, and time travel. 
    The results are displayed without foreground object relighting, allowing the effects of each component removal to be observed more clearly.
    Removing detail-preserving averaging leads to the loss of sharp details, removing Laplacian pyramid warping introduces aliasing in regions with large stretching, and removing time travel makes the cross-view blending noticeably less natural with strong artifacts. 
    The full model avoids these issues and yields sharper and more coherent results.
    }
    \label{fig:ablation}
\end{figure}

\paragraph{Additional ablation study.}
An additional ablation study is performed where we evaluate the effect of three components: detail-preserving averaging, Laplacian pyramid warping, and time travel.
Qualitative results are shown in \cref{fig:ablation}.
To better visualize the impact of each component, we show two zoomed-in regions of an artroom scene without applying foreground object relighting. 
This avoids differences caused by relighting itself, which can introduce slight variations even when using the same random seed, since the input images differ slightly.
In the second row, we observe that removing either detail-preserving averaging or Laplacian pyramid warping leads to notably rough edges and artifacts.
In particular, removing Laplacian pyramid warping introduces aliasing.
Removing time travel, which is designed to improve blending consistency, produces obvious artifacts, including random spots, ghosting, and loss of detail.
In contrast, the full model shows negligible aliasing within the sphere, and the refractions appear smooth and physically plausible.  
The third row highlights background regions outside the sphere, where the advantages of the full model are again evident.
It produces smooth results while the other variants exhibit rough regions, noise, and visual artifacts.
This shows that each component contributes meaningfully to the overall image quality and consistency.

Another quantitative ablation study is given in \cref{tab:ablation2}, showing how removing each component affects performance.
We compare the full model without relighting with variants that remove individual components: detail-preserving averaging (DPA), Laplacian pyramid warping (LPW), and time travel (TT).
First, a note of caution about the interpretation of the metrics.
Those that evaluate the quality of the refracted region---the masked mean absolute error, the peak signal-to-noise ratio, and the LPIPS distance---are all in reference to the Blender-rendered image.
However, this pseudo-ground truth is not harmonized with respect to lighting or reflection.
That is, the light sources are in significantly different locations and have different colors, and the reflected background is entirely missing.
This means that the refracted region in the ``ground truth'' has significant expected differences in color, luminance, and structure from the truth, and so these metrics are not a fully reliable guide to performance.
Moreover, these metrics do not capture effects such as aliasing, which can be observed in the qualitative ablation study (\cref{fig:ablation}).
Nonetheless, we observe small drops in quantitative performance when the different model components are removed, especially when time travel is ablated.
Interestingly, and as previously observed, time travel smooths the images to some extent, which is helpful for the pixel-level metrics but harmful for the image-level ones (\eg, CLIP, ImageReward). 
On the whole, we direct the reader to the visual ablation study in \cref{fig:ablation}, where the actual effects of each component are more obvious.

\begin{table}[!t]
\centering
\caption{
Additional quantitative ablation study of the sphere object across six scenes (artroom, café, living room, dining room, kitchen, and office).
We compare the full model without relighting with variants that remove individual components: detail-preserving averaging (DPA), Laplacian pyramid warping (LPW), and time travel (TT).
We report the masked mean absolute error (MAE), the peak signal-to-noise ratio, and the LPIPS distance, which all measure the fidelity of the refractive region to the Blender-rendered pseudo-ground truth, as well as the CLIP and ImageReward (ImgR) scores, which assess whether the overall image is reasonable.
We note that the Blender image is not harmonized with respect to lighting or reflection, so the refracted region has significant expected differences in color, luminance, and structure, and so these metrics are not a fully reliable guide.
We also note that these metrics cannot capture effects like aliasing, which can be observed in \cref{fig:ablation}.
}
\label{tab:ablation2}
\setlength{\tabcolsep}{3pt}{
\begin{tabular}{l c c c c c}
\toprule
Model Variant & MAE$\downarrow$ & PSNR$\uparrow$ & LPIPS$\downarrow$ & CLIP$\uparrow$ & ImgR$\uparrow$ \\
\midrule
Ours & \textbf{0.0953} & \underline{18.21} & \underline{0.24} & \underline{34.22} & \underline{0.51}  \\
w/o DPA & 0.0955 & 18.15 & \underline{0.24} & 34.18 & 0.46  \\
w/o LPW & 0.0957 & \textbf{18.23} & \textbf{0.23} & 34.08 & 0.46 \\
w/o TT & 0.1002 & 17.82 & 0.26 & \textbf{34.56} & \textbf{0.58}  \\
\bottomrule
\end{tabular}}
\end{table}

\end{document}